\documentclass[10pt]{article}
\date{} 
\usepackage[hidelinks]{hyperref}
\usepackage{amsmath}
\usepackage{eqnarray,amsmath}
\usepackage{amssymb}
\usepackage{amsfonts}
\usepackage[usenames,dvipsnames]{pstricks}
\usepackage{graphicx}
\usepackage{subcaption}
\usepackage[labelformat=parens,labelsep=quad,skip=3pt]{caption}
\usepackage{qtree}
\usepackage{algorithm}
\usepackage[noend]{algpseudocode}
\usepackage[utf8]{inputenc}
\usepackage{amsthm}
\usepackage{pstricks-add,}
\usepackage{mathtools}
\usepackage{dirtytalk}

\usepackage{xcolor}
\usepackage{hyperref}

\usepackage[english]{babel} % needed for "blindtext"
\usepackage[pangram]{blindtext}

\usepackage{array}
\usepackage{booktabs}
\usepackage{rotating}
\usepackage{tikz}
\usepackage{blkarray}
\usepackage{multirow}
\usepackage{float}%Float

\usepackage{MnSymbol} % It is for \udots command

\usepackage{bbm}

\usepackage{paralist}
\usepackage{nicematrix}
\NiceMatrixOptions{
code-for-first-row = \color{blue} ,
code-for-last-row = \color{blue} ,
code-for-first-col = \color{blue} ,
code-for-last-col = \color{blue}
}

\hypersetup{colorlinks=false,linkbordercolor=red,linkcolor=green,pdfborderstyle={/S/U/W 1}}

\tikzset{use/.style={}}
\tikzset{rectangle/.append style={circle, fill=blue!25}}

\newcommand{\PosetInsertion}[3]{
    \ensuremath{
    \mathop{
    \begin{tikzpicture}[scale=0.25]
        \ifnum#1=1 \fill[black!75] (0,0.5) rectangle (0.5,1); \fi
        \ifnum#2=1 \fill[black!75] (0,0) rectangle (0.5,0.5); \fi
        \ifnum#3=1 \fill[black!75] (0.5,0) rectangle (1,0.5); \fi
        \draw[thin](0,0.5) rectangle (0.5,1);
        \draw[thin](0,0) rectangle (0.5,0.5);
        \draw[thin](0.5,0) rectangle (1,0.5);
    \end{tikzpicture}
    }}
}

\newcommand{\PosetComposition}[2]{
    \,
    \begin{tikzpicture}[scale=0.25]
        \ifnum#1=1 \fill[black!75] (0,0.5) rectangle (0.5,1); \fi
        \ifnum#2=1 \fill[black!75] (0.5,0) rectangle (1,0.5); \fi
        \draw[thin](0,0.5) rectangle (0.5,1);
        \draw[thin](0.5,0) rectangle (1,0.5);
    \end{tikzpicture}
    \,
}

\begin{document}

\title{Machine Learning and Data Analysis Using Posets: A Survey}

\author{ Arnauld Mesinga Mwafise \\ \small arnauld.mesinga@gmail.com \thanks{Corresponding author: Arnauld M. Mwafise (arnauld.mesinga@gmail.com)}}
% \\

\maketitle
%
%\begin{abstract}
%
%\noindent

% Keywords command
\providecommand{\keywords}[1]
{
  \small	
  \textbf{\textit{Keywords---}} #1
}

\begin{abstract}

\noindent Partially ordered sets (posets) are discrete mathematical structures that formalize the notion of comparison without forcing every pair of objects to be comparable. This makes them a natural representation for the many machine learning and data-analysis settings in which objects are related by dominance, containment, priority, or refinement relations rather than by a single scalar score. Over the past two decades, a substantial and fragmented literature has connected posets and lattice theory to ranking, clustering, formal concept analysis, multidimensional and multi-criteria data analysis, structured and safe learning, graph and topological deep learning, and explainable artificial intelligence, spanning a wide range of application domains. Despite this activity, the field has lacked (i) an organizing taxonomy that relates these disparate strands of work, and (ii) an up-to-date account extending through 2025--2026 that incorporates recent developments in poset-structured learning methods, including poset-structured safety layers for reinforcement learning, poset-valued neural pooling operators, functor-calculus approaches to multiparameter persistent homology, and order-theoretic formulations of abstract dynamic programming. This survey addresses both gaps. We propose a four-axis taxonomy of poset-based methods (representation, learning paradigm, data modality, and task), provide a comprehensive and comparative review of representative models and algorithms organized along this taxonomy, curate an extensive collection of datasets, software packages, and algorithmic resources, and close with a critical discussion of open theoretical and practical problems -- including model depth and expressivity, scalability--fidelity trade-offs, heterogeneity of order-structured data, and dynamicity of posets over time -- that we argue define a research agenda for the next generation of order-aware machine learning.

\end{abstract}


\section{Introduction}

Data analysis plays a pivotal role in today's data-driven world. Machine learning is used to automate the data analysis process, and enables the extraction of  valuable insights by making accurate predictions from large  complex datasets.  Data often have complex structure which can usually be endowed with a natural order. Such  complex data can be derived from text, social and behavioral sciences,  chemical structures, biological structures, images, videos, governmental and business settings, and so forth. Mathematical techniques and concepts serve as powerful tools in the development of new algorithms and methods for machine learning and data analysis. For some time now, many studies based on the use of partially ordered sets (posets) for machine learning and data analysis have emerged. Furthermore, over this time the literature on these topics has significantly expanded in volume, scope, and the range of its applications in various fields. Yet this literature is scattered across statistics, discrete mathematics, environmental science, topology, and machine learning venues, using inconsistent terminology and with no organizing framework that connects the underlying methods. This impedes the ability of researchers and machine learning scientists alike to build on prior work or to identify which poset-based tool is appropriate for a given task. In order to address this gap, we present an up-to-date overview with an emphasis on the most prominent and currently relevant works on data analysis and machine learning methods in which the role of posets has been established, including a substantial body of work published in 2025 and 2026 that has not previously been surveyed. By covering earlier works as well as these recent advances, this survey aims to provide researchers, applied mathematicians, data scientists, statisticians, social scientists, and practitioners new to the field, as well as all those aiming to keep pace with innovations in this growing and promising research area, with a solid understanding of the different conceptual and methodological approaches that incorporate posets into data analytics or machine learning applications.

\subsection{Contributions}

This survey makes the following contributions.

\begin{itemize}

\item \textbf{A new taxonomy.} No existing survey organizes poset-based machine learning and data-analysis methods along a common set of axes. We propose a four-axis taxonomy (Section~\ref{taxonomy}) that classifies work by (i) the mathematical representation of the poset used (order-relational, matrix-based, metric/distance-based, algebraic-categorical), (ii) the learning paradigm it belongs to (supervised, unsupervised, semi-supervised, reinforcement/control, or purely descriptive/statistical), (iii) the data modality it targets (tabular/multi-indicator, text, vision, graph/hypergraph, time series, or topological), and (iv) the task it solves (ranking, classification, clustering, model selection, metric computation, or explanation). This taxonomy is used throughout the survey to situate individual papers relative to one another, including papers that have not previously been discussed in a poset-and-machine-learning context.

\item \textbf{A comprehensive and comparative review.} We provide what is, to our knowledge, the most comprehensive review to date of posets in machine learning and data analysis, giving detailed descriptions of representative models, drawing explicit comparisons between competing approaches to the same problem (e.g., alternative ranking, clustering, and metric-learning schemes for partially ordered data), and summarizing the corresponding algorithms and their complexity where this information is available in the primary sources.

\item \textbf{Abundant resources.} We collect an extensive set of resources -- datasets, open-source software packages, and algorithms -- relevant to posets in data analysis and machine learning. The survey is intended to serve as a hands-on reference for understanding, using, and developing data-analysis, machine-learning, and deep-learning approaches for real-world applications involving order-structured data.

\item \textbf{A research agenda.} We discuss the theoretical foundations of posets in data analysis and machine learning, analyze the limitations of existing methods, and lay out a concrete research agenda organized around model depth and expressivity, scalability--fidelity trade-offs, heterogeneity of order-structured data, dynamicity of evolving posets, safety and constrained learning, and the emerging connections between posets, category theory, and topological data analysis.

\end{itemize}

\noindent The rest of this survey is organized as follows. Section~\ref{taxonomy} introduces the taxonomy that organizes the remainder of the paper. Section~\ref{basics} introduces and outlines the key concepts of poset theory and lattice theory. Section~\ref{mlnotes} presents a summary of research studies on machine learning and deep learning with respect to poset theory, lattice theory, and formal concept analysis, including safe and constrained learning, reinforcement learning and dynamic programming, and explainable AI. Section~\ref{clustersec} focuses on cluster analysis using posets and formal concept analysis methods. Section~\ref{multidim} presents multidimensional data analysis from a poset-theoretic perspective, and Section~\ref{exploratory} gives a unified treatment of its exploratory and descriptive aspects. Section~\ref{appsec} presents a collection of applications across various domains, including software packages, datasets, and algorithms selected for this purpose. Section~\ref{futuredir} discusses possible future research directions. Section~\ref{conclusionsec} summarizes the paper.

\section{A Taxonomy of Poset-Based Methods}\label{taxonomy}

To organize the heterogeneous body of work reviewed in this survey, we introduce a taxonomy along four largely orthogonal axes: the mathematical \emph{representation} used for the poset, the \emph{learning paradigm} the method belongs to, the \emph{data modality} it is applied to, and the \emph{task} it addresses. Table~\ref{tab:taxonomy} summarizes the taxonomy together with representative sections of this survey and representative references; most methods can be located at the intersection of one entry from each axis. We use this taxonomy in the remainder of the survey to situate each method discussed.

\begin{table}[ht]
\centering
\caption{A four-axis taxonomy of poset-based methods in machine learning and data analysis.}
\label{tab:taxonomy}
\small
\begin{tabular}{@{}>{\raggedright\arraybackslash}p{2.6cm} >{\raggedright\arraybackslash}p{6.9cm} >{\raggedright\arraybackslash}p{3.9cm}@{}}
\toprule
\textbf{Axis} & \textbf{Categories} & \textbf{Representative sections} \\
\midrule
Representation &
Order-relational (Hasse diagrams, cover relations); matrix-based (incidence, cover, mutual-ranking-probability, poset matrices); metric/distance-based (poset metrics, model-oriented graph distances); algebraic/categorical (operads of posets, operads of poset matrices, poset cocalculus, order polytopes)
& \S\ref{basics}, \S\ref{mlnotes}, \S\ref{futuredir} \\
\addlinespace
Learning paradigm &
Unsupervised (clustering, FCA, concept lattices); supervised (classification, learning to rank); semi-supervised; reinforcement learning and control (safe learning, abstract dynamic programming, poset/lattice generation); purely descriptive/statistical (depth functions, multi-indicator ranking)
& \S\ref{mlnotes}, \S\ref{clustersec}, \S\ref{multidim} \\
\addlinespace
Data modality &
Tabular/multi-indicator (socio-economic, environmental); text/NLP (semantic parsing, dependency order); vision (pose estimation, image classification); graph/hypergraph (GNNs, partial-order hypergraphs); time series (event sequences); topological (multiparameter persistence)
& \S\ref{mlnotes}, \S\ref{appsec} \\
\addlinespace
Task &
Ranking and rank aggregation; classification and model selection; clustering and concept discovery; metric/distance computation and isomorphism testing; generation and enumeration; explanation and auditing (XAI)
& \S\ref{mlnotes}, \S\ref{multidim}, \S\ref{descriptive}, \S\ref{futuredir} \\
\bottomrule
\end{tabular}
\end{table}

The taxonomy is intentionally coarse-grained: individual papers frequently combine categories from more than one axis (for example, a poset-pooling convolutional filter is simultaneously an algebraic/categorical representation, a supervised-learning paradigm, a vision-data modality, and a classification task). We use it as an organizing device rather than a strict partition, and we return to it in Section~\ref{futuredir} when discussing gaps in the current literature -- most notably the comparative scarcity of methods addressing heterogeneous or dynamically evolving posets. We also note that the \emph{representation} axis is not strictly partitioned in practice: poset matrices are simultaneously a matrix-based computational representation and, once equipped with composition operations, an algebraic/categorical one~\cite{operadposetmatrices}, and generation and isomorphism testing -- added to the \emph{task} axis to accommodate recent work discussed in Section~\ref{futuredir} -- are dual problems in the sense that a scalable generator is only as useful as its ability to cheaply determine whether a newly generated candidate duplicates one already found~\cite{posetrlgen,posetisodecomp}.

\section{Basics of Partial Order Theory}\label{basics}

Order theory is a fundamental branch of mathematics which  focuses on the arrangement of elements within various structures based on certain rules that can intuitively be captured using binary relations. It plays a crucial role in understanding mathematical sequences, hierarchies, and the organisation of data in fields such as computer science, economics, social sciences, environmental sciences,  biomedical sciences etc. Generally, order theory deals with the structure and properties of partial orders which occurs naturally in subset relations and integer relations. The analysis of partially ordered sets is a well studied topic in discrete mathematics particularly in combinatorics  
\cite{posetcombin}.   The key concept in partial order theory is the  \lq{concept of comparison}\rq. Objects are mutually compared, and  transitivity is assumed \cite{failedstate}. That is  if object x is better than object y, and object y is better than object z then object x is better than object z. However, comparison can be independently possible without the requirement of transitivity in some scenarios. For instance, in the theory of tournaments in sports, it is possible for  team A to beat team B and team B beats team C without necessarily implying that team A beats team C.
Mathematically, we can summarize the definition of partial order sets as follows. A nonempty set $X$ of size $n$ which is endowed with the order relation ``$ \preccurlyeq$''(comparison operator) is called a partially ordered set or simply a poset, 
 denoted by $ P=(X, \preccurlyeq )$, if the relation $ \preccurlyeq $ satisfies the following conditions:
 \begin{enumerate}
  \item reflexive i.e. $x\in X, x\preccurlyeq  x;$ 
  \item  antisymmetric i.e.  $x,y\in X$ if $x\preccurlyeq  y$ and $y\preccurlyeq  x\implies x=y;$ 
  \item transitive i.e.  $x,y,z\in X$ if $x\preccurlyeq y$ and $y\preccurlyeq z\implies x\preccurlyeq z.$
 \end{enumerate}
Moreover, reflexivity means that a given object can be compared with itself. Anti-symmetry means that if both comparisons are valid, i.e., y is better than x and at the same time, x is better than y, then this axiom requires that x is identical to y. Transitivity means that if the objects are characterized by properties which are at least ordinal scaled, then any measurable quantity like height, length, price, tournament team rankings, order of product quality, order of agreement or satisfaction  etc., implies transitivity\cite{inequalitiesEU}.
If either $x\preccurlyeq  y$  or $y\preccurlyeq  x$ then  $x$ and $y$ are  \textit{comparable.} For partially ordered sets it is not required that all  pairs $x,y\in X$ are comparable  either as $x\preccurlyeq  y$  or $y\preccurlyeq  x.$ In the case where all pairs $x,y\in X$  are comparable, then the set $X$  is referred to as a \textit{totally ordered set} or an \textit{ordered set.} The numerical comparison operator $ \leq$ is used to denote the relationship  between any two pairs of elements(objects) within an ordered set. With an ordered set, ranking is always possible  and  any sorting algorithm may be applied using $ \leq$ as the comparison operator for choosing whether to swap the objects in an ordered list. Classical problems of sorting and searching assume an underlying linear ordering of the objects being compared. An important result in order theory is that every partial order can be extended to a linear ordering or a total order. In order theory, a linear extension of a partial order is a total order (or linear order) that is compatible with the partial order. More formally, a \textit{linear extension}\cite{linearextensions,RiskeyLinearEXt} of a partially ordered set P is a permutation of the elements $p_1, p_2, p_3,...$ of P such that $p_i<p_j$  implies  $i<j$. For example, the linear extensions of the partially ordered set $((1,2), (3,4))$ are $1234, 1324, 1342, 3124, 3142,  \text{and}\;3412,$  all of which have $1$ before $2$ and $3$ before $4$.  
Some key concepts on partially ordered sets $ P=(X, \preccurlyeq )$ are described as follows:

 \begin{itemize}

 \item \textit{Maximal elements (or objects)}  of the  poset  $P$ are the set of elements  $x\in X$ in which no other element  $y\in X$  satisfies  the relation  $y>x.$  If $x$  is the only maximal element then it is referred to as the  “greatest" element. In a totally ordered set, the terms maximal element and greatest element  coincide.

 \item \textit{Minimal elements (or objects)}  of the  poset  $P$ are the set of elements $x\in X$ in which no other element $y\in X$  exist such that  $y < x$. If $x$ is the only minimal element,  then $x$  is referred to as the  “least” element. In a totally ordered set, the terms minimal element and least element coincide.

\item \textit{Chain} is a subset C of X, in which any object (or element)  is mutually comparable with all other elements of C. That is, a chain C is a poset such that no element can be added between any two of its elements without losing the property of being totally ordered. 

\item \textit{Antichain} is a subset $C^{\prime}$ of X, in which each object (or element)  of $C^{\prime}$ is mutually incomparable with all
other elements  in $C^{\prime}$. That is, all the elements of the set $C^{\prime}$ are pairwise incomparable and as such they can never have the property of being totally ordered. 

\item  A \textit{cover relation} for ${\mathcal P}=(X, \preccurlyeq )$ is the set of pairs $(x,y)$ such that $x,y\in X,$ and $y$ covers $x$ whenever there exist no element $z\in X$  such that $x\prec z\prec y.$

\end{itemize}

\noindent Beyond this relational description, a poset also admits an equivalent algebraic representation as a binary matrix, which is the representation underlying some of the algorithmic and machine-learning literature surveyed in this paper. Following the formulation in~\cite{posetisodecomp,mwafise2022poset}, a \textit{poset matrix} $PM \in \{0,1\}^{n\times n}$ encodes a poset $P=(X,\preccurlyeq)$ on $n$ elements via $PM_{ii}=1$ (reflexivity), $PM_{ij}=1 \Rightarrow PM_{ji}=0$ for $i\neq j$ (antisymmetry), and $PM_{ij}=1 \wedge PM_{jk}=1 \Rightarrow PM_{ik}=1$ (transitivity), the last of which can be written compactly in Boolean arithmetic as $PM^2 \preccurlyeq PM$. If the elements of $P$ are topologically ordered, $PM$ takes a unit lower-triangular form, though this representation is not unique: simultaneous permutation of rows and columns yields another valid poset matrix for the same underlying poset, which is precisely why testing whether two poset matrices represent isomorphic posets is a nontrivial algorithmic problem in its own right (Section~\ref{futuredir}). Posets also admit a \textit{dual} matrix representation $PM^{*}=[p_{n+1-j,\,n+1-i}]$, with a poset called self-dual when it is isomorphic to its own dual. This matrix-theoretic view of posets underlies not only the classical incidence- and rank-matrix methods used throughout Sections~\ref{multidim} and~\ref{descriptive}, but also a growing body of purely combinatorial and algorithmic work that treats poset matrices as first-class objects of study: algebraically, via operad structures defined directly on the collection of poset matrices~\cite{operadposetmatrices}, and algorithmically, via hierarchical matrix-decomposition methods for poset isomorphism testing~\cite{posetisodecomp} and reinforcement-learning-based generation of posets and lattices~\cite{posetrlgen}, both discussed further in Section~\ref{futuredir}.

\noindent Partially ordered sets are graphically visualized using Hasse diagrams. A Hasse diagram can be derived from a directed acyclic graph, where the vertices are representing the objects and
a line relates object $x$ with $y$ whenever $x \prec y$. In the case of a transitivity relation  whenever  $x\prec y$ and $y\prec z$ then it suffices to draw a line only for $x\prec y$ and $y\prec z$ and not for $x\prec z$. Hasse diagrams can be viewed as a powerful tool for unifying ideas and concepts. Hasse diagrams reveal an intricate network of comparabilities and incomparabilities, maximal and minimal elements. In addition, Hasse diagrams are also characterized with a structure that comprises levels, chains and antichains.

Hasse diagram:  \begin{minipage}[c]{.20\textwidth}
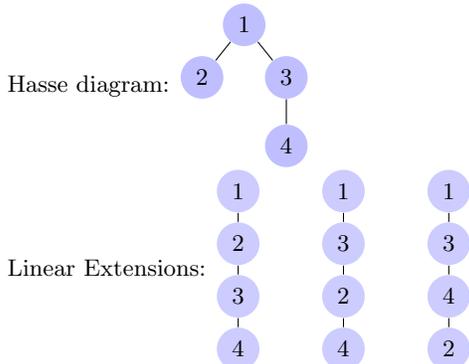
\begin{tikzpicture}
  [scale=.7,auto=center,every node/.style={circle,fill=blue!20}]

\node[rectangle] (c1) at (6.2,3) {2};
  \node[rectangle] (c2) at (7.8,1.7) {4};
  \node[rectangle] (c4) at (7.8,3) {3};
  \node[rectangle] (c8) at (7,4)  {1};

      \draw (c4) -- (c2);
    \draw (c8) -- (c4);
   \draw (c8) -- (c1);

 \end{tikzpicture}

\end{minipage}

Linear Extensions:  \begin{minipage}[c]{.20\textwidth}
\begin{tikzpicture}
  [scale=.7,auto=center,every node/.style={circle,fill=blue!20}]
\node(d1) at (10,4) {2};
\node(d6) at (10,3)  {3};
\node(d7) at (10,5) {1};
\node(d8) at (10,2) {4};
\draw (d6) -- (d1);
\draw (d1) -- (d7);
\draw (d6) -- (d8);

\node(d11) at (12,4) {3};
\node(d66) at (12,3)  {2};
\node(d77) at (12,5) {1};
\node(d88) at (12,2) {4};
\draw (d66) -- (d11);
\draw (d11) -- (d77);
\draw (d66) -- (d88);

\node(d111) at (14,4) {3};
\node(d666) at (14,3)  {4};
\node(d777) at (14,5) {1};
\node(d888) at (14,2) {2};
\draw (d666) -- (d111);
\draw (d111) -- (d777);
\draw (d666) -- (d888);

\end{tikzpicture}
\end{minipage}

\noindent Finite posets can also be  represented by  square matrices which comprise:  incidence \cite{riordanposets,operadposet}, cover \cite{covermatrices},
and mutual ranking probability matrices\cite{probmatrices}.  The mutual ranking probability matrices are  a class of matrices which comprise and convey information on the dominance among statistical units\cite{probmatrices}.  All the three main types of poset matrices are useful in conveying essential information on the structure of the order relation from different perspectives.
Posets can also be subdivided into variour classes such as series-parallel posets, semantic posets, and ranking posets etc. Series-parallel posets are characterized by the  N-free partial order structure. Series-parallel partial orders have been applied to machine learning of event sequencing in time series data \cite{seqdata}. Semantic posets are used for compositional generalization in semantic parsing  \cite{decodecomp}. 
The ranking poset $\mathfrak{M}_n$\cite{rankposet}	 is the poset in which the elements are all possible cosets $\mathfrak{G}_{\lambda}\pi $,
where $\lambda$ is an ordered partition of $n$ and $\pi\in\mathfrak{G}_n.$ The ranking poset is useful in model selection and validation in machine learning.
More recently, Fueyo et al.~\cite{leveledposets} introduced \textit{leveled partially ordered sets}, a class of posets equipped with a level (height) function generalizing Birkhoff's classical notion of element height, originally motivated by the problem of ordering the layers of points generated during additive manufacturing (3D printing). A leveled poset satisfying a Jordan--Dedekind-type chain condition can always be completed into a bounded lattice by adding links to its existing order relation rather than by introducing new elements, which makes this class attractive for applications, discussed further in Section~\ref{futuredir}, in which data arrives incrementally in layers or batches. From an algebraic and categorical perspective, posets can also be studied as \textit{algebras over the operad of posets}: Arciniega-Nev\'{a}rez, Berghoff, and Dolores-Cuenca~\cite{operadwixarika} use the language of operads to formalize composition of posets and study a nontrivial suboperad, the Wix\'{a}rika posets, together with its associated algebras, illustrating how order-theoretic combinatorics can be organized using the same categorical toolkit used elsewhere in machine learning (e.g., for compositional model architectures).

Posets are natural models for many statistical applications \cite{statmodel, dataAnalytic, montecarlo, modelselection}. Partial orders are also the natural mathematical structure for comparing multivariate data that lacks a natural order \cite{multivariate}. For instance, multivariate data on colours has no natural ordering and therefore the rank features in colour spaces can be modelled as partial orderings, and this approach is generalizable in similar contexts \cite{posetcolorspace}. It also plays a key role in decision making in environmental informatics and computational chemistry \cite{envchem, rankhazard,chemsimilar,chem,chemposets, rankenvsoft, incomparable}, social sciences\cite{socialorder, rankinadequate} etc. Moreover, in data analysis a typical  binary relation is defined for a set of objects and a set of attributes. For instance,  objects are transactions in a supermarket, attributes are items in the supermarket, and the relation consists of pairs (transaction, item) such that the item occurs in the transaction\cite{orderdataanalysis}.

The study of partially ordered sets and lattices has been covered by a significant number of mathematical publications. Birkhoffs (1967) \cite{barrettlattice} and
Gr$\ddot{\text{a}}$tzer's (1978)\cite{gratzerlattice} books on lattice theory are considered to be classics. Lattices\cite{latticehiech} are a broad structured class of posets. Lattice theory has shown its ability to contribute to solving problems in a wide variety of applications. Formally, a lattice\cite{latticehypothesis} is a poset, in which every pair of elements has both a least upper bound and a greatest lower bound. In other words, it is a structure with two binary operations: join and meet. The following sub-definitions summarize a lattice structure.
A meet semilattice is a poset for which any two elements a and b have a greatest lower bound denoted $a\wedge b$. The greatest lower bound of a and b is the largest element that is still less than both of them. In a lattice, the greatest lower bound must be unique. The greatest lower bound of a and b is also called the meet or infimum of $a$ and $b$. A join semilattice is a poset for which any two elements $a$ and $b$ have a least upper bound, denoted $a\vee b$. The least upper bound of $a$ and $b$ is the smallest element that is still greater than both. In a lattice, the least upper bound must be unique.
The least upper bound of $a$ and $b$ is also called the join or supremum of $a$ and $b$. If a poset is both a meet semilattice and a join semilattice, then the poset is also a lattice.

Example of a lattice : \begin{minipage}[c]{.20\textwidth}
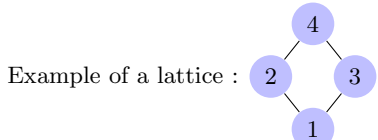
\begin{tikzpicture}
  [scale=.7,auto=center,every node/.style={circle,fill=blue!20}]

\node[rectangle] (c1) at (6.2,3) {2};
  \node[rectangle] (c2) at (7,2) {1};
  \node[rectangle] (c4) at (7.8,3) {3};
  \node[rectangle] (c8) at (7,4)  {4};

      \draw (c1) -- (c2);
        \draw (c2) -- (c4);
    \draw (c8) -- (c4);
   \draw (c8) -- (c1);

 \end{tikzpicture}

\end{minipage}

In the example above: The maximal elemnt is $4$ and the greatest element is $4.$ Similarly, the minimal element is $1$ and the least element is $1.$

\section{Machine learning using posets and formal concept analysis}\label{mlnotes}

\subsection{Overview}

In general, machine learning involves the creation of knowledge from a training dataset by using models and algorithms to learn from data in order to make predictions, find patterns, classify data and automate decision making processes. Machine learning has progressed significantly over the past two decades, and this has been driven by the development of new learning algorithms, theory, and high performance computing architecture \cite{machinehighcompute}. There are many different kinds of machine learning algorithms. The most well-known ones are supervised, unsupervised, semi-supervised, and reinforcement learning. Supervised learning is useful in scenarios for which some relationship between input and output labelled data has been established, unsupervised learning is effective for uncovering hidden patterns in unlabelled data, and semi-supervised learning utilizes a combination of labelled and unlabelled data to train models. The goal of any machine learning algorithm is to optimize the performance of a system when handling new instances of data through user defined programming logic in a given environment \cite{mlperformance}.  In order to build more efficient machine learning algorithms or models so as  to improve on their performance and accuracy in most predictive task, various ensemble learning methods have been developed.  Ensemble learning\cite{ensembLE} is an approach which aggregates the predictions of  two or more models fitted to the same data in order to minimize error. In the area of supervised learning, the most popular ensemble machine learning techniques are bagging, boosting, and stacking,  which  use multiple learning algorithms or models to produce one optimal predictive model. Ensemble methods can broadly be categorized into  sequential ensemble techniques and parallel ensemble techniques. Bagging(bootstrap aggregating) aims to reduce variance by adopting parallel ensemble learning on homogenous weak learners, boosting aims to reduce bias by adopting sequencial ensemble learning on homogenous weak learners, and stacking aims to improve prediction accuracy by adopting parallel ensemble learning on heterogeneous weak learners. Sequential ensemble techniques generate base learners in a sequence that are characterized by the dependence between the base learners. On the other hand the  parallel ensemble technique   base learners are generated in parallel, and  in such a way that independence is enforced between the base learners.  Dagging(Disjoint samples aggregating) is another useful type of parallel ensemble technique that was introduced in \cite{daggingF}. Dagging is similar to bagging, but instead of bootstrapping, it uses stratified sampling by creating a number of disjoint groups and stratified data from the original learning data set, with each considered as a subset of learning. In unsupervised learning, ensemble clustering\cite{ensembleRR, EnsembleBB} aims at combining the outputs of several clustering algorithms to form a single clustering structure (crisp or fuzzy partition, hierarchy).

In recent years, deep learning\cite{deeLSurvey}  has been the most popular computational approach in the field of machine learning and has unique advantages when dealing with high-dimensional nonlinear problems. The performance of machine learning algorithms relies heavily on the representation, and finding a good representation can facilitate the discovery of structure in the input data by the learning algorithm. Deep learning as a model is a specific type of representation learning\cite{representationL} that uses layered algorithms known as artificial neural networks, which attempts to mimic the human brain through a combination of data inputs, weights, and biases, to learn representations of data.  It has succesfully been used  in a wide range of disciplines such as cybersecurity, natural language processing, visual recognition, machine translation, robotics, recommendation systems etc. There are several types of deep learning algorithms. These are: Convolutional Neural Network(CNN), Long Short Term Memory, Graph Neural Networks etc. Traditionally, machine-learning and deep learning algorithms have been used on data represented in Euclidean space such as  image data which can be represented as a regular grid of pixel values.  On the other hand,  graph data cannot be represented on Euclidean space \cite{grapheuclid}. Graph Neural Networks (GNNs)\cite{gnnsurvey}  are a class of deep learning methods designed to perform inference on data described by graphs which are non-Euclidean (non-metric) structured. They are highly influenced by CNN which  learns features by inspecting neighboring pixels(nodes in GNN) in three dimensional image data for classification and object recognition purposes. Posets are special classes of directed graphs. There is also a growing  use of graph neural networks for the modeling and analysis of partially ordered data   \cite{hypergraph,weigtposet,decodecomp,gnnposet11, gnnposet22}. More recently, the problem of machine learning on meet/join lattices and posets was extensively studied in  \cite{swiss}. The main focus of the study was on developing methods that are based on generalized convolutions and sparse Fourier transforms algorithms  which are capable of learning set functions. A complementary, more architecture-level use of order theory in deep learning is given by Dolores-Cuenca et al.~\cite{poolingorder}, who observe that every four-point poset corresponds to an order polytope and, via the correspondence between integer-valued neural networks and tropical rational functions, to a $2\times2$ convolutional filter; the resulting \textit{poset pooling} filters can be inserted into any neural network and are reported to update weights during backpropagation with greater precision than average, max, or mixed pooling, without adding trainable parameters, while the composition of such poset-neural architectures can itself be organized using the operad of posets discussed above~\cite{operadwixarika}.
The first lattice-based machine learning models were introduced by V.K. Finn\cite{finnjsm, finncomputer, finnsemantics}. It was based on a closure system which uses the JSM(John Stuart Mille)-method of automated hypothesis generation. In this model, positive hypotheses are searched among intersections(similarity as meet operations) of positive example descriptions (object intents), likewise for negative hypotheses \cite{jsmoperations}. The JSM method also identifies data patterns by means of induction and it can also be used as a logical rule-based classification method  formulated in terms of formal concepts \cite{MathConcepts}.
Machine learning has been studied from many perspectives in the context of poset theory and lattice theory. These include : machine learning performnce comparison\cite{depthfns,rankposet}, multilabel classification \cite{dempster}, ensemble  classification\cite{rankposet}, sequential clasification \cite{seqdata, sequence, dataAnalytic, sequenceexp}; natural language processing \cite{weigtposet, compgen, decodecomp}, deep unsupervised learning \cite{silarlearn}, semi-supervised learning\cite{hypergraph}, learning to rank\cite{predrankABs}, time series modelling \cite{bayesianpartialorder,seqdata}, ensemble clustering \cite{dempster}, model selection\cite{modelselection}, topological deep learning \cite{TDL}, attention-based neural networks \cite{HOA}.

\subsection{Safe Reinforcement Learning, Control, and Order-Theoretic Dynamic Programming}\label{safeRL}

Two closely related bodies of work use posets not merely to represent data but to represent the \textit{constraints or the mathematical machinery of the learning problem itself}. The first concerns safety-constrained control and reinforcement learning. Deploying learning-based controllers in safety-critical robotic systems typically requires enforcing multiple safety constraints simultaneously, and existing approaches usually do so either uniformly or via a fixed priority order, which can be infeasible or brittle when safety requirements are genuinely heterogeneous. Wong, Xiao, and Rus~\cite{posafenet} instead formalize this setting as \textit{poset-structured safety}: safety constraints are modeled as a partially ordered set in which some constraints are mutually comparable (one strictly takes precedence) while others are inherently incomparable, and safety composition is treated as a structural property of the policy class rather than as an externally imposed schedule. Building on this formulation, they propose PoSafeNet, a differentiable neural safety layer that enforces safety via sequential closed-form projection under poset-consistent constraint orderings, allowing the controller to adaptively select or mix valid safety executions while provably preserving priority semantics. This is a direct, and to our knowledge novel, instance of a neural architecture whose layer-level computation is explicitly organized by a partial order supplied by the problem specification, complementing the poset-pooling architectures of Section~\ref{mlnotes} in which the poset instead organizes the receptive field of a convolutional filter.

The second body of work uses partial orders as the foundational structure of dynamic programming itself, rather than of the state or action space in the usual sense. Sargent and Stachurski~\cite{dposets} represent a dynamic program abstractly as a family of policy operators acting on a partially ordered set, and derive an optimality theory -- covering value function iteration and policy iteration -- from purely order-theoretic assumptions; because the framework does not presuppose a real-valued Bellman equation, it uniformly covers standard Markov decision processes together with nonlinear recursive-preference models, robust-control objectives, and distributional dynamic programs whose value ``functions'' take values in a space of distributions. Peng, Stachurski, and Yang~\cite{abstractdynprog} strengthen this framework by pairing the order-theoretic assumptions with topological and metric stability conditions (global stability and contractivity of the policy operators), from which they recover convergence of value function iteration, Howard policy iteration, and optimistic policy iteration, together with a proof that stationary policies dominate nonstationary policy plans under weak assumptions; applications include optimal stopping without discounting and Bayesian sequential analysis, for which their results weaken existing assumptions. These two papers do not analyze machine-learned models directly, but they establish exactly the order-theoretic convergence theory that any future poset-structured reinforcement learning architecture -- such as PoSafeNet -- would need to invoke to obtain formal optimality or convergence guarantees, and we return to this connection in Section~\ref{futuredir}.

A third and distinct role for RL in this space is as a generator of poset-structured objects rather than a consumer of a fixed poset-structured problem specification. Mwafise~\cite{posetrlgen} trains a policy-gradient agent to construct finite lattices, join-semilattices, and meet-semilattices directly, motivated by the combinatorial explosion of exact lattice enumeration: the number of non-isomorphic lattices on $n$ elements grows from $5{,}994$ at $n=10$ to over $23$ trillion at $n=20$ (OEIS A006966), which confines exhaustive, orderly-generation algorithms to $n \lesssim 24$. Rather than visiting graph nodes sequentially, as in GraphRNN-style autoregressive generators, the agent visits element \emph{pairs} in a freshly randomized order at every rollout and chooses among ``no relation'', $u\!\to\!v$, or $v\!\to\!u$ under a hard pre-softmax legality mask that enforces the poset axioms structurally; an independent, exact tensor-based verification oracle then certifies the target closure property (lattice, join-, or meet-semilattice) for every candidate, so that no confirmed structure can be a false positive regardless of policy quality. The central empirical finding is that this order-agnostic construction alone -- independent of any other architectural choice -- improves the lattice-discovery rate at $n=20$ by a factor of $4.7$ (from $10.5\%$ to $49.7\%$) relative to fixed sequential visitation, because sequential, node-by-node construction structurally starves early-visited elements of the admissible relations a valid join or meet requires, regardless of how well the policy is trained. Combined with property-aware reward shaping that discourages the easily-reached distributive lattice class while rewarding the rarer modular non-distributive ($M_3$) and non-modular semimodular classes, an entropy-augmented Generalized Advantage Estimation variant, and GPU-vectorized batch generation, the framework sustains mean discovery rates of $60$--$80\%$ (peaking at $80$--$100\%$) across $n \in \{30,40,50\}$, a regime roughly double the practical ceiling of exact enumeration. The verified-oracle design is methodologically significant beyond this specific application: because the sparse, non-differentiable validity predicate is checked exactly rather than learned, training instability in the generator can depress discovery rates but can never manufacture a false positive, a separation-of-concerns principle directly analogous to the exact verification used elsewhere in poset-based safe learning (PoSafeNet's projection step, above) and one we return to as a broader design pattern in Section~\ref{futuredir}.

\subsection{Posets in Explainable Artificial Intelligence}\label{xaiSec}

Posets are increasingly used in explainable artificial intelligence (XAI) to manage complexity, provide transparency, and represent hierarchical structure in model behavior and in data, precisely because they allow a model to represent that two objects, features, or explanations are \textit{genuinely incomparable} rather than forcing an arbitrary total order between them. This is a natural fit for interpretability, where forcing a single linear ranking (as in a scalar composite score or a single feature-importance ranking) can hide the fact that different explanations are valid along different, mutually incompatible dimensions. Several strands of the material already reviewed in this survey can be read specifically through this lens.

\begin{itemize}

\item \textit{Transparent decision-making.} Rather than reducing complex, multi-dimensional data to a single score, the multi-indicator and multidimensional poset methods of Section~\ref{multidim} preserve the original comparability structure of the data and make explicit which objects dominate which others and along which criteria, which is the central requirement of a transparent, auditable ranking or scoring procedure~\cite{compindic,measuredev}.

\item \textit{Formal concept analysis (FCA) and concept lattices.} As discussed in Section~\ref{fcaTT}, FCA organizes a formal context into a concept lattice whose Hasse diagram directly exposes the hierarchy of object--attribute concepts, providing a visual and mathematically grounded explanation of how instances group together~\cite{FCABook,latticehiech}. Where full concept lattices are too large to interpret directly, reduced skeleton structures known as AOC-posets (posets of attribute- and object-concepts) retain the information relevant for interpretation while substantially reducing the number of nodes that must be inspected.

\item \textit{Interpretable neural architectures.} The poset-pooling filters of Dolores-Cuenca et al.~\cite{poolingorder} (Section~\ref{mlnotes}) are explicitly motivated by interpretability: because each filter corresponds to an explicit poset and an associated order polytope, the effect of the filter on backpropagation can be analyzed geometrically rather than treated as a black box, in contrast to standard max or average pooling.

\item \textit{Model selection as a poset.} As discussed in Section~\ref{basics} and Section~\ref{descriptive}, model-selection problems can themselves be organized as a poset in which a ``least'' element represents a null or baseline model and the order relation encodes model refinement~\cite{modelselection}; this makes explicit, and auditable, which comparisons between candidate models are actually licensed by the data rather than being an artifact of an arbitrary total ranking.

\item \textit{Handling incomparability as a consistency constraint.} Across the applications reviewed in Sections~\ref{appsec}, incomparability is repeatedly identified not as a shortcoming to be resolved but as information: two objects, cities, chemicals, or models may simply not be comparable given the available criteria~\cite{bridgeposets,failedstate}. Explicitly representing this, rather than silently resolving it via aggregation, is a recurring argument in the poset literature for why posets support more honest, trustworthy explanations than composite-index or single-ranking approaches.

\end{itemize}

\noindent Taken together, these threads suggest that posets are already, if implicitly, a recurring representational choice in interpretable and auditable machine learning, even where the original works are not framed using XAI terminology; we discuss the resulting research opportunities further in Section~\ref{futuredir}.

\subsection{Performance Comparison and Model Selection}

Comparing the performance of competing learning algorithms is one of the most common tasks in machine learning, yet it is typically reduced to a single scalar criterion (e.g., accuracy), which discards information whenever algorithms trade off differently across multiple performance measures. Blocher and Schollmeyer~\cite{depthfns} address this by recasting algorithm comparison as a problem over partially ordered data. They extend classical statistical \textit{depth functions}~\cite{stasdepth} -- originally developed to generalize the univariate notions of median, quantiles, and order statistics to multivariate data -- to the space of partial orders itself, via an adaptation of simplicial depth that they call the union-free generic (ufg) depth. A key property of the ufg depth is that it treats a partial order holistically rather than decomposing it into pairwise comparisons, which allows samples of poset-valued random variables (for example, the partial order induced by several performance measures over a set of algorithms) to be analyzed descriptively without collapsing them onto a single criterion such as balanced accuracy. This depth-function perspective is complementary to the multidimensional and descriptive poset-analysis methods discussed in Sections~\ref{multidim}--\ref{descriptive}, applied here specifically to the problem of algorithm and model comparison.

A related but distinct use of posets is in \emph{model selection} itself, where the object of interest is not a ranking of performance scores but the space of candidate models. Taeb et al.~\cite{modelselection} organize classes of models as partially ordered sets in order to address model structures that lack an underlying Boolean logical structure -- a property that is otherwise a prerequisite for controlling the false-positive error rate of standard model-selection procedures. This poset-structured view of model selection recurs later in the survey, both as a component of formal concept analysis (Section~\ref{fcaTT}) and as one of the recurring examples of order-theoretic explainability discussed in Section~\ref{xaiSec}.

\subsection{Natural Language Processing}

Compositionality~\cite{comparse} -- the ability to recombine familiar units such as words into novel phrases and sentences -- is a long-standing goal in natural language understanding, and existing neural sequence models are known to generalize poorly along this dimension~\cite{compProbs}. To improve the compositional generalization of neural encoder--decoder architectures, Guo et al.~\cite{decodecomp} introduce a hierarchical poset decoding paradigm that explicitly accounts for the partial permutation invariance of semantic representations, treating the decoding target itself as a poset rather than a linear sequence. Compositional generalization is particularly relevant to tasks involving complex utterances with multiple equivalent meaning representations, such as semantic dependency parsing, in which a sentence is mapped to a directed graph capturing the semantic relations between its words~\cite{semanticdep}. In a closely related direction, Dyer~\cite{weigtposet} converts dependency trees into surface word orders using syntactic word embeddings combined with edge-weighted posets, training a graph neural network to learn the poset's edge weights on the Universal Dependencies (UD) corpora (Section~\ref{appsec}). Together, these results illustrate that posets can serve not only as a representation of the \emph{output} of a language model, as in the decoding case, but also as the learned structure that mediates between a semantic representation and its surface realization.

\subsection{Classification}

Classification -- the prediction of a discrete target label -- has long been a central concern of machine learning, and many classification tasks are naturally reframed as ranking problems once the target admits a partial order. In the simplest case, a single label $y \in Y$ is associated with a covariate $x$; \emph{conditional ranking} generalizes this by instead assigning $x$ a full or partial ranking of the items in $Y$. Lebanon and Lafferty~\cite{rankposet} propose a unifying algebraic framework for ensemble classification and conditional ranking based on the ranking poset introduced in Section~\ref{basics}: the collection of all full and partial rankings of a fixed set of items, partially ordered by refinement. The structure of this poset induces natural, order-invariant distance functions that generalize both Kendall's Tau~\cite{kendalltau} and the Hamming distance, from which a generative probabilistic model can be derived and subsequently used for model selection and validation~\cite{rankposet}.

A second strand of work derives classification rules directly from lattice structure. Sahami~\cite{classrule} introduces Ruleamer, an inductive algorithm that takes as input a lattice $L$, a set of instance labelings $C$ corresponding to nodes of $L$, and a noise parameter $N$ bounding the fraction of training instances a rule is permitted to misclassify, and outputs a set of symbolic classification rules; this line of work has since been extended to sequential data~\cite{neuralseqclass,binaryclassrule}. A related problem arises when classification states themselves form a lattice and response distributions vary by experiment: Tatsuoka~\cite{sequenceexp} proposes a Bayesian framework for sequential classification under this assumption, building on an earlier data-analytic framework for fitting and validating latent, complex classification models~\cite{dataAnalytic} and on the theory of asymptotically optimal sequential experiment selection~\cite{sequence,sequence2}, which applies whenever the parameter space is finitely partially ordered. These lattice-based classification models are a natural fit whenever classification states are more plausibly partially than totally ordered, as in cognitive and neuropsychological assessment, educational testing, and group-testing data (Section~\ref{appsec}).

A third connection between posets and classification arises through multi-label classification and reasoning under uncertainty. Multi-label classification~\cite{multilabelclass} allows each instance to be associated with a set of class labels, but becomes computationally expensive once classes overlap in feature space or the label space grows large, since the number of admissible label combinations grows exponentially. Dempster--Shafer theory (DST)~\cite{shafer,dempsterOri} offers a generalized framework for reasoning under such uncertainty by blending probability with set-theoretic logic, but its computational cost likewise grows sharply with the number of events under consideration. Denoeux~\cite{dempster} shows that when the frame of discernment carries a lattice structure, the set of events under consideration can be restricted to intervals of that lattice, making DST computationally tractable for demanding tasks such as multi-label classification and the ensemble-clustering methods discussed in Section~\ref{clustersec}.

\subsection{Deep Learning}

This subsection reviews three interconnected uses of posets in deep learning: unsupervised representation learning, attention-based architectures, and topological deep learning, extending the poset-pooling, safety-layer, and generative (reinforcement-learning) architectures already introduced in Sections~\ref{mlnotes} and~\ref{safeRL}. Deep unsupervised learning~\cite{deepunsupervisedlearn} is attractive precisely because unlabelled data is far cheaper to obtain than labelled data, which matters for tasks such as visual similarity learning~\cite{surveyDUNN} that would otherwise require millions of labelled training pairs when addressed with standard convolutional neural networks (CNNs)~\cite{ccsilearn}. Bautista et al.~\cite{silarlearn} propose an unsupervised alternative that frames visual similarity learning as a combination of surrogate (artificially constructed) classification tasks and a partial ordering over samples. This combination of classification with poset-structured ordering directly addresses three known limitations of purely surrogate-class approaches: many training samples cannot be confidently assigned to any surrogate class because their mutual similarity is not easily established; joint optimization across surrogate tasks is hindered by mutually conflicting relationships once transitivity fails to hold; and obtaining sufficiently many labelled samples for training remains costly.

Attention-based architectures constitute a second point of contact between deep learning and order theory. Neural attention models~\cite{visualattentionmodel} restrict computation to informative regions of an image rather than processing it uniformly, originally motivated by both computational efficiency and an analogy to human visual attention, and have since become central to a wide range of vision and language tasks~\cite{anmsurvey,surveydeepattentionNN}. Hajij et al.~\cite{HOA} generalize this idea with higher-order attention networks (HOANs), defined over a \textit{combinatorial complex} (CC) -- a structure that unifies simplicial and cell complexes with hypergraphs, and that is itself a poset ordered by set inclusion. They show that any combinatorial complex reduces to a Hasse graph, which both characterizes the computational structure of HOANs in purely graph-theoretic terms and supplies a natural notion of equivariance for such networks.

This Hasse-graph reduction subsequently underpinned the broader development of topological deep learning (TDL)~\cite{ziaTDA}, a framework that combines topological data analysis (TDA) with deep learning by using topological features to inform model design. TDL traces its origins to topological signal processing (TSP)~\cite{TSP}, which first demonstrated the value of modeling higher-order, multi-way relationships among data points that pairwise graph structures cannot capture. Hajij et al.~\cite{TDL} formalize this direction with combinatorial complex neural networks (CCNNs), an abstract architecture class that generalizes convolutional and attention-based networks and whose neighborhood structure is fully determined by the incidence, adjacency, and coadjacency matrices of the underlying combinatorial complex. Because every CC reduces to a Hasse graph describing the poset structure between its cells, Hajij et al.~\cite{TDL} show that (i) any CCNN-based computation can be realized as a message-passing scheme over a subgraph of the augmented Hasse graph of the CC, and (ii) the resulting tensor-diagram representation of a CCNN is likewise fully realizable on augmented Hasse graphs -- reinforcing, at the architectural level, the connection between posets and multiparameter topological structure that we revisit in Section~\ref{futuredir}.

\subsection{Semi-Supervised Learning}

Semi-supervised learning~\cite{surveysemisuperv} bridges supervised and unsupervised paradigms by training on a combination of labelled and unlabelled data, and is particularly valuable when labelled data is scarce or expensive to obtain. Graph-based semi-supervised methods~\cite{graphsemilearn} have proven effective across many domains due to their structural flexibility and scalability, but they are typically restricted to ordinary graphs, in which each edge connects exactly two vertices. Hypergraphs generalize this by allowing an edge to join an arbitrary number of vertices, yet conventional hypergraph representations still treat each hyperedge as an unordered set, discarding any ordering relationship among its member vertices -- relationships that are present in much real-world relational data. To address this gap, Feng et al.~\cite{hypergraph} introduce the Partial-Order Hypergraph, a data structure that explicitly injects partial-ordering relations among vertices into a hyperedge, together with a regularization-based learning theory that generalizes conventional hypergraph learning by incorporating logical rules encoding these partial-order relations. Feng et al. further apply a graph convolutional network (GCN)~\cite{gcn} in a semi-supervised setting over the resulting partial-order hypergraphs, demonstrating that the order-aware hypergraph construction is compatible with standard graph-neural-network training pipelines.

\subsection{Time Series Modeling}

Many machine learning and statistical models exist for modeling sequential data~\cite{mlseqdata}, one important source of which is a series of discrete events occurring over time -- as arises, for instance, in web browsing, e-commerce, and process monitoring. A central problem in mining such event sequences is recovering an overview of the ordering relationships they encode. Mannila and Meek~\cite{seqdata} address this by treating a partial order as a generative model for event sequences, describing an observed set of sequences via a mixture of partial-order models; they show that the likelihood of a given partial order is inversely proportional to the number of total orders (linear extensions) compatible with it, and restrict the computation of this quantity to series-parallel posets, for which it can be evaluated efficiently. A greedy search algorithm over the space of partial orders on the set of possible events is then used for learning, and is able to test compatibility between a candidate partial order and an observed sequence in linear time. Nicholls et al.~\cite{bayesianpartialorder} apply a complementary Bayesian approach, inferring partial orders from random linear extensions of time-series data describing social hierarchy across the eleventh and twelfth centuries. Their model represents actors listed in order of precedence as realizations of a queue whose position is constrained by an underlying social-hierarchy partial order, providing a rank-order time-series framework for modeling how that hierarchy evolves over the observed period.

\subsection{Learning to rank} Learning to rank  (LTR)\cite{caoLTR,classifrules} describes a class of algorithmic techniques that apply supervised machine learning to solve ranking problems from a  wide range of domains such as information retrieval, recommendation systems,  search engine optimization etc.  It arises from the need to obtain prediction results based  on the best order of their relevance to a machine learning classification problem. Several methods have been proposed for LTR and they can be categorized or grouped  into three main approaches: pointwise, pairwise , listwise.  Pointwise approach involves scoring items independently and then ranking them based on their scores. In contrast the pairwise or listwise ranking methods,  consider the relative positions of items in pairs or lists respectively. The goal for the ranker is to minimize the number of inversions in ranking i.e. cases in which the pair of results are in the wrong order relative to the ground truth. In particular, the listwise learning approach addresses the ranking problem, by  analyzing ranked lists of objects  as input instances and then trains a ranking function through the minimization of a listwise loss function defined on the predicted list and the ground truth list. 
In  \cite{predrankABs}, it is noted that not all rankings can follow a strict total order in the context of learning to rank. As such a  proposed method of  relaxing the conventional setting such that predictions are given in terms of partial instead of total orders is outlined. The key idea is that if a model is  uncertain about the relative order of two alternatives, in which case it is unable to clearly determine whether the former should precede the latter  and vice-versa, it may decline or pospone the making of a conclusive decision  and instead declare such pair of alternatives as incomparable.
In general,  learning to rank problems as presented in \cite{predrankABs2} can be formulated as follows :
Given a set of training instances $\{x_1,...,x_n\}\subseteq \mathcal{X}$ and a set of labels $\mathcal{Y}=\{y_1,...,y_k\}$ endowed with an order $y_1<y_2<,...,<y_k$ such that for each training instance $x_l$ it can be associated a label $y_l.$ The problem is to determine a ranking function that orders a new set of instances $\{x^{\prime}_j\}^{t}_{j=1}$ according to their (unknown) preference degrees. The performance meaures for this are: AUC($k = 2$, for the bipartite ranking) and C-index in the polytomous case ($k > 2$, for  $k$-partite ranking). Since the ranker has the ability to reject predictions, there is a trade-off between correctness and completeness. Correctness is measured by gamma rank correlation\cite{gammarank} is a measure of rank correlation, i.e., the similarity of the orderings of the data when ranked by each of the quantities.  Completeness measure penalizes the abstention from comparisons that should actually be made. Furthermore,  a  preference relation $P:A\times A\rightarrow \left[0,1\right]$ provides a
measure of support for the pairwise preference $a\succ b$ with $P(a,b)=1-P(b,a)$ for all 
$a,b\in A,$ which can be considered an application of a generic approach\cite{predrankABs2} to transform every  ranker into a partial ranker via ensembling such that:

\begin{enumerate}

\item for any ranker $L$, train $k$ ranking models  $M_1 … M_k$  by resampling from the original data set, i.e., by $k$  bootstrap samples. By
querying these models, $k$ rankings  $\succ_1...\succ_k$ are generated;

\item for each pair of alternatives $a$ and $b$,  the degree of
preference is defined as: $P(a,b)=\frac{1}{k}\lvert\{i\mid a\succ_i b\}  \rvert.$

\end{enumerate}

\subsection{Metrics and Distances over Posets}\label{metricsub}

A prerequisite for applying nearest-neighbor methods, kernel machines, or clustering algorithms (Section~\ref{clustersec}) to poset-valued data is a well-behaved notion of distance between posets, or between elements of a poset. Taeb, Guo, and Henckel~\cite{modelgraphdist} address the specific case of distances \textit{between graphs} that represent statistical models, motivated by the observation that generic graph distances, such as the structural Hamming distance, ignore the structure of the underlying model space. They organize the graphs of interest -- probabilistic undirected graphs, causal directed acyclic graphs, and several partially directed generalizations -- into a poset ordered by model inclusion, which induces a neighborhood structure, and define the model-oriented distance as the length of a shortest path through this neighborhood structure; the resulting metric is shown to behave more consistently than existing alternatives when used to evaluate graph-structure-learning estimators. Olave~\cite{posetdistfam} takes a more purely order-theoretic approach, introducing a family of extended metrics defined directly on path-connected and fence-connected posets without requiring any additional valuation structure; these metrics arise as a shortest-path distance that accounts for both path length and the number of order-direction alternations along the path, converge to a shortest-fence metric on discrete posets, characterize most discrete path-connected posets up to isomorphism, and coincide with interleaving distances when the poset is viewed as a thin category -- connecting this line of work to the poset-cocalculus and multiparameter-persistence material discussed in Section~\ref{futuredir}. Together, these two contributions suggest that poset-valued distance learning is beginning to mature from bespoke, domain-specific rank-correlation measures (Section~\ref{multidim}) toward general-purpose metrics with formal characterization results, though, as discussed in Section~\ref{futuredir}, their computational scalability to large posets remains to be systematically evaluated.

A closely related, and in a precise sense prerequisite, problem is \textit{poset isomorphism testing}: any distance measure that assigns a poset to itself distance zero only up to relabeling requires an efficient way to decide whether two poset matrices (Section~\ref{basics}) represent the same underlying order relation. Because a poset on $n$ elements admits up to $n!$ distinct labeled matrix representations, naively comparing poset-valued data at scale -- for example, deduplicating a large collection of posets before clustering, or indexing them for nearest-neighbor lookup -- inherits the same factorial search space as the general graph isomorphism problem, of which poset isomorphism is a structured instance. Mwafise~\cite{posetisodecomp} addresses this directly with a tiered hierarchical matrix-decomposition framework that maps a poset matrix to a Hierarchical Poset Matrix Tree (HPMT) by recursively stripping universal bounds and partitioning disconnected cores, reducing isomorphism testing to a sequence of structured tree comparisons rather than an explicit search over permutations; a first tier resolves the common case of matrices with nontrivial universal bounds in empirically near-quadratic time (a $150\times$ speedup over the VF2 baseline at $n=100$, with VF2 timing out entirely by $n=500$), a second tier handles reducible posets via direct-sum decomposition, and a third tier -- required for highly symmetric, non-reducible structures such as crown posets and $N$-dimensional hypercube (Boolean lattice) posets, where degree-based invariants collapse -- extracts maximal disconnected principal submatrices in $O(n^4)$ time, remaining tractable (e.g., $62$ seconds for a $512$-element hypercube $Q_9$) at sizes where VF2-based backtracking times out. The framework is accompanied by formal invariance theorems establishing that the decomposition commutes with relabeling, so that isomorphic posets provably yield identical hierarchical fingerprints. For the metric-learning methods surveyed above, fast canonicalization of this kind is a natural and currently missing preprocessing step: it is what would make pairwise poset-distance computation tractable at the scale needed for the nearest-neighbor and kernel-based learning tasks motivating this subsection in the first place, a connection we return to in Section~\ref{futuredir}.

\subsection{Machine Learning Using Formal Concept Analysis}\label{fcaTT}

Formal concept analysis (FCA)~\cite{chemdualFCA,FCABook} originates from partial order and lattice theory and provides a mathematically grounded method for conceptual knowledge representation and data analysis, introduced in the seminal work of Wille~\cite{latticehiech} as an attempt to restructure classical order and lattice theory around the notion of a \textit{concept}. The basic data format in FCA~\cite{fcaontology} is a cross-table given by a triple $(O, A, I)$ called a \textit{formal context}, where $O$ is a set of formal objects, $A$ is a set of formal attributes, and $I \subseteq O \times A$ is the incidence relation between them; a pair $(X, Y)$, where $X$ is a maximal set of objects (the \textit{extent}) sharing a maximal set of common attributes $Y$ (the \textit{intent}), constitutes a \textit{formal concept}. The set of all formal concepts of a context is partially ordered by extent inclusion and forms a complete lattice, whose representability by ordered sets of meet- and join-irreducibles~\cite{fcaordreFR} is the fundamental result underlying FCA. Because a formal context is naturally represented by a binary object--attribute matrix -- rows indexed by objects, columns by attributes, with a $1$ entry wherever an object possesses an attribute~\cite{MathConcepts} -- properties of lattice theory can be applied directly to the analysis of such matrices~\cite{mlauto}. A range of discretization and Booleanization procedures further allow diverse datasets to be converted into formal contexts or concept lattices~\cite{dataformconc,FCAConvert,VFCAdata}, so that FCA can be viewed, in effect, as an unsupervised machine learning technique that takes a binary relation as input and returns the natural concepts of the data, organized as a Hasse diagram~\cite{conceptbook,fcAgarrica,fcAIganov}. This generality has driven a growing adoption of FCA across data science tasks~\cite{fcaML}, with several FCA-based methods reported to be competitive with classical machine learning approaches~\cite{fcarandom}. In an application to image classification, Khatri~\cite{fcaimage} identifies two specific advantages of FCA over convolutional neural networks: (i) FCA provides interpretability by construction, since the classification hierarchy is directly visualized as a lattice, and (ii) data can be added to or removed from the lattice without retraining the model, unlike a trained CNN -- with the resulting FCA-based classifier reported to outperform the random forest classifier, itself considered a relatively interpretable baseline. A further advantage of FCA-based classification more generally is that it makes no distributional assumptions about the underlying data~\cite{proclasmodels}, in contrast to many classical statistical classifiers.

This lattice-theoretic view of classification extends naturally to \textit{learning from positive and negative examples}, in which a learning system constructs a generalization of positive examples that excludes (does not ``cover'') negative examples~\cite{mlauto}; several FCA-based models address this problem directly~\cite{fcaML,fcaclassbiclus,mlauto}. Jabin~\cite{gafca} applies a genetic algorithm to automatically learn object-oriented hierarchies closely related to lattice-based concept hierarchies from large datasets, while Ikeda and Yamamoto~\cite{fcaclassLang} classify data by constructing a concept lattice and then selecting the formal concepts within it that are most informative for a given classification task, improving feature-selection robustness by retaining both selected and redundant concepts. FCA has also been integrated with ensemble-learning strategies -- Dagging~\cite{fcadagging,fcarandom}, Boosting~\cite{fcaboosting}, Bagging~\cite{fcabagging}, and recommendation-based multiple classifier systems~\cite{fcareccO} -- extending the ensemble-clustering connections already discussed in Section~\ref{clustersec}. Xie~\cite{Zhipeng} embeds simple base classifiers (Naive Bayes and $k$-nearest-neighbor) into each node of a concept lattice, producing composite Concept Lattice Naive Bayes (CLNB) and Concept Lattice Nearest Neighbor (CLNN) classifiers that outperform their respective base classifiers -- and, in the case of CLNB, several state-of-the-art classifiers -- across 26 benchmark datasets.

FCA's inherently visual, lattice-based structure also makes it a natural tool for the explainability problem introduced in Section~\ref{xaiSec}: because many machine-learning-based AI systems are designed as black boxes, achieving interpretability in the face of increasingly complex model architectures remains a central challenge. Sangroya et al.~\cite{fcadlexplain} propose a general concept-lattice-based framework for explaining deep learning outcomes, in which a model prediction and a domain ontology are combined to identify the explanation that points a user to the most salient feature set underlying that prediction.

Finally, FCA supports the automated construction of domain-specific ontologies directly from textual descriptions of domain entities~\cite{fcAOntoInduct,fCAAutoOnt,fcaonto5}. \textit{Ontology learning}~\cite{ontolearnD} is the broader process of automatically extracting knowledge structures -- typically annotated taxonomies, concept hierarchies, or domain ontologies -- from unstructured or semi-structured sources such as text, speech, images, or sensor measurements; ontologies of this kind are a key building block of the semantic web, capturing domain knowledge in a machine-processable form. \textit{Relational Concept Analysis} (RCA)~\cite{datasetrelation} extends the FCA framework to multi-relational datasets by generating a family of concept lattices, one per object category, and both FCA and RCA underpin a range of approaches to ontology learning and extraction~\cite{fcaonto1,fcaonto2,fcaonto3,fcaonto4,fcalconcept}.

\section{Clustering Partially Ordered Data}\label{clustersec}

Cluster analysis is a multivariate technique for identifying structural patterns in complex data by grouping similar objects together, typically by evaluating similarity or dissimilarity across a set of shared attributes. Under the unsupervised-learning branch of the taxonomy in Table~\ref{tab:taxonomy}, several distinct lines of work have brought poset and lattice structure to bear on this problem: an \emph{ordinal} model of clustering built directly from order theory, \emph{hierarchical} clustering methods that operate on the Hasse graph of a poset, \emph{ontology}-driven clustering based on the subset relation, and \emph{conceptual} clustering derived from formal concept analysis (Section~\ref{fcaTT}).

The ordinal model for clustering with posets, introduced by Janowitz~\cite{clusteroder}, shows that characterizing flat cluster methods~\cite{semiflatcluster} reduces to a universal mapping problem in the theory of partially ordered sets, with generalized notions of adjoints of order-preserving mappings between posets recurring as a persistent underlying theme~\cite{clusterreport}. Building on this, Janowitz~\cite{clusterreport} develops a clustering scheme based directly on dissimilarities measured over posets, in keeping with the broader observation that most classificatory clustering methods operate on a dissimilarity coefficient defined over a set of objects~\cite{pyramidORD}.

Cluster analysis is traditionally divided into hierarchical and non-hierarchical methods: hierarchical clustering produces a nested sequence of partitions, while non-hierarchical clustering produces a single partition. Hierarchical clustering further splits into \emph{divisive} methods, which proceed top-down by successively dividing a single cluster as inter-object distance increases, and \emph{agglomerative} methods, which proceed bottom-up by successively merging clusters as inter-object distance decreases; agglomerative clustering is commonly further categorized into single linkage (minimum distance between clusters) and complete linkage (maximum distance between clusters)~\cite{singlecomplete}. Several authors have combined these classical hierarchical schemes with poset structure. Sabara et al.~\cite{clusteragglo} apply both single- and complete-linkage agglomerative clustering directly to the Hasse graph of a poset in a multidimensional data-analysis setting (Section~\ref{multidim}), while Wu et al.~\cite{clusterhres} and Kardaetz et al.~\cite{envshallowlake} likewise combine poset methodology with hierarchical clustering from a multidimensional perspective; Wu et al.~\cite{clusterhres} in particular develop a hierarchical stratification method, interpretable via the partial-order Hasse graph, that divides multi-dimensional indicators into layers with distinct properties. Janowitz~\cite{clusterrelate} provides a comprehensive review of this broader area, including the use of lattices to generalize tree-based clustering.

A related strand of work uses posets to formalize the relationship between clustering and \emph{ontologies}, which represent data as hierarchies of possibly overlapping classes and are thus closely related to clustering hierarchies in their own right. Liu et al.~\cite{clusterpair} show that modeling ontologies as posets over the subset relation allows classical dissimilarity-matrix-based clustering algorithms to incorporate all available ontological information without loss, and use this observation as the basis for a clustering algorithm that produces a partially ordered set of clusters directly from a dissimilarity matrix. Because a dissimilarity matrix has size $\mathcal{O}(N^2)$ in the number of objects $N$, independent of the dimensionality of the objects themselves, this poset-of-clusters construction sidesteps some of the difficulties otherwise associated with clustering high-dimensional data.

A fourth strand connects clustering directly to formal concept analysis (FCA), which -- beyond its role in classification (Section~\ref{fcaTT}) -- is itself a form of \emph{conceptual clustering}~\cite{clusterrelate}, a branch of machine learning introduced by Michalski~\cite{mikhalskiconcept} that groups unlabelled objects into classes. Carpineto~\cite{conceptret} identifies three defining features of conceptual clustering methods: (i) each output class is characterized both extensionally, by the objects it covers, and intensionally, by the concept it represents; (ii) output classes are arranged into a hierarchy ordered by generality; and (iii) class formation proceeds incrementally, so that processing the $n$th object does not require reprocessing the preceding $n-1$ objects. The theory of concept (Galois) lattices offers a natural formalization of this process: the Galois lattice generated from a binary relation~\cite{incgaloisFO} is itself a concept hierarchy, and Carpineto's GALOIS algorithm~\cite{galoiscluster} computes the concept lattice for a given set of objects with an update-time complexity ranging from $O(n)$ to $O(n^2)$ in the number of concepts $n$, and has been shown useful for both class discovery and class prediction. Related conceptual-clustering perspectives include Fisher's treatment of conceptual clustering as an extension of numerical taxonomy~\cite{fisherConcept}; Restrepo et al.'s use of hierarchical cluster analysis (HCA) to reduce the number of elements in a poset to a set of representatives, thereby improving the interpretability of the resulting Hasse diagrams~\cite{HCAHDT}; and Zhang et al.'s distance-function-based hierarchical conceptual clustering, which mitigates the NP-hardness of exhaustively determining all concepts of a formal context by subsetting the feature set used to define the clusters~\cite{computeconc}. Markov~\cite{markovHcluster} instead induces a lattice structure over the clusters via a generalization operator, maximizing overall clustering quality by evaluating the hierarchy level by level in a bottom-up fashion, while Yoneda et al.~\cite{grapFCA} use the algebraic closedness property of FCA to learn a graph-structured representation of multivariate data in which each node is a cluster and each edge encodes a subset--superset relationship between clusters; related graph-based hierarchical conceptual clustering methods applicable to partially ordered data are studied in~\cite{joynyergraphClu,joynyergraphClu2}.

Posets also arise naturally within a single clustering run, independent of any external ontology or concept lattice, through the space of \emph{partitions} itself: the frame of discernment in clustering is the set of all partitions of a finite set $E$, denoted $\mathcal{P}(E)$, and this set carries a natural partial order in which a partition $p$ is \textit{finer} than a partition $p'$ (written $p \preccurlyeq p'$) whenever the clusters of $p$ are obtained by splitting those of $p'$, yielding the poset $(\mathcal{P}(E), \preccurlyeq)$. This structure underlies \textit{ensemble clustering}, which combines the outputs of several clusterers into a single clustering structure. Using evidential reasoning grounded in Dempster--Shafer theory~\cite{dempster} -- already introduced in the classification context of Section~\ref{mlnotes} -- one can assume the existence of a ``true'' partition $p^*$, of which each clusterer supplies partial evidence; the evidence from multiple clusterers can then be combined within the partition poset to draw plausible, well-supported conclusions about $p^*$ even in situations where no single clusterer's output is decisive.

\section{Multidimensional Data Analysis}\label{multidim}

Multidimensional data analysis (MDA) originated with the development of relational databases and On-Line Analytical Processing (OLAP) by Codd in 1993~\cite{mdaHtech,dimensionsDEF,defmda,mdaorigins}, whose central idea -- organizing data along \textit{dimensions} -- provides a flexible way to interpret the same dataset from several angles at once. In statistics, econometrics, and related fields, MDA accordingly groups data into \textit{dimensions} (the axes along which analysis is carried out) and \textit{measurements} (the values associated with different dimensional entities), with the resulting hierarchy, sequencing, and dependency relationships among dimensions typically organized into a meaningful structure~\cite{dimensionsDEF,HierarchyMDA,mdaHtech}.

Multi-indicator models~\cite{multipleindicator} -- in which two or more ``alternative'' measures are used for the same underlying concept -- are especially useful in this setting; a familiar example is measuring individual satisfaction with a service via several differently worded survey questions, and this kind of data is common in public-opinion surveys of many thousands of individuals~\cite{posetbigdata,urbandepr} and in environmental and infrastructure monitoring, where a battery of indicator values assesses several features of a site's condition~\cite{bridgeposets}. In both settings, different indicators frequently convey conflicting comparative signals about the same object. Multi-indicator systems are similarly central to modeling business processes that span multiple geographic regions, channels, and products, but quantifying such systems is difficult and time-consuming because of their inherent complexity~\cite{multenv,complexphenomena}: nominal or ordinal multivariate data is often first scaled to a quantitative form, a step that can itself introduce inconsistency~\cite{fuzzypoverty}. The conventional solution -- aggregating indicators via a weighted average or another aggregation scheme (additive~\cite{ADDmethod}, hierarchical~\cite{HHmethod}, etc.) into a single \textit{composite indicator}~\cite{compositeINDIC}, which is then used for ranking~\cite{rankindic} -- is a widely used method in the social sciences for measuring multi-dimensional phenomena such as well-being (which might combine income, employment, health, and education into one score)~\cite{summarydata}. However, the validity and robustness of composite indices constructed this way has been repeatedly challenged, largely because of the unavoidable subjectivity involved in choosing aggregation weights: aggregative methods tend to oversimplify the underlying phenomenon and can produce identical scores for genuinely different situations~\cite{whyagg,measuredev}, and ordinal data of this kind often lacks a clear ordering criterion in the first place, making individual-level comparisons ambiguous~\cite{fuzzypoverty}. A substantial body of work (e.g.,~\cite{summarydata,envshore,envranking,compindic,multenv,posetbigdata,poprank,partialeuro,multidomainsys,cantercrime,fattoresuffer,happiness}) therefore advocates replacing aggregation with partially ordered set theory, which restricts subjectivity to the choice of which properties to consider in the first place, rather than to how they are numerically combined. The underlying motivation is that classical aggregative methods can fail to express the true complexity of a phenomenon, whereas partial orders make explicit \emph{why} an object occupies a given ranking position, and how sensitive that position is to changes in the underlying indicators~\cite{multenv} -- directly complementing the explainability motivation for posets discussed in Section~\ref{xaiSec}. The principal advantages of treating multivariate data as a poset for ordering purposes are summarized below.

\begin{itemize}

\item Partial order theory yields rankings that can accommodate ties without requiring indicator weighting~\cite{failedstate}: the horizontal arrangement of objects within a Hasse diagram, organized by level, already gives a first approximation to a weak order in which tied ranks are not excluded.

\item Partial orders impose no requirement that indicators be mapped onto a single common scale.

\item Phenomena such as subjective well-being can be evaluated consistently and effectively using partial order theory, overcoming the limitations of both composite and simple counting paradigms~\cite{ordbeing}.

\item Partial order theory handles multidimensional systems of ordinal data without variable aggregation into composite indicators, so there is no need to convert ordinal scores into numerical values -- a conversion that can itself introduce inconsistency into the evaluation of the underlying phenomenon~\cite{compindic,partialeuro}.

\item No weighting of evaluation dimensions is required to account for their differing relevance~\cite{compindic}.

\item Objects are compared simultaneously across all indicators, without any need for prior aggregation~\cite{inequalitiesEU}.

\item Partial order ranking is a non-parametric method, requiring no assumptions of linearity or of any particular statistical distribution over the attributes~\cite{apor}.

\item The graphical representation of posets as Hasse diagrams requires no additional information to sort objects; because a Hasse diagram represents a well-defined mathematical structure, further conclusions can be drawn from it beyond a simple sorted order~\cite{envshore,onlinedatabase}.

\item Posets define the structure of comparabilities underlying a multi-indicator system~\cite{measuredev}, permitting mathematical analysis of the phenomenon under study and enabling a reduction in its dimensionality without loss of information, since no difference or ratio of indicator values is ever computed to derive distances or proportions.

\item Because different indicators often convey different comparative messages, there is frequently no single, unambiguous way to rank a set of objects while honoring all of the available indicator information~\cite{bridgeposets}: object $A$ may dominate object $B$ on one indicator while $B$ dominates $A$ on another. This complicates prioritization based on raw indicator scores, but the properties of posets and their associated rank matrices provide additional information useful for decision-making beyond a single overall ordering.

\item Representing a phenomenon by a Hasse diagram avoids the arbitrariness inherent in constructing a single ranking index~\cite{envshore}, and offers a holistic view of all objects to be ranked without introducing artificial ranking indices~\cite{envrankaqua}.

\end{itemize}

\noindent We now summarize how a poset structure is obtained from a subjective dataset in practice. Let $k$ denote the number of subjective indicators used to study a phenomenon of interest, and let $n$ denote the number of individuals in the study, so that $A = \{q_1, q_2, \ldots, q_k\}$ represents the set of subjective-indicator variables. Each indicator is scored on an $m$-degree ordinal scale $(y_1, y_2, \ldots, y_m)$ with $y_1 < y_2 < \cdots < y_m$, and each of the $n$ individual profiles is determined by the $m$ ordinal scores assigned across the $k$ indicators in $A$. Pairwise comparison of all $n$ individual profiles then yields a poset structure whenever the resulting comparisons are not totally ordered, which can be depicted as a Hasse graph of size $n$. For a four-degree scale, for example, the profile $(y_2, y_3, y_2, y_3)$ dominates $(y_4, y_1, y_3, y_1)$, whereas $(y_3, y_3, y_4, y_2)$ and $(y_4, y_3, y_3, y_2)$ are incomparable. Several complementary methodologies have been developed for analyzing multivariate datasets of this kind, based respectively on (i) a fuzzy set approach~\cite{fuzzydeprv,fuzzypoverty}, (ii) partial order linear extensions~\cite{compindic}, (iii) average rank~\cite{posetbigdata}, (iv) poset theory combined with the Adjusted Mazziotta--Pareto Index~\cite{multiindictaor}, (v) interpersonal comparability~\cite{interagrr}, and (vi) fuzzy first-order dominance~\cite{poprank}, summarized in turn below.

\begin{itemize}

\item The fuzzy set approach, originating in Zadeh's seminal work~\cite{fuzzysetZ}, models uncertainty over vague concepts by allowing partial set membership. Combined with posets, it has been used to statistically evaluate ordinal data on socio-economic phenomena while overcoming the limitations of classical composite-indicator aggregation~\cite{fuzzydeprv,fuzzypoverty,migrantspos}, and its main advantage is that the measurement level of the data is fully respected, avoiding any improper rescaling.

\item Fattore et al.~\cite{compindic} rank a finite collection of objects -- each represented by a suite of indicator values, i.e., a point in a multidimensional cloud -- by treating their relative positions as defining a partial order, avoiding the arbitrary assignment and aggregation of a composite numerical score; incomparable pairs are simply left incomparable. Each object's possible rank interval is then determined from the Hasse diagrams of all linear extensions compatible with the partial order. Since exhaustively enumerating linear extensions becomes computationally intractable for large datasets, Bruggemann et al.~\cite{envranking} instead propose Discrete Markov Chain Monte Carlo (MCMC) sampling of linear extensions~\cite{fastlinear}, followed by cumulative-frequency ranking based on the sampled extensions, while Lerche~\cite{rankprob} develops ranking-probability estimates from random linear extensions -- both approaches make it practical to predict ranking probabilities and average rank for posets too large to enumerate exhaustively.

\item Caperna and Boccuzzo~\cite{posetbigdata} address big, complex datasets in which some attributes require measuring complex concepts on ordinal or dichotomous scales, by sampling units from the population using a simple criterion and then applying the central limit theorem to compare group-level results via standard statistical tests on the means; their Height of Groups by Sampling method then compares average rank across groups defined by one or more socio-demographic variables.

\item Patil et al.~\cite{multivariate} review procedures for extracting ordering properties embodied in multivariate datasets, applicable to configuring sets of indicators more generally.

\item Alaimo et al.~\cite{multiindictaor} synthesize the evolution of multi-indicator systems over time by combining partial order theory with the Adjusted Mazziotta--Pareto Index, applied to one of the fifteen Sustainable Development Goals; posets define the comparability structure underlying the multi-indicator system, scores are then evaluated to reduce the phenomenon's dimensionality, and temporal posets are obtained by merging posets across time periods, with an embedded five-level scale (minimum, maximum, and the first, second, and third quartiles of the indicator values) used to further refine the measurement.

\item Sen~\cite{interagrr} develops a framework for interpersonal comparability in aggregating individual welfare measures, considering an aggregation relation between social states that reduces to the Pareto quasi-ordering under noncomparability and becomes a complete ordering under unit or full comparability.

\item Fattore and Maggino~\cite{ordbeing} outline a comprehensive evaluation procedure for multidimensional ordinal settings, comprising identification of evaluation dimensions, assignment of attribute relevance, threshold selection, and computation of evaluation scores at both the individual and population level.

\item Fattore~\cite{ordinaldeprive} proposes a partial-order-based procedure for assessing multidimensional deprivation with ordinal data, focusing on achievement profiles and their comparabilities and incomparabilities; the resulting synthetic indicators account for both the vagueness and the intensity of multidimensional deprivation, incorporate attribute-importance information, and handle missing data.

\item Fattore~\cite{poprank} introduces fuzzy first-order dominance (F-FOD), which combines partially ordered set theory with fuzzy relational calculus to overcome limitations of earlier first-order-dominance algorithms, producing full pairwise comparison matrices from which partial orderings and rankings of statistical units can be derived.

\item Bruggemann et al.~\cite{chemrank} propose an average-rank method based on the structure of the Hasse diagram associated with the order relations encoded in the data matrix.

\end{itemize}

\section{Exploratory and Descriptive Data Analysis}\label{exploratory}\label{descriptive}\label{descsection}

Exploratory and descriptive data analysis are closely related stages of the data-analysis pipeline, and posets have been used extensively in both: exploratory data analysis (EDA) is the preliminary investigation of a dataset to uncover patterns, detect anomalies, test hypotheses, and check assumptions, while descriptive analysis summarizes a dataset's primary properties and uses historical trends and relationships between variables to inform further analysis and decision-making. We treat both together in this section, since the poset-based techniques that support them -- handling noise and missing data on one hand, and visualizing the order structure of a dataset on the other -- form a continuous methodological pipeline.

\subsection{Noise, Trend, and Missing-Data Handling}

A major challenge in the exploratory phase is identifying noisy data, characterized by a degree of variability that the rest of the dataset cannot explain, since noise can alter the ordering relations between objects and thus introduce genuine data uncertainty into a partial ordering~\cite{rankparameter}. Bruggemann and Carlsen~\cite{noisyposet} address this by constructing a probability scheme that specifies an explicit noise model, from which the distribution of noisy values can be derived analytically; this supports \textit{a priori} estimation of how a given noise level affects the order relation between any pair of objects, so that the expected magnitude of noise-induced perturbation can be quantified in advance. Carlsen and Bruggemann~\cite{infnoise} extend this analysis with a fuzzy approach in which the partially ordered set itself is treated as a function of the noise level, making it possible to identify the range of noise over which the original partial order remains stable.

Trend analysis -- tracking how a phenomenon evolves over time -- is a natural fit for poset-based exploratory analysis. Alaimo et al.~\cite{multiindictaor} show that analyzing the evolution of Hasse diagrams over time yields a preliminary depiction of a phenomenon's temporal trend, though missing data remains a well-known obstacle to nonlinear trend analysis more generally~\cite{trendmiss}. Fattore~\cite{ordinaldeprive} addresses this obstacle directly with a poset methodology for handling missing values, applied to a case study of deprivation data; the underlying idea is that pairwise comparison of poset elements is inherently robust to missing information, since the absence of sufficient information to compare one object to a particular other object does not preclude using the remaining, available comparisons elsewhere in the dataset~\cite{summarydata}. In the specific context of formal concept analysis (Section~\ref{fcaTT}), this same robustness -- together with FCA's human-centered, visually grounded representation of concepts -- makes it particularly well suited to exploratory data analysis~\cite{formalexplore,fcAIganov}.

\subsection{Visualization via the Hasse Diagram Technique}

Turning from exploration to description, poset theory has been used extensively for data summarization and visualization, most centrally through the Hasse diagram, which conveys a considerable amount of information about a dataset's partial order structure in a single graphical representation. Hasse diagrams have accordingly been used for visualization tasks spanning causality~\cite{hassDis}, socioeconomic analysis~\cite{visualsocio}, linear models and analysis of variance~\cite{hassANOVA}, experimental design~\cite{hassExp}, learning analytics~\cite{hassLearntics}, image analytics~\cite{HassVis}, medical data analysis~\cite{diazwhasse}, molecular structure prediction~\cite{reactiondiag,molepred}, spatial analysis~\cite{spatial,envgraph}, multidimensional analysis~\cite{humandev,decisionsupprt,multenv}, biomonitoring~\cite{biomonitoring}, multivariate analysis~\cite{rankinadequate}, sustainability analysis~\cite{foodwate}, multi-criteria decision analysis~\cite{stakeholder}, environmental data analysis~\cite{envdataH}, and decision support~\cite{decisionsupportposet,bridgeposets}, among other domains reviewed further in Section~\ref{appsec}. The overarching goal of such a visualization tool, for multidimensional and partially ordered datasets alike, is to represent the data structure directly, reducing its complexity while retaining its essential patterns~\cite{visualsocio}.

Among the formal methods built around the Hasse diagram, the most established is the Hasse Diagram Technique (HDT), introduced by Bruggemann and Voigt~\cite{hassetechchem} as an application of partial order theory to a data matrix, used to analyze the structure of multivariate datasets whose objects are characterized by multiple attributes (indicators). The HDT sorting rule~\cite{chemrank} can be stated concisely: given a dataset of objects to be sorted with respect to some criterion, let $m(i, x)$ denote the value of the $i$th attribute of object $x$, for $i = 1, \ldots, n$; then $x \geq y$ (i.e., $x$ is evaluated as at least as good as $y$) if and only if $m(i, x) \geq m(i, y)$ for every $i = 1, \ldots, n$. The HDT is especially useful for multi-criteria ranking problems in which each object must be assessed relative to all others~\cite{onlinedatabase}, for several reasons: it makes the ranking process transparent; it requires no normative constraints on how the ranking is performed; it enables extraction of structural information from, and visualization of, the resulting Hasse diagrams~\cite{envshore}; and it supports visualization of sensitivity-analysis parameters~\cite{chemenvtest}.

Bruggemann et al.~\cite{chemrank} apply the HDT to sort chemicals by potential environmental hazard, establishing dominance of one chemical over another whenever every attribute simultaneously supports that dominance. One limitation of the basic HDT is its lack of weighted aggregating functions, which would otherwise allow conflicting descriptor values to be resolved; Simon et al.~\cite{decisionsup} address this with METEOR (METhod of Evaluation by ORder), an extension of the HDT procedure that, unlike the base method, combines transparent decision support with the ability to incorporate stakeholder preferences~\cite{decisionsup2}, resolves the problem of obtaining a single top-ranked object, and avoids costly trial-and-error selection of descriptor priorities by directly computing the probability of a given linear order under a specified descriptor prioritization~\cite{chemrankpatterns}. METEOR has subsequently been applied to the computational evaluation of chemicals and environmental hazards~\cite{envpollutionMETEOR,rankenvsoft}. In a related vein, Halfon et al.~\cite{rankhazard} propose a vectorial ranking procedure for partial ordering that applies broadly across problems in environmental toxicology, motivated by the poset's ability to represent both comparable and genuinely incomparable chemicals with respect to environmental hazard; the resulting rankings are visualized via computer-generated Hasse diagrams applicable to datasets of any size, with the chain and antichain structure of the diagram corresponding respectively to its vertical and horizontal components. Bruggemann~\cite{rankpolychlorinate} shows that tripartite graphs are useful for interpreting complex datasets of this kind by clarifying which indicators drive the incomparabilities reflected in a Hasse diagram's horizontal components, and Carlsen et al.~\cite{analyticperform} demonstrate, in an application to analytic-performance evaluation, that such incomparabilities -- an inherent feature of partial order ranking rather than a defect -- can themselves reveal important characteristics of the data, with summary statistics (absolute $z$-score, absolute skewness, standard deviation) visualized directly on the Hasse diagram to provide a holistic view of the results. A related graphical technique, POSAC~\cite{crime2,cantercrime}, maps the rows of a data matrix (e.g., geographic regions) into a two-dimensional space that maximizes preservation of their partial order, placing similar objects in close proximity; POSAC has proven useful for tracking the evolution of crime analytics across large geographical areas over time.

Formal concept analysis offers a complementary route to visualization: its concept lattice, together with the associated Hasse graph, provides a conceptual framework for structuring, analyzing, and visualizing data in a more concise and comprehensible form~\cite{ontsim}, distinguished from a plain Hasse diagram by its ``symmetric'' treatment of objects and properties rather than objects alone~\cite{rankpolychlorinate}. Large datasets nonetheless remain computationally challenging for FCA, since the number of formal concepts derived from a dataset is the key factor determining whether the resulting concept lattice is usable for visualization. Andrews and Orphanides~\cite{dataformconc} show that interpretable results can still be obtained from data sources that would otherwise be intractable to visualize, by first focusing on the information of interest and then reducing noise in the formal context, thereby revealing readable lattices that faithfully represent the conceptual structure of large datasets.

A general limitation of the Hasse diagram as a representation, independent of FCA, is that it becomes difficult to interpret as the number of vertices and edges grows, and it provides no metric information even when such information is available in the original data. Cluster analysis (Section~\ref{clustersec}) addresses the complexity problem but is not designed to preserve comparability and incomparability information. Bruggemann and Carlsen~\cite{clusterincomp} combine the two, producing a visual output that lets end users jointly grasp both the partial order and the metric structure of the data via a three-step process detailed by Fattore et al.~\cite{visualsocio}: (i) reduce dataset complexity through clustering based on a self-organizing map (SOM); (ii) build a classical Hasse diagram over the resulting population of clusters, associated with the SOM weight vectors; and (iii) visually annotate the diagram with information on statistical units, clusters, and covariate values. Alternatively, combining cluster analysis and principal component analysis~\cite{envshore} with the HDT can help obtain a statistically relevant data representation while avoiding insignificant numerical differences between attributes that would otherwise generate spurious comparabilities and incomparabilities, and correspondingly overcomplicated Hasse diagrams.

\section{Applications}\label{appsec}

This section surveys the datasets, software, and algorithmic resources that support the practical analysis of partially ordered data. Since most machine-learning applications of posets have already been discussed in Section~\ref{mlnotes}, the focus here is on data-analysis applications across application domains, following the \textit{data modality} and \textit{task} axes of the taxonomy in Table~\ref{tab:taxonomy}. Partially ordered data and structures are ubiquitous, and a significant share of their applications concern data-driven decisions in sustainable development~\cite{posetanalSD,measuredev,govt,multiindictaor}, where the central use case is analyzing complex, multidimensional systems of ordinal data for multi-criteria decision-making in the socio-economic and environmental sciences~\cite{appliedscience}. Partial order methodology is sometimes used as an interim step before other analytical tools are applied~\cite{inequalitiesEU}, but posets and other lattice-theoretic methods are more generally useful throughout comparative evaluation processes~\cite{onlinedatabase}: binary comparisons based on a given criterion are naturally order-theoretic, so poset methodology is often better suited to comparative evaluation than purely statistical tools. Because posets can be visualized directly, additional structural results can be derived from the underlying mathematical theory of relations between objects -- order relations support identifying and evaluating the most relevant objects in a study, and deriving the relative importance of criteria used in an ordinal ranking process -- and the Hasse Diagram Technique (Section~\ref{descsection}) offers a particularly powerful tool for such comparative evaluations~\cite{envgraph}. Newlin and Patil~\cite{bridgeposets} outline a procedure for identifying the minimal and maximal elements of a poset from its Hasse graph in complex scenarios, which is of central importance whenever a priority-setting procedure must be carried out. More generally, partially ordered sets encode a substantial amount of information about the degree of dominance among their elements, and the tools reviewed throughout this survey exist precisely to extract that information and convert it into rankings for the wide range of data-analysis purposes surveyed below.

\subsection{Environmental Data Analysis}

The concept of partially ordered sets and their visualisation by Hasse diagrams turns out to be very useful in many applications of environmental pollution data studies where evaluative considerations and assessments are important  \cite{rankhazard,FATTOREENV, envnerveagents,envhazard,  envpollution, conceptAnal, envchempesticide,envpollutemon,envpahrm,envSWsoils,firstrank,posetvscluster}. Environmental data have been extensively studied by
Hasse Diagram Technique in  \cite{envrankontario,envgraph,envshore,envchem} with the  goal to identify significant relationships among sediment samples (objects) and degradation indices (attributes). Further investigations using formal concept analysis\cite{chemdualFCA}   with the aim to highlight  interaction among hygienic compounds and a synergism between  toxicity tests applied to the sediment samples from surface water sources. Hasse diagram technique\cite{envgraph} supports the visualization of more than two-dimensional problems to identify
pollution patterns by characterizing geographical regions using their chemical pollution levels, and then suggesting priority regions for further examination. The pollution pattern which determines the location of the regions within a Hasse diagram is useful in the search for remediation strategies. In an alternative approach outlined by Restrepo \& Bruggemann\cite{HCAHDT}, the pollution pattern from regional studies was examined using a methodology that combined hierarchical clustering analysis(HCA) and Hasse Diagram Technique. The HCA was used for classification of the objects in the set in order to find similarity classes resulting to reduced set of object repreentatives, thereby making the Hasse diagrams for analysis depicting the network structure more easily understandable.
Posets can also be very useful when ranking environmental  chemicals by  multicriteria analysis \cite{rankenvsoft,delosrank,annrankchem}. Posets in this case  provides a solid formal framework for the ranking of objects without assigning a common scale or weights to the criteria. In the prediction of toxicity levels of chlorobenzene in  environmental chemical polychlorinated hydrocarbons, a  scheme is developed by Ivanciuc et al.\cite{quantposet} based on poset theory and embedded within an overall reaction network.

\subsection{Socio-Economic Data Analysis}

Partially ordered data are prevalent in many branches of the social and behavioral sciences. This has been driven by the fact that partial order theory employed in the applied sciences overcomes the intrinsic disadvantage hidden in aggregation if a multiple attribute system is available\cite{partialeuro}. A typical example is in response categories: ``Agree", ``Neutral", ``Disagree", and ``Don't Know", of which the first three can be ordered and the last forms a category of its own.  This type of data can be derived from a  wide range of applications covering different topics such as  multidimensional poverty, economic development, inequality measurement etc. Partially ordered set theory has been shown to be particularly suitable to address multi-criteria decision problems  since  it shows where multidimensional indicator data values are expressing a conflict which can then be identified and resolved by appropriate corrective action \cite{decisionsupprt}. The use of  posets with quantitative data reduces the set of operations and choices to be made in order to synthesize indicators (normalization, aggregation), even if they are the natural representation of multidimensional ordinal data \cite{ordinaldeprive, fattoreWELL}. Using this approach, it is possible not only to investigate the nature of the phenomenon in more detail, but also to help policy-makers in their assessments. Infact, it can be considered an effective policy tool aimed at promoting the identification of disadvantaged and under-developed profiles and contexts  \cite{measuredev}. The application of the poset-based method provides an  understanding of the  complexity of the evolution of a social phenomena in terms of temporal trends and comparisons between regional structures in a countrywide basis using a multidimensional data source for the process \cite{multiindictaor,posetbigdata}. Incomparability captures the existence of intrinsically different forms of it, thereby providing a more realistic picture of the phenomenon under investigation. In this respect, the impracticability to compare the profiles of different cities, due to the existence of dimensions where they perform in conflicting ways, reveals the irreducible complexity of sustainability.  In census data on deprivation within regions of a country, poset theory has been applied because they can  account for incomparabilities which are at the basis of deprivation complexity\cite{migrantspos,fuzzypoverty,fuzzydeprv}.  Partial order theory has been extensively applied to the construction of synthetic indicators in various socio-economic contexts \cite{multiindictaor,measuredev,anonifuzzy,whyagg,badingerfiscal,gendergap,rankindic,carlsenfragile,eukeys,dellasocio,fattoresuffer,fattoreWELL,FATTOREFIN,Rimodifragile,RIMOLDIMUN}.
In the study of crime analytics\cite{cantercrime}, POSAC(Partially Ordered Scalogram Analysis with Coordinates) is a graphical representation, providing dimensional reduction procedures that preserve partial order. It has been used to study the structure of criminal networks in terms of their longevity,  biographical data the, network size and so forth. Levy \cite{crime} performs  data analysis by employing the technique of partial order analysis of crime indicators characterizing cities. The results were visualized with a scalogram based on the method of Partial Order Structuple Analysis which was developed for non-metric data analysis. Furthermore, it is noted that the approach described in Levy\cite{crime}, can be useful in  a broad range of problems on the stratification of cities, individuals, as well as several varieties of social indicators for classification.

 \subsection{Neurocognition Modelling}

Finite partially ordered sets  are natural models for cognition as it is reasonable to assume that some cognitive states have higher levels of functionality
than others\cite{dataAnalytic,sequence,sequencephd,diagnostics,latentclassif}. Additionally, finite partially ordered classification models are useful for many statistical applications
including cognitive modelling.  In particular, a data analytic framework for implementing latent finite partially ordered classification models introduced by Tatsuoka\cite{dataAnalytic}, provides  useful methods for evaluating  cognitive applications that are latent and complex.  Poset models are flexible and can become quite rich and complex, enabling them to
be effective models for describing response phenomena from educational test data or neuropsychological assessment data\cite{dataAnalytic,sequence2,neuronumeric}. An objective of neuropsychological assessment is to determine differences in cognitive functioning in clinical settings. Carr et al.\cite{cognitionamyloid} used poset classification models of neuropsychological test data to classify samples into detailed cognitive profiles using ADNI2(Alzheimer's Disease Neuroimaging Initiative) and AIBL(Australian Imaging, Biomarker  \& Lifestyle ) datasets. In the risk of disease progression of Alzheimer  disease for individuals with mild cognitive impairment
(MCI), Tatsuoka et al. \cite{cognitivemodeling} suggest that  poset-based modeling methods may be useful in providing more precise classification of cognitive subgroups among MCI for imaging and genetics studies, and for developing more efficient and focused cognitive test batteries. An order structure arises naturally with skill profiles. The flexibility to not necessarily assume that one state is greater than another is an appealing feature of posets\cite{diagnostics}. In addition,  posets are comprised of states, into which cases are classified, that are associated with distinct patterns of attribute strengths and weaknesses \cite{neuroshizo}. Posets have several advantages over conventional
statistical methods for handling large numbers of  polyfactorial neuropsychological test variables such as the ability to mimic the expert judgment of a clinical neuropsychologist for each case in a large sample. Posets are efficient in these tasks since as valid conclusions can be drawn based on relatively few measures, and classification of large samples can be accomplished rapidly \cite{neurocognitivep}.
On the other hand, Jaeger et al.\cite{neurocognitivep} noted that the primary limitation in  poset modeliing of large conventional neuropsychological test datasets is based on its restriction to a selected set of attributes, while additional important distinctions remain to be tested.

 \subsection{Other}
 
The application of posets and lattice theory are not limited to the aforementioned domains and tasks for data analysis. Posets have also been used in: 1) hypothesis management in large scale research projects \cite{latticehypothesis}; 2) longitudinal data analysis\cite{longid}; 3) social macroeconomics analysis\cite{macrosocio} ; 4) adaptive testing for cognitive assessment  \cite{sequenceexp}; 5) psychometrics\cite{beliefitems}; 6) medical statistics \cite{medstats}; 7) machine learning of reviewer’s paper preferences \cite{predrankABs}; 8)  visual similarity learning using pose estimations \cite{silarlearn}; 9) gender analysis \cite{gendereq}; 10)  Pattern Mining \cite{loglinearposets}; 11) large-scale data mining, where AI-integrated lattice and poset structures are combined with the algebraic features of these structures to process complex relational data, discover hidden patterns, and scale to large datasets~\cite{jdmscmining}. A further, weaker connection to poset structure appears in compiler optimization: pass-ordering search spaces are naturally partially ordered by which optimization passes must precede others, and recent reinforcement-learning-and-graph-neural-network frameworks for multi-objective compiler phase ordering~\cite{milestone} implicitly search this space, although the poset structure is not the explicit focus of that line of work and we flag it here only as an adjacent application rather than a core contribution to poset theory.

\subsection{Some Selected Datasets}

\begin{enumerate}

\item \textbf{Computer Vision Datasets}

\begin{itemize}

\item The Olympic Sports dataset  \cite{olympicdata}, The Leeds Sports Pose(LSP)\cite{lspdataset},  MPII Pose \cite{mp11pose}. They are  employed in the study of visual similarity learning from  pose estimation using posets.\cite{silarlearn}.  

\item  Micro-video dataset \cite{videodata}. It is used for the learning problem on hypergraph partial order \cite{hypergraph}.

\end{itemize}

\item  \textbf{Natural Language Datasets}:
\begin{itemize}

\item The Compositional Freebase Questions (CFQ)(Keysers et al., 2020)\cite{cfqdataset}   is a dataset that is specifically designed to measure compositional generalization. It is used for the study of  hierarchical poset decoding for compositional generalization in language  \cite{decodecomp}.

\item  Universal Dependencies (UD) corpora \cite{UDC}. It is used in the learning  algorithm for generating a surface word order for a sentence given
its dependency tree \cite{weigtposet}.

\end{itemize}

\item \textbf{Sustainanble Development Data}:

The sustainable datasets are categorized into socio-economic and environmetal  datasets.

\textbf{ Socio-economic data}

\begin{itemize}
\item  Equitable and sustainable well-being data  \cite{equidata}. Used in \cite{multiindictaor,whyagg} for multidimensional data analysis in the context of sustainable development.

 \item Service Performance Dataset   \cite{servicedata}.  Used for multidimensional data analysis in  \cite{partialeuro}. 

 \item Data on crime rates and their ranking for sixteen  American cities is described in Levy\cite{crime}.

 \end{itemize}

\item  \textbf{Environmental data}

\begin{itemize}

 \item A battery of biochemical, microbiological and bioassay tests were used to identify degraded or degrading sediments in waters  \cite{envrankontario,envshore,chemdualFCA}. The dataset is available in  \cite{seddata}. 

 \item Collection of toxicity data \cite{quantposet}. 

 \item  Multiple indicator data for stream channel stability at bridge crossing is described in  \cite{bridgeposets}.

 \end{itemize}

\item  \textbf{Other datasets}:

\begin{itemize}

 \item (i)Data covering a time period up to the end of 2022  on the historical head-to-head matches of six professional tennis players.(ii) Data on educational testing from 15 OECD countries  in.reading comprehension based on test performance in 2015 from the Programme for International Student Assessment (PISA). Both of these are described in \cite{modelselection} for the purpose of partial ranking of tennis players and total ranking of educational systems respectively.

 \item  A mass spectroscopy dataset consisting of 11 phosphoproteins and phospholipids  containing approximately 854 measurements of abundance levels in an observational setting described in the supplementary material  in  \cite{proteindata}. It is used in  Taeb et al. \cite{modelselection} for the purpose of  learning  causal relations and structures in proteins.

 \item UCI machine learning repository  \cite{ucidata}, used in \cite{fcaboosting}.

\item Poset/Hypergraph Currvature Datasets\cite{hypergraphdata},  used in  \cite{hypergraphpaper} to   perform an empirical study involving computation and analysis of the Forman–Ricci curvature of hyperedges in 12 real-world hypergraphs.

\end{itemize}

\end{enumerate}

\subsection{ Selected Algorithms and  Software Packages}

\begin{enumerate}

\item  \textbf{Machine learning and Deep learning}

\begin{itemize}

\item Python implementation of deep  visual similarity learning using posets \cite{pythonVS}. Used in  \cite{silarlearn}. 
\item Given as input an edge-weighted poset, the algorithm in \cite{weigtposet}  constructs a total order such that nodes with smallest weights are adjacent. The algorithm works by  attempting to order a set of words as closely as possible to their original surface  realization in the Universal Dependencies(UD) corpus. Due to the fact that words may repeat in the sentence, each order is instead represented by a list of integers, and it is these lists of integers which are compared \cite{weigtposet}.  For example, assuming a target reference order of [1,2,3] for the red horse, the generated order of red the horse would be [2,1,3].

\item Causal Structure Discovery Algorithm-Greedy Sparsest Poset (GSPo) , python implementation in \cite{causaldag}. Used  for causal structure learning in the presence of latent variables \cite{causal}.

\item  A  learning  algorithm  for discovering partial orders from sequences of events is described in \cite{seqdata}.

\item The learning algorithm of pertinent concept described in \cite{fcaboosting} uses the Adaboost algorithm with formal concept analysis on a learning dataset  to discover lattice concepts used for classification rules. 

\item  An algorithm framework for integrating base classifier into concept node of concept lattice is described in \cite{Zhipeng}.

\item   Greedy sequential algorithm for model selection described in   \cite{modelselection}. It is used in partial ranking of tennis players, total ranking in educational systems, and causal structure learning on proteins dataset.

\end{itemize}

\item  \textbf{Multidimensional data analyis}: 

\noindent  The Hasse Diagram Technique (HDT) described in section \ref{descsection}  provides a huge collection of methods that are  simple but tend to be complicated if the number of elements in the poset increases. To address this chalenge, special software packages have been developed to support the HDT. Some of these are described as follows.

\begin{itemize}

\item  The  PARSEC\cite{softReval,parsec} is an R package for poset-based evaluation of socio-economic data. Its main goal is to provide socio-economic scholars with an integrated set of elementary functions for multidimensional poverty evaluation based on ordinal information. The package is organized in four main parts  \begin{itemize}
\item 
\begin{inparaenum}[i)]
    \item Data management..
    \item Basic poset analysis.
    \item Poset-based evaluation.
\item  OPHI counting approach.
  \end{inparaenum}

\end{itemize}

\item  An algorithm based on posets for multidimensional data analysis is described in \cite{posetalgo}. 

\item Self-organizing map algorithm(SOM)\cite{selfmap} is a popular tool for non-linear dimensionality reduction and pattern recognition. It is used for multidimensional data analysis described in \cite{migrantspos}.

\item    A non-aggregative partial order algorithm for the construction of sustainability synthetic indicators on multi-indicators systems is described by Arcagni et al. \cite{posetalgo}.

\item   An application of partial order theory for object rank correlation analysis of  multiple variables is implemented  by the software package PO Correlation  \cite{posoftwaRe} . The
design is made transparent for rank correlation analysis by a detailed mapping of the rank relations between all objects.

\item  \  The software WHASSE is applied for chemical monitoring data analysis \cite{chemrank,hassechem,hassesoftw}, medical data analysis \cite{diazwhasse}. 

\item  PLMIX\cite{MollicaR} an R package for modeling and clustering partially ranked data.

\item PyHasse software \cite{pyHasse, pyhasse2,pyhasse3} is used for the purpose  of   ordinal analysis in data matrices,  identification and analysis of partial order relations as well as in computing ranks. Furthermore, it has been used in \cite{envdatahealth,rankenvsoft,envSWchem} for the analysis and evaluation of environmental data and in \cite{stakeholder} for multi-criteria decision analyses.

\item ProRank\cite{softposrank} software for partial order ranking. It is used in the evaluation of environmental databases  \cite{prorank2env}.

\end{itemize}

\end{enumerate}

\section{Limitations, Open Problems, and Future Directions}\label{futuredir}

Posets have proven their usefulness in a broad range of applications. However, several theoretical and practical challenges limit the extent to which poset-based methods have been adopted in mainstream machine learning, and a number of very recent results (2025--2026) suggest concrete directions in which these limitations can be addressed. We organize this discussion around six themes, using the taxonomy of Section~\ref{taxonomy} to relate each theme to the representation, paradigm, modality, and task it primarily affects.

\subsection{Model Depth and Expressivity}

A major challenge in data science is the identification of geometric structure in high-dimensional data. The structural understanding of data is very relevant for designing efficient algorithms for optimization and machine learning. Classically, the structure of data has been studied under Euclidean assumptions: the fundamental representation of the features for any machine learning model is the vector, and its multidimensional generalization, the tensor. Many machine learning algorithms and data-analysis pipelines have accordingly been developed with the aim of computing vectors or matrices of real numbers, and a wide range of well-studied tools and algorithms assume such structure. However, many scientific fields study data with an underlying structure that can only be represented in non-Euclidean space, such as graphs and manifolds, and posets -- as a class of directed acyclic graphs -- have only recently begun to receive attention with regard to developing appropriate \emph{deep} learning architectures rather than shallow order-theoretic statistics. Wendler~\cite{swiss} introduced methods for Fourier-sparse learning on data indexed by lattices and posets, and the graph neural network class of algorithms applies to posets as discussed in Section~\ref{mlnotes}, but there is not yet widespread use of graph neural networks relative to the number of applications based on posets and lattices. A concrete recent step toward genuinely deep, multi-layer poset-structured architectures is the poset-pooling construction of Dolores-Cuenca et al.~\cite{poolingorder}, who show that convolutional filters derived from four-point posets and their associated order polytopes update network weights during backpropagation with greater precision than average, max, or mixed pooling, without introducing additional trainable parameters, and who formalize the composition of such poset-neural architectures using the algebra of the operad of posets. This raises an open question of central importance for model depth: how does the expressivity of a poset-structured network (in the sense of, e.g., function classes realizable by stacking poset-pooling or poset-projection layers) scale with network depth, and what is lost or gained relative to unconstrained architectures when the intermediate representations are themselves required to respect a partial order? The success of machine learning algorithms generally depends on data representation, since different representations encode different explanatory factors of variation behind the data~\cite{representationL,daglatent}. There is accordingly enormous scope to develop deeper, more expressive architectures for learning the structural characteristics of data represented as Hasse graphs of lattices and posets, together with model-evaluation metrics designed specifically for poset-valued outputs.

\subsection{Scalability--Fidelity Trade-offs}

Many of the classical poset-analytic techniques surveyed in Sections~\ref{multidim}--\ref{descriptive} -- enumeration of linear extensions, exact computation of mutual ranking probabilities, exhaustive Hasse-diagram construction -- are worst-case exponential in the number of incomparable elements, which is precisely the situation in which poset methods are most informative relative to a forced total order. Sampling-based approximations such as Markov Chain Monte Carlo over linear extensions~\cite{envranking,fastlinear} trade exactness for scalability, but the bias--variance behavior of these approximations for downstream machine learning tasks (as opposed to descriptive ranking) is not well characterized. A related, and largely open, scalability question concerns \emph{distance and metric computation} over posets, which underlies nearest-neighbor, clustering, and kernel-based learning on order-structured data. Two 2026 contributions are directly relevant here. Taeb, Guo, and Henckel~\cite{modelgraphdist} propose a model-oriented framework in which each graph is treated as a statistical model organized in a poset by inclusion, and define a distance as the length of a shortest path through poset neighbors; this framework applies to probabilistic undirected graphs and to several classes of causal DAGs, but its computational cost scales with the size of the neighborhood structure of the poset, which is not addressed in closed form for large model spaces. Olave~\cite{posetdistfam} introduces a family of extended metrics on path-connected and fence-connected posets that do not require additional algebraic structure (e.g., a valuation), showing that these metrics characterize discrete path-connected posets up to isomorphism and coincide with interleaving distances when posets are viewed as thin categories; this is a promising building block for scalable metric learning on posets, but its practical computational complexity for large, densely comparable posets has not yet been benchmarked. Understanding when a cheaper, approximate poset metric preserves the statistical or predictive guarantees of the exact one -- analogous to Johnson--Lindenstrauss-type results for Euclidean embeddings -- remains an open problem.

The same exponential-versus-polynomial tension recurs, in sharper form, in two further problems that have not previously been connected to this scalability discussion: \emph{generation} and \emph{isomorphism testing} of posets and lattices at scale. On the generation side, exact enumeration algorithms produce a complete, duplicate-free catalog of all lattices of a given size, but scale in the size of the output itself -- which grows from $5{,}994$ non-isomorphic lattices at $n=10$ to over $23$ trillion at $n=20$ (OEIS A006966) -- confining exhaustive approaches to $n \lesssim 24$ in practice. Mwafise's reinforcement-learning generator~\cite{posetrlgen} (Section~\ref{safeRL}) instead trades the completeness guarantee of exact enumeration for a sampling procedure that scales in the size of a single candidate rather than the size of the target class, extending practical lattice discovery to $n \in \{30,40,50\}$ with mean discovery rates of $60$--$80\%$; the two approaches are explicitly complementary rather than competing, with exhaustive enumeration remaining preferable whenever a complete catalog at moderate $n$ is the actual goal. On the isomorphism-testing side, the general graph isomorphism problem is GI-complete and not known to admit a polynomial-time solution, and the classical poset-isomorphism solvers used throughout the descriptive and clustering literature of this survey (VF2-style backtracking) inherit this worst-case factorial cost. Mwafise's hierarchical matrix-decomposition framework~\cite{posetisodecomp} (Section~\ref{metricsub}) instead achieves empirically polynomial ($O(n^2)$ to $O(n^4)$, depending on structural regularity) isomorphism testing for a broad class of posets by recursively decomposing the poset matrix rather than searching over relabelings, with formally proved invariance under permutation. These two results are best read together: a scalable generator is only as useful as its ability to cheaply determine that a newly generated candidate is not a duplicate of one already found, which is exactly the isomorphism-testing bottleneck the decomposition framework targets, and we return to this generate-and-canonicalize relationship as a concrete open research direction in Section~\ref{genplanning}.

\subsection{Heterogeneity of Order-Structured Data}

Most poset-based methods surveyed in this paper assume a single, homogeneous order relation over a single population of objects. Real applications, however, frequently combine several heterogeneous and only partially compatible order relations: for example, multiple stakeholders may impose different priority orders over the same set of safety constraints, or a multi-indicator system may need to merge orders derived from measurements of different types and reliabilities (Section~\ref{multidim}). Wong, Xiao, and Rus~\cite{posafenet} address a version of this problem directly in the setting of safe reinforcement learning: rather than imposing a single fixed priority order or enforcing all safety constraints uniformly, they formalize safety requirements as a poset in which some constraints are comparable and others are genuinely incomparable, and introduce PoSafeNet, a differentiable neural safety layer that performs sequential closed-form projection consistent with the poset structure, enabling adaptive selection or mixing of valid safety executions while preserving priority semantics by construction. This is, to our knowledge, the first neural architecture designed explicitly around the possibility that constraints are only partially ordered rather than either freely combinable or strictly ranked, and it suggests a broader research direction: extending poset-structured layers to settings with heterogeneous, possibly conflicting orders supplied by different sources (annotators, sensors, stakeholders), and developing principled methods for reconciling or completing such orders, in the spirit of the completion results for leveled posets discussed below.

\subsection{Dynamicity of Evolving Posets}

The vast majority of the theory and algorithms surveyed in this paper -- Hasse diagram construction, linear extension counting, formal concept lattices, poset-based ranking -- is formulated for a static poset. Many of the application domains reviewed in Sections~\ref{multidim} and~\ref{appsec}, however, are inherently temporal: multi-indicator systems evolve year over year~\cite{multiindictaor}, safety constraints in control systems can activate or deactivate along a trajectory~\cite{posafenet}, and streaming or additive-manufacturing data is generated incrementally layer by layer. Fueyo et al.~\cite{leveledposets} introduce \emph{leveled partially ordered sets} to manage exactly this kind of layer-indexed data, motivated by an industrial 3D-printing application in which points must be ordered layer by layer; they generalize Birkhoff's notion of element height to the leveled setting and prove that, under conditions related to the Jordan--Dedekind chain condition, a leveled poset can always be completed into a bounded lattice purely by adding links to the existing order relation, without introducing new elements. This is a rare example of a completion result that could plausibly be adapted to online or streaming settings, where new layers (or new time steps) arrive incrementally and the poset -- and any downstream lattice used for analysis -- must be updated rather than recomputed from scratch. Formalizing and analyzing such incremental poset-update algorithms, together with their approximation guarantees relative to full recomputation, remains largely open.

\subsection{Posets, Category Theory, and Multiparameter Topological Data Analysis}

Topological data analysis (TDA)~\cite{toplodata} is a fast-growing field providing topological and geometric tools to infer relevant features from complex data. In (single-parameter) persistent homology, the shape of a dataset is often encoded into a system of vector spaces and linear maps over a totally ordered set, and a growing body of literature connects posets more generally to topological data analysis~\cite{jardinetda,caputitda,botmantda,tombaritda}. The genuinely hard case is \emph{multiparameter} persistence, where the indexing structure is a poset (typically a lattice) rather than a chain, and modules over this poset generally fail to decompose into simple interval summands, unlike in the single-parameter case. Hem~\cite{posetcocalculus} introduces \emph{poset cocalculus}, a variant of functor calculus defined for functors from a poset (or, more specifically, a distributive lattice) to a model category, and shows that it produces stable degree-$n$ approximations of a functor under an appropriate interleaving distance. In a companion paper, Hem~\cite{multiperscocalc} applies poset cocalculus to prove that a pointwise finite-dimensional bipersistence module is middle-exact if and only if it is isomorphic to the homology of a homotopy-degree-1 functor, yielding a new, more synthetic proof of the interval decomposability of middle-exact bipersistence modules, and gives a decomposition theorem expressing a middle-exact multipersistence module as a direct sum of a projective, an injective, and a bidegree-1 module. These results connect the purely combinatorial poset structures used throughout this survey to the algebraic and homotopy-theoretic machinery of functor calculus, and, together with the algebraic/operadic perspective on posets developed by Arciniega-Nev\'{a}rez, Berghoff, and Dolores-Cuenca~\cite{operadwixarika} (who study posets as algebras over an operad and use this framework to analyze the combinatorics of a nontrivial suboperad, the Wix\'{a}rika posets), point toward a category-theoretic reformulation of several of the ranking and clustering constructions in Sections~\ref{clustersec}--\ref{descriptive}. There is more generally a need to study multiway interactions within data endowed with a poset structure, for example by leveraging concepts such as Latent Topology Inference (LTI)~\cite{TID}, and to connect the resulting topological deep learning models~\cite{TDL} to the functor-calculus formalism above.

A second, complementary operadic perspective targets the \textit{matrix} representation of posets directly rather than the poset as an abstract order relation. Cheon, Choi, Giraudo, and Mwafise~\cite{operadposetmatrices} define a family of partial composition operations on poset matrices that build larger poset matrices from smaller ones, prove that three of these operations satisfy the operad axioms -- giving the collection of poset matrices itself a genuine operad structure, distinct from and complementary to the Wix\'{a}rika suboperad of~\cite{operadwixarika} -- characterize the corresponding dual operations, and use the resulting framework to address open questions in poset enumeration. Notably, the same operadic vocabulary reappears independently, and in an entirely algorithmic register, in the isomorphism-testing framework of Section~\ref{metricsub}: Mwafise~\cite{posetisodecomp} explicitly frames the Hierarchical Poset Matrix Tree(HPMT) as a combinatorial invariant developed for the efficient resolution of the poset isomorphism problem. The HPMT Tree can also be viewed within the broader framework of machine learning over combinatorial tree structures, a growing area spanning structural learning and geometric deep learning. Its recursive boundary-stripping and core-extraction procedure exploits the ``operadic properties of poset matrices\cite{mwafise2022poset}'' by decomposing a poset into hierarchically composed substructures. This suggests that its algorithmic tree decomposition may provide a natural bridge between operadic composition and learnable representations of hierarchical poset structure– is both plausible and currently unestablished.

\subsection{Safe and Constrained Learning, and Order-Theoretic Reinforcement Learning}

A further theme that has not previously been surveyed in connection with posets is the use of order-theoretic formalisms in reinforcement learning and control. In addition to the safety-layer work of Wong, Xiao, and Rus discussed above~\cite{posafenet}, Sargent and Stachurski~\cite{dposets} represent a dynamic program as a family of operators acting on a partially ordered set and give an optimality theory based purely on order-theoretic assumptions, covering applications ranging from traditional Markov decision processes to nonlinear recursive preferences, robustness objectives, and distributional dynamic programming. Peng, Stachurski, and Yang~\cite{abstractdynprog} extend this line of work by pairing the order-theoretic approach with topological and metric foundations, showing that readily verifiable forms of topological stability (global stability and contractivity of the policy operators) deliver both the fundamental optimality properties of dynamic programming and convergence of value function iteration, Howard policy iteration, and optimistic policy iteration, with applications including optimal stopping and Bayesian sequential analysis. Together with the classical machine-learning-and-formal-concept-analysis material of Section~\ref{fcaTT}, this suggests that order theory can serve not only as a data-analytic and representational tool, as in most of this survey, but also as a foundational framework for the algorithmic and convergence theory of reinforcement learning itself -- an angle that, to our knowledge, has not previously been connected to the broader poset-and-machine-learning literature surveyed here. A more indepth review of existing and new models for machine learning using formal concept approaches, and of graph-based learning-to-rank methods~\cite{gnnLTR2,gnnLTR1} applied to datasets with an underlying partial order (Sections~\ref{multidim} and~\ref{descriptive}), would also be a useful complement to the material collected in this survey.

\subsection{Toward a Generate--Canonicalize--Compose Pipeline for Large-Scale Poset Discovery}\label{genplanning}

Collectively, the reinforcement-learning generator~\cite{posetrlgen}, the hierarchical isomorphism-decomposition framework~\cite{posetisodecomp}, and the operad of poset matrices~\cite{operadposetmatrices} discussed across Sections~\ref{safeRL}, \ref{metricsub}, and~\ref{futuredir} above chain naturally into a three-stage pipeline that, to our knowledge, has not been proposed as such in the existing literature: \emph{generate} candidate posets or lattices at scale beyond the reach of exact enumeration; \emph{canonicalize} them cheaply via polynomial-time isomorphism testing, deduplicating discoveries and enabling large-scale indexing; and \emph{compose} validated primitives into larger structures, or decompose known large lattices into their generating components, using a formal operad. Each stage already exists in isolation, but their combination suggests concrete extensions beyond what any one paper claims. First, canonicalization is not merely a downstream convenience for a generator: the discovery rates reported by the RL framework are defined over \emph{labeled} structures, precisely because no fast isomorphism-reduction step was integrated into its training loop; wiring the hierarchical decomposition framework's near-linear Tier-1/Tier-2 checks (Section~\ref{metricsub}) directly into the novelty bonus of the generator's reward function would convert a labeled discovery rate into an isomorphism-aware coverage measure -- a gap the RL paper's own stated future work identifies as open, alongside its separately stated interest in ``integration of graph-based machine learning techniques for enhanced structural analysis.'' Second, the operad of poset matrices offers a principled, order-theoretically legal action space: rather than the free pairwise edge choices of the current generator, an RL agent could instead be restricted to operadic composition moves, which would guarantee every intermediate and terminal candidate is a well-formed poset matrix by construction rather than by post hoc verification, potentially easing the combinatorial sparsity that motivates the current framework's exact-verification-oracle design (Section~\ref{safeRL}). Third, beyond generation and testing, this same toolchain is a plausible basis for constructing the first large-scale, deduplicated benchmark corpus of labeled and canonicalized posets and lattices, of the kind needed to systematically evaluate the poset-pooling architectures, poset-valued metrics, and safety layers surveyed throughout Sections~\ref{mlnotes}--\ref{metricsub} -- none of which currently have access to anything resembling a standardized, large-scale evaluation dataset. We flag all three directions as concrete, currently unrealized extensions of existing work rather than results established by the papers cited.

\subsection{Summary of Open Problems}

Table~\ref{tab:futuredir} summarizes the open problems discussed above together with the taxonomy axis (Table~\ref{tab:taxonomy}) each one primarily stresses, to make explicit that the current literature is comparatively dense along the representation and task axes but comparatively sparse along the heterogeneity and dynamicity dimensions of the learning-paradigm and data-modality axes.

\begin{table}[H]
\centering
\caption{Open problems in poset-based machine learning and the taxonomy axis each primarily stresses.}
\label{tab:futuredir}
\small
\begin{tabular}{@{}>{\raggedright\arraybackslash}p{3.6cm} >{\raggedright\arraybackslash}p{6.8cm} >{\raggedright\arraybackslash}p{2.6cm}@{}}
\toprule
\textbf{Open problem} & \textbf{Description} & \textbf{Taxonomy axis stressed} \\
\midrule
Model depth / expressivity & Scaling poset-structured layers (e.g. poset pooling) to deep, multi-layer architectures & Representation \\
Scalability--fidelity trade-off & Approximate linear-extension sampling, poset metrics, and RL-based lattice generation at scale & Task (metric computation, generation) \\
Heterogeneity & Reconciling multiple, only partially compatible order relations (safety, multi-stakeholder) & Learning paradigm \\
Dynamicity & Online/incremental posets and lattice completions for streaming or layered data & Data modality \\
Category-theoretic / topological integration & Poset cocalculus and operadic structure (of posets and of poset matrices) for multiparameter TDA & Representation \\
Order-theoretic RL & Convergence theory for dynamic programming, and generation, posed directly over partial orders & Learning paradigm \\
Generate--canonicalize--compose pipeline & Coupling RL-based generation, polynomial-time isomorphism testing, and operadic composition into a unified, benchmark-ready toolchain & Task (generation, isomorphism testing) \\
\bottomrule
\end{tabular}
\end{table}

\section{Conclusion}\label{conclusionsec}

This survey has provided a comprehensive, taxonomy-organized review of the use of partially ordered sets in machine learning and data analysis. The breadth of application of poset theory across machine learning and data analysis -- from ranking and clustering to formal concept analysis, safe reinforcement learning, and multiparameter topological data analysis -- demonstrates its continued relevance. Building on an extensive account of the classical literature on descriptive, multidimensional, and exploratory poset-based data analysis, we incorporated a substantial body of methods published in 2025--2026 that had not previously been connected to this literature, including poset-structured safety layers for reinforcement learning~\cite{posafenet}, order-theoretic formulations of abstract dynamic programming~\cite{dposets,abstractdynprog}, new families of metrics over posets~\cite{modelgraphdist,posetdistfam}, poset cocalculus for multiparameter persistence~\cite{posetcocalculus,multiperscocalc}, two complementary operadic/algebraic perspectives on poset combinatorics~\cite{operadwixarika,operadposetmatrices}, leveled posets for layer-indexed and streaming data~\cite{leveledposets}, poset-pooling neural architectures for interpretable deep learning~\cite{poolingorder}, reinforcement-learning-based generation of finite lattices and semilattices~\cite{posetrlgen}, and a polynomial-time hierarchical matrix-decomposition framework for poset isomorphism testing~\cite{posetisodecomp}. This last pair -- a generator that samples posets at scale beyond the reach of exact enumeration, and a canonicalization procedure that tests isomorphism far faster than classical backtracking -- suggested a concrete, currently unrealized generate--canonicalize--compose pipeline (Section~\ref{genplanning}) that we believe is a promising direction for building the large-scale, deduplicated poset benchmarks this field currently lacks. Organizing this material along the four-axis taxonomy introduced in Section~\ref{taxonomy} makes clear that the field is comparatively mature along the representation and task axes but still comparatively underdeveloped with respect to heterogeneous and dynamically evolving posets, and with respect to the expressivity of deep, multi-layer poset-structured architectures -- gaps that we outlined as a concrete research agenda in Section~\ref{futuredir}. We aim for this survey to provide both a foundational reference for researchers newly entering this area and a practical map of open problems for those already working at the intersection of order theory and machine learning.


\begin{thebibliography}{9}


\bibitem{multiindictaor}Alaimo, L.S., Arcagni, A., Fattore, M. et al. Synthesis of Multi-indicator System Over Time: A Poset-based Approach. Soc Indic Res 157, 77–99 (2021).

\bibitem{measuredev}Alaimo, L.S., Ciacci, A. \& Ivaldi, E. Measuring Sustainable Development by Non-aggregative Approach. Soc Indic Res 157, 101–122 (2021).

\bibitem{whyagg}L. S. Alaimo \& F. Maggino, 2020. ``Sustainable Development Goals Indicators at Territorial Level: Conceptual and Methodological Issues--The Italian Perspective," Social Indicators Research: An International and Interdisciplinary Journal for Quality-of-Life Measurement, Springer, vol. 147(2), pages 383-419.

\bibitem{LTRauto}A. Albuquerque, T. Amador, R. Ferreira, A. Veloso and N. Ziviani, ``Learning to Rank with Deep Autoencoder Features," 2018 International Joint Conference on Neural Networks (IJCNN), Rio de Janeiro, Brazil, 2018, pp. 1-8, doi: \url{10.1109/IJCNN.2018.8489646}.

\bibitem{fcarandom}M. A. Ali, A. Jaoua and S. A. Al-Maadeed, ``A novel Conceptual Machine Learning Method using Random Conceptual Decomposition," 2020 IEEE International Conference on Informatics, IoT, and Enabling Technologies (ICIoT), Doha, Qatar, 2020, pp. 18-22.

\bibitem{learnclassruledata}A An.,  Learning Classification Rules from Data, (2003) \url{http://www.cs.yorku.ca/~aan/research/paper/cam03.pdf}.

\bibitem{covermatrices}Anđelić, M., da Fonseca, C.M. Cover matrices of posets and their spectra. Czech Math J 59, 1077–1085 (2009). \url{https://doi.org/10.1007/s10587-009-0075-6}

\bibitem{FCAConvert}Andrews, S.: Data Conversion and Interoperability for FCA. In: CS-TIW 2009, pp. 42-49, \url{http://www.kde.cs.uni-kassel.de/ws/cs-tiw2009/proceedings_final_
15July.pdf} 

\bibitem{dataformconc} S. Andrews and C. Orphanides, “Analysis of large data sets using
formal concept lattices,” In: M. Kryszkiewicz and S. Obiedkov, (eds.) \textit{Proceedings of the 7th International Conference on Concept Lattices and Their Applications}. Seville, University of Seville, 2010, 104-115.

\bibitem{VFCAdata}Andrews, S., Polovina, S., Visualising computational intelligence through converting data into formal concepts(2011), \url{https://shura.shu.ac.uk/2720/}.

\bibitem{chemdualFCA} P. Annoni, R. Bruggemann, The dualistic approach of FCA: A further insight into Ontario Lake sediments, \textit{Chemosphere} \textbf{70} (2008) 2025--2031.


\bibitem{partialeuro} Annoni, P., Brüggemann, R. Exploring Partial Order of European Countries. Soc Indic Res 92, 471–487 (2009).

\bibitem{annrankchem}P. Annoni, R. Brüggemann, A. Saltelli, Partial order investigation of multiple indicator systems using variance-based sensitivity analysis,
\textit{Environmental Modelling \& Software}, Vol.  26;  7, 2011, pp. 950-958, \url{https://doi.org/10.1016/j.envsoft.2011.01.008}

\bibitem{anonifuzzy}Annoni, P., Fattore, M. \& Bruggemann R. (2011). A Multi-Criteria Fuzzy Approach for Analyzing Poverty structure. Statistica \& Applicazioni, Special
Issue, 7-30.

\bibitem{operadwixarika}Arciniega-Nev\'{a}rez, J.A., Berghoff, M., \& Dolores-Cuenca, E. (2026). Operad of posets 101: The Wix\'{a}rika posets. \textit{Journal of Prime Research in Mathematics}, 22(1), 29--43. arXiv:2406.07370.

\bibitem{softReval}Arcagni, A., \& Fattore, M. (2014). PARSEC: An R package for poset-based evaluation of multidimen-sional poverty. In R. Bruggemann, L. Carlsen, \& J. Wittmann (Eds.), \textit{Multi-indicator systems andmodelling in partial order}. Berlin: Springer. 

\bibitem{migrantspos}Arcagni, A., Barbiano di Belgiojoso, E., Fattore, M. et al. Multidimensional Analysis of Deprivation and Fragility Patterns of Migrants in Lombardy, Using Partially Ordered Sets and Self-Organizing Maps. Soc Indic Res 141, 551–579 (2019).

\bibitem{posetalgo}Arcagni  A., Cavalli L., and Fattore, M., Partial Order Algorithms for the Assessment of Italian Cities Sustainability (February 3, 2021). FEEM Working Paper No. 1.2021, Available at \url{ https://ssrn.com/abstract=3778559 or http://dx.doi.org/10.2139/ssrn.3778559}

\bibitem{multidomainsys}Arcagni, A., A. Avellone, and M. Fattore (2022). Complexity reduction and approximation of multidomain systems of partially ordered data. \textit{Computational Statistics \& Data Analysis 173}, 107520.

\bibitem{EnsembleBB}Avogadri, R., Valentini, G. (2008). Ensemble Clustering with a Fuzzy Approach. In: Okun, O., Valentini, G. (eds) Supervised and Unsupervised Ensemble Methods and their Applications. Studies in Computational Intelligence, vol 126. Springer, Berlin, Heidelberg.\url{ https://doi.org/10.1007/978-3-540-78981-9_3}

\bibitem{grapheuclid}N. A. Asif et al.,``Graph Neural Network: A Comprehensive Review on Non-Euclidean Space," in \textit{IEEE Access}, vol. 9, pp. 60588-60606, 2021, doi: \url{10.1109/ACCESS.2021.3071274}.

\bibitem{jsmoperations}M. A. Babin and S. O. Kuznetsov, Enumerating Minimal Hypotheses and Dualizing Monotone Boolean Functions on Lattices 

\bibitem{summarydata}J. Bachtrögler, H. Badinger, A. F. de Clairfontaine, and W. H.Reuter, “Summarizing data using partially ordered set theory: anapplication to fiscal frameworks in 97 countries,” \textit{Stat. J. IAOS}, vol.32, no. 3, pp. 383–402 

\bibitem{badingerfiscal}Badinger H., Reuter W. H. (2015). Measurement of Fiscal Rules: Introducing the Application of Partially Ordered Set (POSET) Theory. Journal of Macroeconomics, 43, 108--23.

\bibitem{hassANOVA}R. A. Bailey (2021) Hasse diagrams as a visual aid for linear models and analysis of variance, Communications in Statistics - Theory and Methods, 50:21, 5034-5067, DOI: 10.1080/03610926.2019.1676443

\bibitem{fcAOntoInduct}Bain, M. (2003). Inductive Construction of Ontologies from Formal Concept Analysis. In: Gedeon, T.(.D., Fung, L.C.C. (eds) AI 2003: Advances in Artificial Intelligence. AI 2003. Lecture Notes in Computer Science(), vol 2903. Springer, Berlin, Heidelberg.\url{https://doi.org/10.1007/978-3-540-24581-0_8}.

\bibitem{TSP}S. Barbarossa and S. Sardellitti, ``Topological signal processing over simplicial complexes,” \textit{IEEE Transactions on Signal Processing, 2020.}
 
\bibitem{fcaordreFR}Barbut, M. and Monjardet, B., Ordre et classification, II, Paris: Hachette, 1970.

\bibitem{socialorder}Barthélemy, J.P., Flament, C., Monjardet, B. (1982). Ordered Sets and Social Sciences. In: Rival, I. (eds) Ordered Sets. NATO Advanced Study Institutes Series, vol 83. Springer, Dordrecht.

\bibitem{mlperformance}S. M. Basha, D. S. Rajput, Survey on Evaluating the Performance of Machine Learning Algorithms: Past Contributions and Future Roadmap, \url{https://doi.org/10.1016/B978-0-12-816718-2.00016-6}.

\bibitem{TID}Battiloro C. et al, From Latent Graph to Latent Topology Inference: Differentiable Cell Complex Module, arXiv:2305.16174v2.

\bibitem{silarlearn}M.A. Bautista, A. Sanakoyeu, B. Ommer,  Deep Unsupervised Similarity Learning using Partially Ordered Sets, \textit{2017 IEEE Conference on Computer Vision and Pattern Recognition(CVPR)}, Honolulu HI USA 2017,pp. 1923--1932, doi: 10.1109/CVPR.2017.208.

\bibitem{formalexplore}Belohlavek, R., Sklenar, V., \& Zacpal, J. (2004). Formal concept analysis with hierarchically ordered attributes. International Journal of General Systems, 33(4), 383–394. \url{https://doi.org/10.1080/03081070410001679715}

\bibitem{representationL}Y. Bengio, A. Courville, \& P. Vincent†,(2014) Representation Learning: A Review and New Perspectives: arXiv:1206.5538v3.

\bibitem{causal} Bernstein, D.I.; Saeed, B.; Squires, C.;Uhler, C.; Ordering-Based Causal Structure Learning in the Presence of Latent Variables, Proceedings of the 23$^{\text{rd}}$ International Conference on Artificial Intelligence and Statistics (AISTATS) 2020, Palermo, Italy. PMLR: Volume 108.

\bibitem{pyramidORD} Bertrand, M.F. Janowitz, Pyramids and weak hierarchies in the ordinal model for clustering, Discrete Applied Mathematics, Vol. 122;  1–3, 2002, pp.  55-81,
\url{https://doi.org/10.1016/S0166-218X(01)00354-7}.

\bibitem{multilabelclass} Bi, W. \& ;  Kwok, J.. (2013). Efficient Multi-label Classification with Many Labels. \textit{Proceedings of the 30th International Conference on Machine Learning}, in \textit{Proceedings of Machine Learning Research} 28(3):405-413 Available from \url{https://proceedings.mlr.press/v28/bi13.html.}

\bibitem{barrettlattice} G. Birkhoff, Lattice Theory,  American Mathematical Society,  Vol. 25; 3 1967.

\bibitem{statmodel}Blocher, H., Schollmeyer, G., Jansen, C. (2022). Statistical Models for Partial Orders Based on Data Depth and Formal Concept Analysis. In: Ciucci, D., et al. Information Processing and Management of Uncertainty in Knowledge-Based Systems. IPMU 2022. Communications in Computer and Information Science, vol 1602. Springer, Cham. \url{https://doi.org/10.1007/978-3-031-08974-9_2}.

\bibitem{depthfns}H. Blocher, G. Schollmeyer, C. Jasen,  M. Nalenz, Depth Functions for Partial Orders with a Descriptive Analysis of Machine Learning Algorithms, Proceedings of Machine learning Research 215: 59--71, 2023.

\bibitem{botmantda}MB Botnan, Topological Data Analysis, Lecture Notes, \url{https://www.few.vu.nl/~botnan/lecture_notes.pdf}

\bibitem{surveydeepattentionNN}G. Brauwers and F. Frasincar, "A General Survey on Attention Mechanisms in Deep Learning" in \textit{IEEE Transactions on Knowledge \& Data Engineering}, vol. 35, no. 04, pp. 3279-3298, 2023. doi: \url{10.1109/TKDE.2021.3126456}.

\bibitem{HierarchyMDA}Brisaboa, N.R., Cerdeira-Pena, A., López-López, N., Navarro, G., Penabad, M.R., Silva-Coira, F. (2016). Efficient Representation of Multidimensional Data over Hierarchical Domains. In: Inenaga, S., Sadakane, K., Sakai, T. (eds) String Processing and Information Retrieval. SPIRE 2016. Lecture Notes in Computer Science, vol 9954. Springer, Cham. \url{https://doi.org/10.1007/978-3-319-46049-9_19}.

\bibitem{posetvscluster}Brüggemann, R., Münzer, B., \& Halfon, E. (1994). An algebraic/graphical tool to compare ecosystems with respect to their pollution. The German River Elbe as an example. I : Hasse-diagrams. \textit{Chemosphere}, 28, 863-872.

\bibitem{chemenvtest}R. Bruggemann, J. Schwaiger, R. D. Negele, Applying Hasse diagram technique for the evaluation of toxicological fish tests, \textit{Chemosphere} \textbf{30} (1995) 1767–1780.


\bibitem{onlinedatabase}Bruggemann, R. and K. Voigt (1995). “An Evaluation of Online Databases by Methods of Lattice Theory.” Chemosphere 31: 3585-3594.

\bibitem{envrankaqua}Bruggemann, R.; Oberemm, A.; Steinberg, C. Ranking of Aquatic Effect Tests Using Hasse Diagrams. \textit{Toxicol. EnViron. Chem.} 1997, 63, 125-139.

\bibitem{envrankontario}Bruggemann, R.; Halfon, E. Comparative Analysis of Nearshore
Contaminated Sites in Lake Ontario: Ranking for Environmental Hazard. \textit{J. EnViron. Sci. Healt}h 1997, A32(1), 277-292.

\bibitem{envgraph}R. Bruggemann, S. Pudenz, K. Voigt, A. Kaune, K. Kreimes, An algebraic/graphical tool to compare ecosystems with respect to their pollution. IV: Comparative regionalanalysis by Boolean arithmetics, \textit{Chemosphere} \textbf{38}(1999) 2263–2279.

\bibitem{firstrank} Br\"{u}ggemann, R. and H.-G. Bartel (1999) A Theoretical Concept to Rank Environmentally Significant Chemicals. J.Chem.Inf.Comp.Sc. 39, 211-217 .

 \bibitem{envshore}R. Bruggemann, E. Halfon, G. Welzl, K. Voigt, C. Steinberg, Applying the concept of partially ordered sets on the ranking of near-shore sediments by a battery of tests,  \textit{J. Chem. Inf. Comp. Sci.}  \textbf{ 41 }(2001) 918--925.

\bibitem{chemrank}R. Bruggemann, U. Simon, S. Mey, Estimation of averaged ranks by extended local  partial order models,\textit{ MATCH Commun. Math. Comput. Chem.} \textbf{54} (2005) 489--518.

\bibitem{envchem}Brüggemann, R.; Carlsen, L. Partial Order in Environmental Sciences and Chemistry; Springer, 2006.

\bibitem{hassechem}R. Bruggemann, G. Restrepo, K. Voigt, Structure--fate relationships of organic chemicals derived from the software packages E4CHEM and WHASSE, \textit{J. Chem. Inf.Model}. \textbf{46} (2006) 894–902.

\bibitem{hassetechchem}R. Bruggemann, K. Voigt, Basic principles of Hasse diagram technique in chemistry, \textit{Comb. Chem. \& High Throughput Screen}. \textbf{11} (2008) 756–769.

\bibitem{rankenvsoft}R. Bruggemann, K. Voigt, G. Restrepo, U. Simon, The concept of stability fields and hot spots in ranking of environmental chemicals, \textit{J. Environ. Model. \& Soft.} \textbf{ 23} (2008) 1000–1012.

\bibitem{rankenvsoft22} R. Bruggemann, K. Voigt, Analysis of partial orders in environmental systems apply-ing the new software PyHasse, in: J. Wittmann, M. Flechsig (Eds), \textit{Simulation inUmwelt- und Geowissenschaften- Workshop Potsdam} 2009, Shaker–Verlag, Aachen, 2009, pp. 43–55.

\bibitem{multenv} Bruggemann, R., Patil, G.P. Multicriteria prioritization and partial order in environmental sciences.\textit{ Environ Ecol Stat} 17, 383–410 (2010).\url{ https://doi.org/10.1007/s10651-010-0167-3}

\bibitem{avgrank}R. Bruggemann, L. Carlsen, An improved estimation of averaged ranks of partial orders, \textit{MATCH Commun. Math. Comput. Chem}. \textbf{65} (2011) 383–414.

\bibitem{rankindic} Bruggemann R, Patil G.P. Ranking and prioritization for multi-indicator systems—Introduction to partial order applications.Springer; 2011.

\bibitem{rankpolychlorinate}Brüggemann, R. and K. Voigt (2011). “A New Tool to Analyze Partially Ordered Sets. Application: Ranking of Polychlorinated Biphenyls and Alkanes/Alkenes in River Main, Germany.” \textit{MATCH: Communications in Mathematical and in Computer Chemistry} 66: 231--251.

\bibitem{clusterincomp}R.  Bruggemann, L. Carlsen, Incomparable: What now II? Absorption of incomparabilities by a cluster method, Quality \& Quantity 49(4), (2014).

\bibitem{pyHasse}Brüggemann, R.; Carlsen, L.; Voigt, K.; Wieland, R. PyHasse Software for Partial Order Analysis: Scientific Background andDescription of Selected Modules. In \textit{Multi-Indicator Systems and Modelling in Partial Order}; Springer: New York, NY, USA, 2014; pp. 389–423. 

\bibitem{incomparable}Bruggemann, R.; Carlsen, L. Incomparable—What now? \textit{MATCH Commun. Math. Comput. Chem}. \textbf{2014}, 71, 694–716.

\bibitem{noisyposet}Bruggemann, R. and Carlsen, L. An attempt to Understand Noisy Posets. MATCH Commun.Math.Comput.Chem., \textbf{2016}, 75, 485-510.

\bibitem{complexphenomena}Bruggemann, R. et al.(Eds.)(2021). \textit{Measuring and Understanding Complex Phenomena: Indicators and Their Analysis in Different Scientific Fields}. London, UK, Springer Nature.

\bibitem{pyhasse2} Bruggemann, R., Kerber, A., Koppatz, P., Pratz, V. (2021). PyHasse, a Software Package for Applicational Studies of Partial Orderings. In: Bruggemann, R., Carlsen, L., Beycan, T., Suter, C., Maggino, F. (eds) Measuring and Understanding Complex Phenomena. Springer, Cham.\url{ https://doi.org/10.1007/978-3-030-59683-5_18}.

\bibitem{rankinadequate}Brunsdon, C. Rank Inadequacy: A Partially Ordered Set Approach For Multivariate Data Analysis. \url{https://huckg.is/gisruk2017/GISRUK_2017_paper_31.pdf}.

\bibitem{fastlinear}Bubley R, Dyer M (1999) Faster random generation of linear extensions. Discrete Math 201(1–3):81–88.

\bibitem{cantercrime}Canter, D. A Partial Order Scalogram Analysis of Criminal Network Structures. Behaviormetrika 31, 131–152 (2004). \url{https://doi.org/10.2333/bhmk.31.131}

\bibitem{caoLTR} Cao et al., Learning to Rank: From Pairwise Approach to Listwise Approach, \textit{ICML '07: Proceedings of the 24th international conference on Machine learningJune} 2007Pages 129–136 \url{https://doi.org/10.1145/1273496.1273513}.

\bibitem{posetbigdata} Caperna, G., \& Boccuzzo, G. (2018). Use of poset theory with big datasets: A new proposal applied to the analysis of life satisfaction in Italy. \textit{Social Indicators Research}, 136, 1071–1088.

\bibitem{caputitda}Caputi, L., Collari, C. \& Di Trani, S. Combinatorial and topological aspects of path posets, and multipath cohomology. \textit{J Algebr Comb} 57, 617–658 (2023). \url{https://doi.org/10.1007/s10801-022-01180-9}.


\bibitem{envnerveagents}L. Carlsen, Partial order ranking of organophosphates with special emphasis on nerve agents,  \textit{MATCH Commun. Math. Comput. Chem}. \textbf{54} (2005) 519–534.

\bibitem{apor}Carlsen L. Assessment of chemicals applying partial order ranking techniques. Comb Chem High Throughput Screen. 2008 Dec;11(10):794-805. doi: 10.2174/138620708786734280. PMID: 19075601

\bibitem{envpollution}  Carlsen, R. Bruggemann, Partial order ranking as a tool in environmental impact assessment. PAH and PCB pollution of the river Main as an illustrative example, in:G. T. Halley, Y. T. Fridian (Eds.), \textit{Environmental Impact Assessment}, Nova Science Publishers, 2009, pp. 335–354.

\bibitem{envhazard} Carlsen, B. N. Kenessov, S. B. Batyrbekova, A QSAR/QSTR study on the human health impact of the rocket fuel 1,1-dimethylhydrazine and its transformation products. Multicriteria hazard ranking based on partial order methodologies, \textit{Environ. Tox. Pharm}. \textbf{27} (2009) 415–423.

\bibitem{conceptAnal}L. Carlsen,The interplay between QSAR/QSPR studies and partial order ranking and formal concept analyses, \textit{Int. J. Mol. Sci.} \textbf{10} (2009) 1628–1657.

\bibitem{analyticperform}Carlsen L.; Bruggemann R.; Kenessova O.; Erzhigitov E. Evaluation of analytical performance based on partial order methodology. Talanta 2015, 132, 285-293.

\bibitem{infnoise}Carlsen, L.; Bruggemann, R. On the influence of data noise and uncertainty on ordering of objects, described by a multi-indicatorsystem. A set of pesticides as an exemplary case. \textit{J. Chemom}.  \textbf{2016}, 30, 22–29 .

\bibitem{carlsenfragile}Carlsen L., Brueggemann R. (2017). Fragile State Index: Trends and Developments. A Partial Order Data Analysis. Social Indicators Research 133, 1-14).

\bibitem{eukeys}  Carlsen L. (2017). An Alternative View on Distribution Keys for the Possible
Relocation of Refugees in the European Union. Social Indicators Research.,
130, 1147-1163.

\bibitem{happiness}Carlsen, L. (2018). Happiness as a sustainability factor.The world happiness index: A posetic-based dataanalysis. \textit{Sustainability Science}, 13(2), 549–571 

\bibitem{failedstate}Carlsen, L., \& Bruggemann, R. (2019). An analysis of the ‘Failed States Index’ by partial order methodology. J \textit{ournal of Social Structure}, 14(1), 1–31.

\bibitem{inequalitiesEU}Carlsen, L.; Bruggermann, R. Inequalities in the European Union—A Partial Order Analysis of the Main Indicators,  \textit{Sustainability} (2021), 13, 62--78.

\bibitem{gendereq}Carlsen, L., \& Bruggemann, R. (2021). Gender equality in Europe: The development of the sustainabledevelopment goal No. 5 illustrated by exemplary cases. Social Indicators Research, 158(3), 1127-1151.  

\bibitem{decisionsupprt}Carlsen, L.; Bruggemann, R. Partial Order as Decision Support Between Statistics and Multi-criteria Decision Analyses. \textit{Standards} \textbf{2022}, 2, 22.

\bibitem{stakeholder}Carlsen, L.; Bruggemann, R. Combining different stakeholders'  opinions in multi-criteria decision analyses applying partial ordermethodology. \textit{Standards} \textbf{2022}, 2, 35.

\bibitem{foodwate}Carlsen, L. Food Waste: The Good, the Bad, and (Maybe) the Ugly, \textit{Standards} 3(1):43-56, 2023.

\bibitem{galoiscluster} Carpineto C. and Giovanni R., “GALOIS: An Order-Theoretic Approach to Conceptual Clustering.” \textit{International Conference on Machine Learning} (1993).

 \bibitem{conceptret} Carpineto, C., \& Romano, G. (1996). A Lattice Conceptual Clustering System and Its Application to Browsing Retrieval. \textit{Machine Learning}, 24, 95-122. 

\bibitem{conceptbook}C. Carpineto, G. Romano, \textit{Concept Data Analysis}, Wiley, Chichester, 2004.

\bibitem{cognitionamyloid} Carr et al., Associating Cognition With Amyloid Status Using Partially Ordered Set Analysis. \textit{Front Neurol.} 2019;10:976. 

\bibitem{macrosocio}Cavalletti, B., Corsi, M. ``Beyond GDP” Effects on National Subjective Well-Being of OECD Countries. \textit{Soc Indic Res} 136, 931–966 (2018). \url{https://doi.org/10.1007/s11205-016-1477-0}

\bibitem{toplodata}F. Chanza \& B. Michel, l An Introduction to Topological Data Analysis: Fundamental and Practical Aspects for Data Scientists \url{Front. Artif. Intell.},  Vol. 4 (2021)\url{ https://doi.org/10.3389/frai.2021.667963}

\bibitem{surveyDUNN}Y. Chen, M. Mancini, X. Zhu and Z. Akata, "Semi-Supervised and Unsupervised Deep Visual Learning: A Survey," in \textit{IEEE Transactions on Pattern Analysis and Machine Intelligence}, vol. 46, no. 3, pp. 1327-1347, March 2024, doi: 10.1109/TPAMI.2022.3201576.

\bibitem{predrankABs}Cheng, W., Rademaker, M., De Baets, B., Hüllermeier, E. (2010). Predicting Partial Orders: Ranking with Abstention. In: Balcázar, J.L., Bonchi, F., Gionis, A., Sebag, M. (eds) Machine Learning and Knowledge Discovery in Databases. ECML PKDD 2010. Lecture Notes in Computer Science(), vol 6321. Springer, Berlin, Heidelberg. \url{ https://doi.org/10.1007/978-3-642-15880-3_20}.

\bibitem{predrankABs2}Cheng et al., Partial Orders: Ranking with Abstention, \url{https://www.weiweicheng.com/research/slidesposters/cheng-ecml10aslides.pdf}

\bibitem{riordanposets} G.-S Cheon, B. Curtis, G. Kwon and A.M. Mwafise, Riordan posets and associated incidence matrices, \textit{Linear Algebra Appl.}, 632: 308--331 (2022).

\bibitem{videodata} J. Chen, X. Song, Liqiang Nie, X. Wang, H. Zhang, and Tat-Seng Chua. 2016. Micro tells macro: predicting the popularity of micro-videos via a transductive model. In MM. 898–907.

\bibitem{fcalconcept}Cimiano, P., Hotho, A., \& Staab, S. (2005). Learning Concept Hierarchies from Text Corpora  using Formal Concept Analysis. \textit{Journal of Artificial Intelligence Research} 2005 DOI:\url{10.1613/jair.1648}

\bibitem{binaryclassrule}M Collery(2022), Learning binary classification rules for sequential data, \url{https://strl2022.github.io/files/short1.pdf}.

\bibitem{neuralseqclass}M. Collery et al.(2023), Neural-based classification rule learning for sequential data, arXiv:2302.11286.

\bibitem{humandev}Comim, F. A Poset-Generalizability Method for Human Development Indicators. \textit{Soc. Indic. Res.} \textit{2021}, 158, 1179–1198.

\bibitem{HassVis}M. Crampes, J. Oliveira-Kumar, S. Ranwez, J. Villerd. Visualizing Social Photos on a Hasse Diagram for Eliciting Relations and Indexing New Photos. IEEE Computer Graphics and Applications, 2009, 15 (6), pp.985-992. 

\bibitem{compProbs}V. Dankers, E. Bruni, and D. Hupkes. 2022. The Paradox of the Compositionality of Natural Language: A Neural Machine Translation Case Study. In\textit{ Proceedings of the 60th Annual Meeting of the Association for Computational Linguistics (Volume 1: Long Papers)}, pages 4154–4175, Dublin, Ireland. Association for Computational Linguistics.

\bibitem{delosrank}De Loof K, De Baets B, De Meyer H, Brüggemann R. A hitchhiker's guide to poset ranking. Comb Chem High Throughput Screen. 2008 ;11(9):734-44. doi: \url{10.2174/138620708786306032. PMID: 18991576}.

\bibitem{dellasocio} della Queva, S. (2017). Analysis of Social Participation: A Multidimensional Approach Based on the Theory of Partial Ordering. In M. Fattore, R. Bruggemann Partial Order Concepts in Applied Sciences, Springer.

\bibitem{dempsterOri}Dempster, A. P. (1967). Upper and lower probabilities induced by a multivalued mapping. \textit{The Annals of Mathematical Statistics}. 38 (2): 325–339. doi:\url{10.1214/aoms/1177698950}.

\bibitem{dempster}T.  Denoeux, M-H Masson (2010),  Dempster-Shafer Reasoning in large Partially Ordered Sets: Applications in Machine Learning, Advances in Intelligent and Soft Computing, book series (AINSC) volume 68. Springer, Berlin, Heidelberg.

\bibitem{gendergap}Di Brisco, A., \& Farina, P. (2018). Measuring gender gap from a poset perspective. \textit{Social Indicators Research}, 136, 1109–1124.

\bibitem{diazwhasse} Diaz et al., Predicting Proteome-Early Drug Induced Cardiac Toxicity Relationships (Pro-EDICToRs) with Node Overlapping Parameters (NOPs) of a new class of Blood Mass-Spectra graphs ,The 11th International Electronic Conference on Synthetic Organic Chemistry session Computational Chemistry \url{ https://sciforum.net/paper/view/1371}

\bibitem{daglatent}Diego Colombo, Marloes H Maathuis, Markus Kalisch, and Thomas S Richardson. Learning high-dimensional directed acyclic graphs with latent and selection variables. \textit{The Annals of Statistics}, pages 294–321, 2012.

\bibitem{longid}di Bella, E., Corsi, M., Leporatti, L. (2017). POSET Analysis of Panel Data with POSAC. In: Fattore, M., Bruggemann, R. (eds) Partial Order Concepts in Applied Sciences. Springer, Cham.\url{ https://doi.org/10.1007/978-3-319-45421-4_11}.

\bibitem{deeLSurvey} S. Dong, P. Wang, and K. Abbas. 2021. A survey on deep learning and its applications. \textit{Comput. Sci. Rev.} 40, C (May 2021).

\bibitem{seddata}B.J. Dutka, 1, K. Jones, 1, K.K. Kwan, 1, H. Bailey , 2, R. McInnis 1. Use of microbial and toxicant screening tests for priority site selection of degraded areas in water bodies, \textit{Water Research}, Vol. 22; 4, 1988, 503-510. \url{https://doi.org/10.1016/0043-1354(88)90047-4}

\bibitem{weigtposet}W. Dyer, Weighted Posets: Learning surface order from dependency trees.\textit{Proceedings of the 18th International Workshop on Treebanks and Linguistic Theories (TLT, SyntaxFest 2019) }\url{https://api.semanticscholar.org/CorpusID:204885369}

 \bibitem{gnnLTR1} U. Ergashev, E. C. Dragut, W. Meng, Learning To Rank Resources with GNN, WWW '23: Proceedings of the ACM Web Conference 2023, pp. 3247–3256, \url{https://doi.org/10.1145/3543507.3583360}

\bibitem{leveledposets}Fueyo, F., Abascal, P., Jim\'{e}nez, J., Palacio, A., Serrano, M.L., \& Tepav\v{c}evi\'{c}, A. (2025). Leveled partially ordered sets. \textit{Computational and Applied Mathematics}, 44, article 416. \url{https://doi.org/10.1007/s40314-025-03368-8}

\bibitem{fuzzypoverty} Fattore, M. (2008). Hasse diagrams, poset theory and fuzzy poverty measures. \textit{ Rivista Internazionale Di Scienze Sociali, Anno}, 116(1), 63–75.

\bibitem{fuzzydeprv} Fattore, M., Brueggemann, R., OWSIŃSKI, J., (2011). Using poset theory to compare fuzzy multidimensional material deprivation across regions. In: Ingrassia S., Rocci R. and Vichi, M. New Perspectives in Statistical Modeling and Data Analysis. Berlin: Springer-Verlag.

\bibitem{compindic}Fattore, M., Maggino, F. and Colombo, E. (2012) “From Composite Indicators to Partial Orders: evalu-ating socio-economic phenomena through ordinal data”. In: Maggino F, Nuvolati G (eds.) Quality of Life in Italy: research and reflections. Social Indicators Research Series, 48, 41–68.


\bibitem{visualsocio} Fattore, M., Arcagni, A., Barberis, S., (2014). Visualizing Partially Ordered Sets for Socioeconomic analysis. Revista Colombiana de Estadistica,  34(2), pp. 437–450.

\bibitem{parsec} Fattore, M., Arcagni, A., (2014). PARSEC: an R package for poset-based
evaluation of multidimensional poverty. In: Bruggemann R., Carlsen L. and Wittmann J. Multi-Indicator Systems and Modelling in Partial Order. Berlin: Springer.

\bibitem{ordbeing}Fattore M.; Maggino, F.; Arcagni, A.(2015)Exploiting Ordinal Data For Subjective Well-Being Evaluation \textit{The Measurement of Subjective Well-Being in Survey Research} Vol. 16, No. 3, pp. 409–428.

\bibitem{fattoresuffer} Fattore M., Maggino F., (2015). A new method for measuring and analyzing suffering - Comparing suffering patterns in Italian society. In Anderson R. E.
(Eds.). World Suffering and the Quality of Life. New York: Springer.

\bibitem{ordinaldeprive}Fattore, M. (2016). Partially ordered sets and the measurement of multidimensional ordinal deprivation. \textit{Social Indicators Research}, 128(2), 835–858.

\bibitem{fattoreWELL} Fattore M., Maggino F., Arcagni A. (2016) “Non-aggregative assessment ofsubjective well-being”, In G. Alleva, A. Giommi (eds.), Topics in Theoretical and Applied Statistics, Springer International Publishing Switzerland, 
\bibitem{appliedscience}M. Fattore, R. Bruggemann. \textit{Partial Order Concepts in Applied Sciences}, Springer, Cham, 2017, pp. 71–86. \url{DOI 10.1007/978-3-319-27274-0_20}.


\bibitem{FATTOREENV} Fattore M. (2018) “Non-aggregated indicators of environmental sustainability”, Silesian Statistical Review, 16(22), 7-22.

\bibitem{poprank} Fattore, M., \& Arcagni, A. (2018). F-FOD: Fuzzy first order dominance analysis and populations ranking over ordinal multi-indicators system. \textit{Social Indicators Research}, 144(1), 1–29.

\bibitem{FATTOREFIN} Fattore M., Zenga Ma. (2019) “New posetic tools for the evaluation of financialliteracy”, In Bianco A., Gnaldi M., Conigliaro. P. (eds.), Italian Studies on
Quality of Life, Springer, 978-3-030-06021-3.

\bibitem{probmatrices}Fattore, M., Arcagni, A., \& Maggino, F. (2019). Optimal scoring of partially ordered data,with an application to the ranking of smart cities. \url{https://research.uniroma1.it/pubblicazioni/49725}.

\bibitem{hypergraph}F. Feng, X. He, Y. Liu, L. Nie, and T-S Chua. 2018. Learning on Partial-Order Hypergraphs. In \textit{ WWW 2018: The 2018 Web Conference}, April 23–27, 2018, Lyon, France. ACM, New York, NY, USA 10 Pages.

\bibitem{fisherConcept}Fisher, D.H., \& Langley, P. (1985). Approaches to Conceptual Clustering. International Joint Conference on Artificial Intelligence.

\bibitem{fisherHCluster}H. Fisher. Knowledge Acquisition Via Incremental Conceptual Clustering. \textit{Machine Learning} 2:139--172, 1987. 

\bibitem{finnsemantics} Finn, V.K., and others (1982). Many valued 1ogics as fragments formalize semantics. Acta phi- losophica Fennica, vol.35.



\bibitem{finncomputer}V. K. Finn, M. I. Zabezhailo and O. M. Anshakov On a Computer-Oriented Formalization of Plausible Reasoning In F. Bacon-J. S. Mill'S Style (Main Principles and Computer Experiments) IFAC Artificial Intelligence. Leningrad. USSR 1983.

. \bibitem{finnjsm}Finn, V.K.: Plausible Reasoning in Systems of JSM Type. Itogi Nauki i Tekhniki,  Seriya Informatika 15, 54–101 (1991).

\bibitem{ontsim}A.  Formica, Ontology-based concept similarity in Formal Concept Analysis, Information Sciences, Volume 176, Issue 18, 2006, Pages 2624-2641, https://doi.org/10.1016/j.ins.2005.11.014.

\bibitem{defmda}Francés, O.; Abreu-Salas, J.; Fernández, J.; Gutiérrez, Y.; Palomar, M. Multidimensional Data Analysis for Enhancing In-Depth Knowledge on the Characteristics of Science and Technology Parks. Appl. Sci. 2023, 13, 12595. \url{https://doi.org/10.3390/app132312595}

\bibitem{rankprob} Lerche, D. and P. Sorensen (2003) Evaluation of the ranking probabilities for partial orders based on random linear extensions. Chemosphere 53, 981-992.

\bibitem{FCABook}Ganter, B., Wille, R.: Formal Concept Analysis, Mathematical Foundations. Springer-Verlag (1999).

\bibitem{gnnLTR2}H. Gao et al. “Graph-augmented Learning to Rank for Querying Large-scale Knowledge Graph.” ArXiv abs/2111.10541 (2021): 

\bibitem{fcAgarrica}Garriga, G.C. (2011). Formal Concept Analysis. In: Sammut, C., Webb, G.I. (eds) Encyclopedia of Machine Learning. Springer, Boston, MA.

\bibitem{gennariH}Gennari, J., Langley, P., \& Fisher, D. (1989). Models of incremental concept formation. Artificial Intelligence,
40:12–61.

\bibitem{machinehighcompute}P. Gepner, Machine Learning and High-Performance Computing Hybrid Systems, a New Way of Performance Acceleration in Engineering and Scientific Applications, \textit{Proceedings of the 16th Conference on Computer Science and Intelligence Systems,ACSIS}, Vol. 25, pages 27--36 (2021).

 \bibitem{incgaloisFO}R. Godin, R. Missaoui, H. Alaoui. Incremental concept  formation algorithms based on   Galois(concept) Lattice,  \textit{Computational IntelligenceVolume} 11: 2 ,pp. 246--267.

 \bibitem{latticehypothesis} Goncalves, B. and Porto, F. Research lattices: towards a scientific hypothesis data model. In \textit{Conference on Scientific and Statistical Database Management, SSDBM ’13, Baltimore, MD, USA, July 29 - 31, 2013.} 

\bibitem{comparse}E. Goodwin et al., Compositional Generalization in Dependency Parsing. arXiv:2110.06843 (cs).

\bibitem{gratzerlattice}G. Grätzer, General Lattice Theory: Academic Press, Inc., New York. 1978.

\bibitem{decodecomp}Y. Guo, Z. Lin, J-G Lou, D. Zhang, Hierarchical Poset Decoding for Compositional Generalization in Language, $34^{th}$ conference on Neural Processing Systems(NeurlPS 2020) Vancouver Canada.

\bibitem{rankdnn} Guo et al. RankDNN: Learning to Rank for Few-Shot Learning,(2023) The Thirty-Seventh AAAI Conference on Artificial Intelligence (AAAI-23)

\bibitem{posetcocalculus}Hem, B.G. (2025). Poset functor cocalculus and applications to topological data analysis. arXiv:2501.05996.

\bibitem{multiperscocalc}Hem, B.G. (2025). Decomposing multipersistence modules using functor calculus. arXiv:2510.06178.

\bibitem{hadHcluster}Hadzikadic, M., \& Yun, D. (1989). Concept formation by incremental conceptual clustering. \textit{Proceedings of
the Eleventh International Joint Conference on Artificial Intelligence} (pp. 831–836). Detroit, MI: Morgan Kaufmann.

\bibitem{fcaonto3}M. R. Hacene, A. Napoli, P. Valtchev, Y. Toussaint and R. Bendaoud, "Ontology Learning from Text Using Relational Concept Analysis," 2008 International MCETECH Conference on e-Technologies (mcetech 2008), Montreal, QC, Canada, 2008, pp. 154-163, doi: 10.1109/MCETECH.2008.29.

\bibitem{HOA}M. Hajij et al.,  Higher-Order Attention Networks (2022), \url{https://arxiv.org/abs/2206.00606v1}.

\bibitem{TDL}M. Hajij et al., Topological Deep Learning: Going Beyond Graph Data (2023), arXiv: arXiv:2206.00606v3.

\bibitem{rankhazard}E. Halfon, M.G. Reggiani, On ranking chemicals for environmental hazard \textit{ Environ. Sci. Technol.} 1986, 20, 11, 1173–1179.

\bibitem{hassesoftw}Halfon, E. (2006). Hasse Diagrams and Software Development. In: Brüggemann, R., Carlsen, L. (eds) Partial Order in Environmental Sciences and Chemistry. Springer, Berlin, Heidelberg.

\bibitem{hansonHcluster}Hanson, S., \& Bauer, M. (1989). Conceptual clustering, categorization, and polymorphy. \textit{Machine Learning},
3:343–372.

\bibitem{fcaonto1}B.  A. Hassan, Ontology Learning Using Formal Concept Analysis and WordNet, arXiv:2311.14699, 2023.


\bibitem{govt}Hilckmann, A., Bach, V., Bruggemann, R., Ackermann, R., \& Finkbeiner, M. (2017). Partial order analysis of the government dependence of the sustainable development performance in Germany’s federal states. In M. Fattore \& R. Bruggemann (Eds.), (2017)\textit{ Partial Order Concepts in Applied Sciences}. Springer.


\bibitem{dimensionsDEF}S. Hira, P.S. Deshpande, Data Analysis using Multidimensional Modeling, Statistical Analysis and Data Mining on Agriculture Parameters, Procedia Computer Science, Vol. 54, 2015, pp. 431-439, \url{https://doi.org/10.1016/j.procs.2015.06.050}.

\bibitem{posetanalSD}T.  Hirai a, F.  Comim, Measuring the sustainable development goals: A poset analysis, \textit{Ecological Indicators}, Vol. 145, 2022.

\bibitem{datasetrelation}Huchard, M., Rouane-Hacène, M., Roume, C., Valtchev, P.: Relational concept discovery in structured datasets. Ann. Math. Artif. Intell. \textbf{49}(1-4) (2007) 39-76.

\bibitem{fcAIganov} D. I. Ignatov, Introduction to Formal Concept Analysis and Its Applications in Information Retrieval and Related Fields(2017), arXiv:1703.02819

\bibitem{fcaclassLang}Ikeda, M., \& Yamamoto, A. (2013). Classification by Selecting Plausible Formal Concepts in a Concept Lattice.

\bibitem{beliefitems}Ip, E., Chen, S.-H., \& Quandt, S. (2016). Analysis of multiple partially ordered responses to belief items with don’t know option.  \textit{Psychometrika}, 81(2), 483–505.

\bibitem{quantposet}T. Ivanciuc, O. Ivanciuc, D. J. Klein, Posetic Quantitative Superstructure/Activity Relationships (QSSARs) for Chlorobenzenes,  J. Chem. Inf. Model. 2005, 45, 4, 870–879. \url{ https://doi.org/10.1021/ci0501342}

\bibitem{urbandepr}Ivaldi, E., Ciacci, A., \& Soliani, R. (2020). Urban deprivation in Argentina: A poset analysis.  \textit{Papers in Regional Science}, 99(6), 1723-1747.

\bibitem{ADDmethod}S. D. Langhans, P. Reichert, N. Schuwirth, The method matters: A guide for indicator aggregation in ecological assessments, Ecological Indicators, Volume 45,
2014, pp. 494-507, \url{ https://doi.org/10.1016/j.ecolind.2014.05.014}.

\bibitem{gafca} S. Jabin, Machine Learning methods and applications  using Formal Concept Analysis, International Journal of New Technologies in Science and Engineering  Vol. 2: 3, 2015, 

\bibitem{neurocognitivep}Jaeger J, Tatsuoka C, Berns S, Varadi F, Czobor P, Uzelac S: Associating functional recovery with neurocognitive profiles identified using partially
ordered classification models. \textit{Schizophr Res} 2006, 85:40-48.

\bibitem{neuroshizo}Jaeger J, Tatsuoka C, Berns SM, Varadi F: Distinguishing neurocognitive functions in schizophrenia using partially ordered classification models. \textit{Schizophr Bull} 2006, 32:679-691

\bibitem{semiflatcluster}M.F. Janowitz., Semi-Flat Cluster Methods,  Discrete Mathematics 21 (1978) , 47-60.

\bibitem{clusteroder}M.F. Janowitz., An order theoretic model for cluster analysis.  \textit{SIAM Journal on Applied Mathematics},34:55–72, 1978.

\bibitem{clusterreport} Janowitz, M.F. (2007). Cluster Analysis Based on Posets. In: Brito, P., Cucumel, G., Bertrand, P., de Carvalho, F. (eds) Selected Contributions in Data Analysis and Classification. Studies in Classification, Data Analysis, and Knowledge Organization. Springer, Berlin, Heidelberg.

\bibitem{clusterrelate}M. F. Janowitz:  Ordinal and Relational Clustering. Interdisciplinary Mathematical Sciences 10, World Scientific 2010, pp. 1-200.

\bibitem{jardinetda}R. Jardine, Posets, Metric Spaces, Topological Data Analysis(2020),\url{ http://www.sci.brooklyn.cuny.edu/~noson/CTslides/Jardine20.pdf}

\bibitem{persistposet} W. Kim, F. Memoli, Persistence Over Posets, Notices of American Mathemtical Society, Vol 20, No. 8, (2023).

\bibitem{fcaonto2} Jia, H., Newman, J., Tianfield, H. (2009). A new Formal Concept Analysis based learning approach to Ontology building. In: Sicilia, MA., Lytras, M.D. (eds) Metadata and Semantics. Springer, Boston, MA. \url{https://doi.org/10.1007/978-0-387-77745-0_42}.

\bibitem{joynyergraphClu}I. Jonyer, L. B. Holder, and D. J. Cook. Graph-Based Hierarchical Conceptual Clustering. \textit{Proceedings of the Florida Artificial Intelligence Research Symposium}, pp. 91-95, 2000. 

\bibitem{joynyergraphClu2}I. Jonyer, D. J. Cook , L. B. Holder, Graph-Based Hierarchical Conceptual Clustering, \textit{Journal of Machine Learning Research} 2 (2001) 19-43.

\bibitem{spatial}Kainz, W., Egenhofer, M. J., and Greasley, I. (1993). Modelling spatial relations and operations with partially ordered sets.  \textit{International Journal of Geographical Information Systems}, 7(3):215–229.

\bibitem{hassExp} H-M.  Kaltenbach, Teaching Design of Experiments using Hasse diagrams arXiv:1912.08567v1

\bibitem{fcabagging}Kang, X., Li, D., \& Wang, S. (2011). A multi-instance ensemble learning model based on concept lattice. \textit{Knowl. Based Syst.}, 24, 1203-1213.

\bibitem{envshallowlake}S. Kardaetz, T. Strube, R. Bruggemann, G. Nützmann, Ecological scenarios analyzed and evaluated by a shallow lake model, \textit{J. Environ. Manag}. \textbf{88} (2008) 120–135.

\bibitem{deepunsupervisedlearn}J. Karhunen, T.  Raiko, K.  Cho, Unsupervised deep learning: A short review, \textit{Advances in Independent Component Analysis and Learning Machines}, Academic Press, 2015, Pages 125-142,\url{ https://doi.org/10.1016/B978-0-12-802806-3.00007-5}.

\bibitem{fcareccO}Y. Kashnitsky and D. I. Ignatov, “Can FCA-based Recommender System Suggest a Proper Classi er ? 2 Multiple Classi er Systems,” \textit{FCA do Artif. Intell.} 2014), p. 17, 2014.

\bibitem{kendalltau} M. Kendall, A new measure of rank correlation, Biometrika, Volume 30, Issue 1-2, June 1938, Pages 81–93 (doi:\url{10.1093/biomet/30.1-2.81})

\bibitem{hassLearntics} Kickmeier-Rust, M.D., Albert, D. (2013). Using Hasse Diagrams for Competence-Oriented Learning Analytics. In: Holzinger, A., Pasi, G. (eds) Human-Computer Interaction and Knowledge Discovery in Complex, Unstructured, Big Data. HCI-KDD 2013. Lecture Notes in Computer Science, vol 7947. Springer, Berlin, Heidelberg. \url{https://doi.org/10.1007/978-3-642-39146-0_6}

\bibitem{persistence} Kim, W., Memoli, F.: Persistence over posets. Notices of theAmerican Mathematical Society 2761, 1214–1224 (September 2023).

\bibitem{gcn} T. N Kipf and M. Welling. 2017. Semi-supervised classification with graph convolutional networks. ICLR (2017).

\bibitem{chemsimilar}Klein, D. J. Similarity and Dissimilarity in Posets. \textit{J. Math. Chem.} \textbf{1995}, 18, 321-348.

\bibitem{chemposets}Klein, D. J.; Babic, D. Partial Orderings in Chemistry. \textit{J. Chem. Inf. Comput. Sc.} \textbf{1997}, 37, 656--671.

\bibitem{chem}D. J. Klein, Prolegomenon on partial orderings in chemistry, \textit{MATCH Commun. Math.Comput. Chem.} \textbf{42} (2000) 1–290.

\bibitem{reactiondiag}Klein, D. J., \& Ivanciuc, T. (2006). Directed reaction graphs as posets. In R. Bruggemann \& L. Carlsen (Eds.), Partial order in environmental sciences and chemistry (pp. 35–57). Berlin: Springer.

\bibitem{fcaimage}Khatri, Minal, "Formal Concept Analysis for Image Classification and Machine Learning Models for Anti-CRISPR Protein Discovery in Bioinformatics" (2023). Dissertations and Doctoral Documents from University of Nebraska-Lincoln, 2024–. 47.

\bibitem{daggip}Kotsianti, S.B., Kanellopoulos, D. (2007). Combining Bagging, Boosting and Dagging for Classification Problems. In: Apolloni, B., Howlett, R.J., Jain, L. (eds) Knowledge-Based Intelligent Information and Engineering Systems. KES 2007. Lecture Notes in Computer Science(), vol 4693. Springer, Berlin, Heidelberg. \url{https://doi.org/10.1007/978-3-540-74827-4_62}

\bibitem{pyhasse3} V. Kristina Et Al. , ``Features of PyHasse software used for the evaluation of chemicals in human breast milk samples," \textit{Simulation in Umwelt- und Geowissenschaften} , vol.141, Hamburg, Germany, pp.169-180, 2012.

\bibitem{semanticdep}R. Kurtz(2020), Contributions to Semantic Dependency Parsing.
\url{https://www.diva-portal.org/smash/get/diva2:1457198/FULLTEXT02}.

\bibitem{MathConcepts}S. O. Kuznetsov, Mathematical aspects of concept analysis, \textit{Journal of Mathematical Sciences} 80(2),(1996).

\bibitem{mlauto} Kuznetsov, S.O. Machine Learning on the Basis of Formal Concept Analysis. Automation and Remote Control 62, 1543–1564 (2001). \url{https://doi.org/10.1023/A:1012435612567}

\bibitem{fcaML}Kuznetsov, S.O, Machine Learning and Formal Concept Analysis Conference: Concept Lattices, Second International Conference on Formal Concept Analysis, ICFCA 2004, Sydney, Australia, February 23-26, 2004.

\bibitem{orderdataanalysis} S. Kuznetsov, Ordered Sets for Data Analysis(2019), arXiv:1908.11341 .

\bibitem{rankposet}G. Lebanon, J. Lafferty, Conditional models on the ranking poset, NIPS'02: Proceedings of the 15th International Conference on Neural Information Processing Systems, January 2002, pp. 431--438.

\bibitem{lebowitzHcluster}Lebowitz, M. (1986). Concept learning in a rich input domain: generalization-based memory. In R.S. Michalski,
J.G. Carbonell, \& T.M. Mitchell (Eds.), \textit{Machine Learning: An Artificial Intelligence Approach} (Vol. 2). Morgan Kaufmann, San Mateo, CA.

 \bibitem{mdaorigins}C. Lei et al., The Application of Multidimensional Data Analysis in the EIA Database of Electric Industry, 2011 3rd International Conference on Environmental Science and Information Application Technology (ESIAT 2011), In:  \textit{Procedia Environmental Sciences} 10 ( 2011 ) 1210 – 1215.

\bibitem{crime}Levy, S. (1985). Partial order analysis of crime indicators. \textit{Social Indicators Research}, 16(2), 195–199.

\bibitem{linearextensions} Linear Extensions \url{https://mathworld.wolfram.com/LinearExtension.html}.

\bibitem{clusterpair}J. Liu, Q. Zhang, W. Wang, L. McMillan, J. Prins, Clustering pair-wise dissimilarity data into partially  ordered sets. In Proceedings of the Twelfth ACM SIGKDD International Conference on Knowledge Discovery and Data Mining. (2006) 637--642. 

\bibitem{compgen} Liu, Q., An, S., Lou, J.-G., Chen, B., Lin, Z., Gao, Y., Zhou, B., Zheng, N., and Zhang, D. Compositional generalization by learning analytical expressions. In Proceedings of the 34th International Conference on Neural Information Processing Systems, NIPS’20, Red Hook, NY, USA,  2020. 

\bibitem{marektHcluster}Maarek, Y.S. (1990). An Incremental Conceptual Clustering Algorithm that Reduces Input-Ordering Bias. In: Golumbic, M.C. (eds) Advances in Artificial Intelligence. Springer, New York, NY. \url{https://doi.org/10.1007/978-1-4613-9052-7_7}

\bibitem{seqdata} H. Mannila, C. Meek, Global partial orders from sequential data, KDD'00: Proceedings of the sixth ACM SIGKDD international conference on knowledge discovery and data mining, August 2000, pages  161--168.

\bibitem{latanalytics}L. Markowsky and G. Markowsky, “Lattice Data Analytics and an Exploratory Analysis of the Carver2 Dataset,” to appear in IDAACS 2019.

\bibitem{markovHcluster} Markov, A lattice-based approach to hierarchical clustering.\textit{ Proceedings of the Florida Artificial Intelligence Research Symposium}, pp. 389--393, 2001. 

\bibitem{martHcluster}Martin, J., \& Billman, D. (1994). Acquiring and combining overlapping concepts. \textit{Machine Learning}, 16(1–
2):121–155.

\bibitem{macHcluster}McKusick, K., \& Langley, P. (1991). Constraints on tree structure in concept formation. \textit{Proceedings of the
Thirteenth International Joint Conference on Artificial Intelligence} (pp. 810–816). Sydney, Australia: Morgan
Kaufmann.

\bibitem{fcaboosting}Meddouri, N., Maddouri, M. (2009). Boosting Formal Concepts to Discover Classification Rules. In: Chien, BC., Hong, TP., Chen, SM., Ali, M. (eds) Next-Generation Applied Intelligence. IEA/AIE 2009. Lecture Notes in Computer Science(), vol 5579. Springer, Berlin, Heidelberg. \url{https://doi.org/10.1007/978-3-642-02568-6_51}.

\bibitem{fcadagging} N. Meddouri , H. Khoufi , M. Maddouri, Parallel Learning and Classification for Rules based on Formal Concepts \textit{Procedia Computer Science
Volume} 35, 2014, pp.  358-367.

\bibitem{molepred}  J-P. M$\acute{e}$tivier et al., Discovering Structural Alerts for Mutagenicity Using Stable Emerging Molecular Patterns, J. Chem. Inf. Model. 2015, 55, 5, 925–940, 2015 \url{https://doi.org/10.1021/ci500611v}.

\bibitem{mikhalskiconcept}R.S. Michalski, Knowledge Acquisition Through Conceptual Clustering: A Theoretical Framework and an Algorithm for Partitioning Data into Conjunctive Concepts.\textit{ International Journal for Policy Analysis and Information Systems}. Vol. 4, No. 3, 1980.

\bibitem{michalskiHcluster} Michalski and R. E. Stepp. Learning From Observation: Conceptual Clustering. In R.S.Michalski, J.G. Carbonell, and T.M. Mitchell (Eds.), Machine Learning: An Artificial Intelligence Approach, Volume 1, Tioga Publishing Company, pp. 331-363, 1983. 

\bibitem{visualattentionmodel}V. Mnih, N. Heess, A. Graves, and k. kavukcuoglu, “Recurrent models of visual attention,” in \textit{27th Annual Conference on Neural
Information Processing Systems }(NIPS 2014). Curran Associates,
Inc., 2014, pp. 2204–2212.


\bibitem{operadposet}A. M. Mwafise, G-S Cheon, H. J. Choi, S. Giraudo, Operad Structure of Poset Matrices (2024), arXiv:2401.06814.

\bibitem{MollicaR}Mollica, C., \& Tardella, L. (2020). PLMIX: an R package for modelling and clustering partially ranked data. Journal of Statistical Computation and Simulation, 90(5), 925–959. \url{https://doi.org/10.1080/00949655.2020.1711909}.

\bibitem{ontolearnD}M. Nabeel et al., A survey of ontology learning techniques and applications, Database, Volume 2018, 2018, bay101, \textit{https://doi.org/10.1093/database/bay101}

\bibitem{compositeINDIC}Nardo, M., M. Saisana, A. Saltelli and S. Tarantola: 2005b, Tools for Composite IndicatorsBuilding. European Commission, report EUR 21682 EN (Joint Research Centre, Ispra,Italy).

 \bibitem{bridgeposets}Newlin, J.T., Patil, G.P. Application of partial order to stream channel assessment at bridge infrastructure for mitigation management.  \textit{Environ Ecol Stat} 17, 437–454 (2010). \url{ https://doi.org/10.1007/s10651-010-0162-8}.

\bibitem{bayesianpartialorder}Nicholls GK, Lee JE, Karn N, Johnson D, Huang R, Muir-Watt A, Bayesian inference for partial orders from random linear extensions: power relations from 12th Century Royal Acta, , 2022. arXiv: 2212.05524. 


\bibitem{posetdistfam}Olave, A.A. (2026). A new family of distances over partially ordered sets. arXiv:2606.06377.

\bibitem{abstractdynprog}Peng, C., Stachurski, J., \& Yang, J. (2026). Abstract Dynamic Programming on Partially Ordered Spaces. arXiv:2606.12777.

\bibitem{milestone}Sadr, A., \& Alidoost Nia, M. (2026). MileStone: A Multi-Objective Compiler Phase Ordering Framework for Graph-based IR-Level Optimization. arXiv:2605.23435.

\bibitem{dposets}Sargent, T.J., \& Stachurski, J. (2025). Dynamic Programs on Partially Ordered Sets. \textit{SIAM Journal on Control and Optimization}. arXiv:2308.02148.

\bibitem{modelgraphdist}Taeb, A., Guo, F.R., \& Henckel, L. (2026). Model-oriented Graph Distances via Partially Ordered Sets. arXiv:2511.10625.

\bibitem{posafenet}Wong, K., Xiao, W., \& Rus, D. (2026). PoSafeNet: Safe Learning with Poset-Structured Neural Nets. arXiv:2601.22356.

\bibitem{poolingorder}Dolores-Cuenca, E., Guzm\'{a}n-S\'{a}enz, A., Kim, S., L\'{o}pez-Moreno, S., \& Mendoza-Cortes, J. (2025). Order Theory in the Context of Machine Learning. arXiv:2412.06097.

\bibitem{jdmscmining}P.C. Golar et al. Mining large-scale data using AI-integrated lattice and poset structures (2026). \textit{Journal of Discrete Mathematical Sciences and Cryptography}, 29(2-A), 671--680. \url{https://doi.org/10.47974/JDMSC-2510}.

\bibitem{fcaclassbiclus}Onishchenko, A.A., Gurov, S.I. Classification based on formal concept analysis and biclustering: possibilities of the approach. Comput Math Model 23, 329–336 (2012). \url{https://doi.org/10.1007/s10598-012-9141-2}.

\bibitem{fcaonto4}  Ouyang, C.; Liu, Y. , Formal Concept Analysis Supporting Ontology Learning From Database,  Advanced Science Letters, Volume 7 2012, pp. 473-477(5).

\bibitem{reinLTR}V. Padhye, K. Lakshmanan, A deep actor critic reinforcement learning framework for learning to rank, Neurocomputing, Volume 547, (2023.) \url{https://doi.org/10.1016/j.neucom}.

\bibitem{HHmethod}Pakkar, M.S. A Hierarchical Aggregation Approach for Indicators Based on Data Envelopment Analysis and Analytic Hierarchy Process. Systems 2016, 4, 6. https://doi.org/10.3390/systems4010006

\bibitem{envranking} G. P. Patil, C. Taillie, Multiple indicators, partially ordered sets, and linear extensions: Multi-criterion ranking and prioritization, \textit{Environ. Ecol. Stat.} \textbf{11} (2004) 199–228.

\bibitem{multivariate} G. Patil, W. L. Myers, R. Bruggemann, Multivariate datasets for inference of order: some considerations and explorations, in: R. Bruggemann, L. Carlsen, J. Wittmann (Eds.), \textit{Multi-Indicator Systems and Modelling in Partial Order}, Springer, New York, 2014,pp.13–45.

\bibitem{montecarlo}L. Pellegrina, C. Cousins, F. Vandin, M. Riondato. MCRapper: Monte-Carlo Rademacher Averages for Poset Families and Approximate Pattern Mining. Extended version. \url{https://arxiv.org/abs/2006.09085}.

\bibitem{mdaHtech}Popescu-Spineni, S. (1998). Hierarchy Techniques of Multidimensional Data Analysis (MDA) in Social Medicine Research. In: Rizzi, A., Vichi, M., Bock, HH. (eds) Advances in Data Science and Classification. Studies in Classification, Data Analysis, and Knowledge Organization. Springer, Berlin, Heidelberg. \url{https://doi.org/10.1007/978-3-642-72253-0_87}

\bibitem{proclasmodels}O. Prokasheva, A. Onishchenko, S. Gurov, Classification Methods Based on Formal Concept Analysis, Conference: FCAIR 2013 – Formal Concept Analysis Meets Information Retrieval. Workshop co-located with the 35th European Conference on Information Retrieval (ECIR 2013)At: Moscow.

\bibitem{softposrank} S. Pudenz, ProRank – software for partial order ranking, \textit{MATCH Commun. Math. Comput. Chem.} \textbf{54} (2005) 611–622.

\bibitem{mlbigdata}Qiu, J., Wu, Q., Ding, G. et al. A survey of machine learning for big data processing. \textit{EURASIP J. Adv. Signal Process.} 2016, 67 (2016). \url{https://doi.org/10.1186/s13634-016-0355-x}.

\bibitem{singlecomplete}N.R Rashidah, A Comparison Between Single Linkage and Complete Linkage in Agglomerative Hierarchical Cluster Analysis for Identifying Tourists Segments, IIUM Engineering Journal, Vol. 12, No. 6, 2011: Special Issue in Science and Ethics.

\bibitem{crime2}A. Raveh and S. F. Landau, Partial Order Scalogram Analysis with Base Coordinates (POSAC): Its Application to Crime Patterns in All the States in the United States,  \textit{Journal of Quantitative Criminology} Vol. 9, No. 1.,  pp. 83-99 (1993).

\bibitem{HCAHDT}G. Restrepo \&  R. Brüggemann, Ranking regions using cluster analysis, Hasse diagram technique and topology , $3^{rd}$ International Congress on Environmental Modelling and Software, Burlington, Vermont, USA 2006.

\bibitem{chemrankpatterns}G. Restrepo, R. Bruggemann, M. Weckert, S. Gerstmann, H. Frank, Ranking patterns, an application to refrigerants, \textit{MATCH Commun. Math. Comput. Chem.} \textbf{59} (2008) 555–584.

\bibitem{Rimodifragile} Rimoldi S.M.L., Arcagni A., Fattore M., Barbiano di Belgiojoso E.. (2020)
“Targeting policies for multidimensional poverty and social fragility relief
among migrants in Italy, using F-FOD analysis”, Social Indicators Research
- \url{doi 10.1007/s11205-020-02485-7}

\bibitem{RIMOLDIMUN} Rimoldi S.M.L., Arcagni A., Fattore M., Terzera T. (2020) “Social and
material vulnerability of the Italian municipalities: comparing alternative ap-
proaches”, Social Indicators Research- \url{doi: 10.1007/s11205-020-02330-x}.

\bibitem{ensembLE}Sagi O, Rokach L. 2018. \textit{Ensemble learning: a survey}. Wiley Interdisciplinary Reviews: Data Mining and Knowledge Discovery 8(4):e1249

\bibitem{posetcombin}G.-C. Rota, On the foundations of combinatorial theory I.  Theory of Mobius functions, Z. Wahrsch. Verw. Gebiete 2 (2964) 340-368. 

\bibitem{fcaontology}Rouane-Hacene, M., Valtchev, P., Nkambou, R.: Supporting Ontology Design through Large-Scale FCA-Based Ontology Restructuring. In: ICCS 2011. (2011) 257--269.

\bibitem{gammarank}M. D. Ruiz, E. Hüllermeier, A formal and empirical analysis of the fuzzy gamma rank correlation coefficient, Information Sciences,Vol. 206, 2012, pp. 1-17, \url{ https://doi.org/10.1016/j.ins.2012.04.006} .

\bibitem{RiskeyLinearEXt}F. Ruskey, Generating linear extensions of posets by transpositions, Journal of Combinatorial Theory, Series B, Volume 54, Issue 1, 1992, Pages 77-101,\url{ https://doi.org/10.1016/0095-8956(92)90067-8}.


\bibitem{clusteragglo}I.M. Sabara, F.  Rozi, M. N.  Jauhari, Agglomerative Hierarchical Clustering Analysis Based on Partially-Ordered Hasse Graph of Poverty Indicators in East Java, 
Proceedings of the 12th International Conference on Green Technology (ICGT 2022). 

\bibitem{proteindata}K. Sachs, O. Perez, D. Lauffenburger, G. Nolan, Causal protein-signaling networks derived from multiparameter single-cell data. Science. \url{ https://www.science.org/doi/10.1126/science.1105809}\#supplementary-materials.

\bibitem{fcamlsoft} Salman, H.E. Leveraging a combination of machine learning and formal concept analysis to locate the implementation of features in software variants 2023, Information and Software Technology, vol. 164.

\bibitem{classrule}Sahami, M. (1995). Learning classification rules using lattices (Extended abstract). In: Lavrac, N., Wrobel, S. (eds) Machine Learning: ECML-95. ECML 1995. Lecture Notes in Computer Science, vol 912. Springer, Berlin, Heidelberg.\url{ https://doi.org/10.1007/3-540-59286-5_83}.

\bibitem{fcadlexplain}Sangroya, A., Anantaram, C., Rawat, M., \& Rastogi, M. (2019). Using Formal Concept Analysis to Explain Black Box Deep Learning Classification Models. \textit{FCA4AI@IJCAI.}

\bibitem{anmsurvey}A. Santana, E.  Colombini, Neural Attention Models in Deep Learning: Survey and Taxonomy(2021), arXiv:2112.05909v1.

\bibitem{posetcolorspace}Semeraldi et al., Partial Order Rank Features in Colour Space,Appl. Sci. 2020, 10(2), 499; \url{https://doi.org/10.3390/app10020499}

\bibitem{interagrr} Sen, A. (1970b). Interpersonal aggregation and partial comparability. \textit{Econometrica}, 38(3), 393–409.

\bibitem{trendmiss} Seymour, R. B. (1960). Missing Data in Non-Linear Trend Analysis of Repeated Measurements on the Same Individuals. The Journal of Educational Research, 54(4), 141–144. \url{http://www.jstor.org/stable/27530403}.

\bibitem{shafer} Shafer, G.: \textit{A mathematical theory of evidence}. Princeton University Press, Princeton (1976).

\bibitem{medstats} Silan M, Boccuzzo G, Arpino B. Matching on poset‐based average rank for multiple treatments to compare many unbalanced groups. \textit{Statistics in Medicine}. 2021;40(28):6443–6458. doi: \textit{10.1002/sim.9192}.

\bibitem{decisionsup}U. Simon, R. Bruggemann, S. Mey, S. Pudenz, METEOR- application of a decision support tool based on discrete mathematics, \textit{MATCH Commun. Math. Comput. Chem.} \textbf{54} (2005) 623–642.

\bibitem{decisionsup2}Simon, U.; Brüggemann, R.; Behrendt, H.; Shulenberger, E.; Pudenz, S. METEOR: a step-by-step procedure to explore effects of indicator aggregation in multi criteria decision aiding – application to water management in Berlin, Germany. Acta hydrochim. Hydrobiol. 2006, 34, 126-136. 

\bibitem{graphsemilearn} Song et al.,  Graph-based Semi-supervised Learning: A Comprehensive Review 2021.

\bibitem{rankparameter}Sorensen, P. B.; Mogensen, B. B.; Carlsen, L.; Thomsen, M. The
Influence on Partial Order Ranking from Input Parameter Uncertainty Definition of a Robustness Parameter. \textit{Chemosphere }\textbf{2000}, 595--600.

\bibitem{envchempesticide}P. B. Sorensen, R. Bruggemann, L. Carlsen, B. B. Mogensen, J. Kreuger, S. Pudenz, Analysis of monitoring data of pesticide residues in surface waters using partial order ranking theory, \textit{Envir. Tox. Chem.} \textbf{22} (2003) 661–670.

\bibitem{posoftwaRe}Sorensen, P. B. ; Brüggemann; M. Thomsen; D. B. Lerche,  Application of multidimensional rank-correlation. MATCH Communications in Mathematical and in Computer Chemistry, 54(3), 643-670. 

\bibitem{loglinearposets} M. Sugiyama, Machine Learning and Information Geometry II, \url{https://mahito.info/files/Sugiyama_NII_IISS_2018_02.pdf}.

\bibitem{multipleindicator}J.L. Sullivan \&  S. Feldman  Multiple Indicators - An Introduction (1979), \url{https://www.ojp.gov/ncjrs/virtual-library/abstracts/multiple-indicators-introduction}.

 \bibitem{modelselection} Taeb,  A. ; Buhlmann, P. ; Chandrasekaran, V.  Model Selection over Partially Ordered Sets(2024), ArXiV: 2308.10375 .

\bibitem{posetrlgen} Mwafise, A.M. (2026). Order-Agnostic Generation of Lattices via Reinforcement Learning. \textit{ScienceOpen Preprints}. DOI: 10.14293/PR2199.004125.v2. 

\bibitem{posetisodecomp} Mwafise, A.M. (2026). Matrix Decomposition Algorithms for Accelerated Poset Isomorphism. \textit{ScienceOpen Preprints}. DOI: 10.14293/PR2199.003522.v2. 

\bibitem{operadposetmatrices} Cheon, G.-S., Choi, H.J., Giraudo, S., \& Mwafise, A.M. (2026). Operads of Poset Matrices. \textit{The Electronic Journal of Combinatorics}, 33(2), \#P2.37. DOI: 10.37236/14396.

\bibitem{sequencephd}Tatsuoka C: Sequential classification on partially ordered sets. PhD thesis Cornell University, Statistics Department; 1996.

\bibitem{dataAnalytic}Tatsuoka C. Data analytic methods for latent partially ordered classification models. \textit{ Applied Statistics (Journal of the Royal Statistical Society Series C}). 2002;51:337–350.

\bibitem{sequence}Tatsuoka C, Ferguson T: Sequential classification on partially ordered sets. \textit{Journal of the Royal Statistical Society}, Series B 2003, \textit{65}:143-157.

\bibitem{sequence2} Tatsuoka C, Corrigendum:Data analytic methods for latent partiallyordered classification models, Appl.Statist.(2005) \textbf{54},Part2,pp.465–467

\bibitem{diagnostics}Tatsuoka C. Diagnostic models as partially ordered sets. \textit{Measurement}. (2009) 7:49–53.

\bibitem{latentclassif}Tatsuoka C, Varadi F, Jaeger J. Latent partially ordered classification  models and normal mixtures. \textit{J Edu Behav Stat.} (2013) 38:267–94.

\bibitem{cognitivemodeling}Tatsuoka C, Tseng H, Jaeger J, Varadi F, Smith MA, Yamada T, Smyth KA,Lerner AJ. com AsDNInn: Modeling the heterogeneity in risk of progres-sion to Alzheimer’s disease acrosscognitive profiles in mild cognitiveimpairment. Alzheimer’s Res Ther. 2013;5:1–19. 

\bibitem{sequenceexp}Tatsuoka C. Sequential classification on lattices with experiment-specific response distributions. \textit{Sequent Anal Design Methods Appl.} (2014) 33:400–20.

\bibitem{neuronumeric}Tatsuoka C, McGowan B, Yamada T, Espy KA, Minich N, Taylor HG. Effects of extreme prematurity on numerical skills and executive function in kindergarten children: an application of partially ordered classification modeling. \textit{Learn Individ Differ.} (2016) 49:332–40.

\bibitem{daggingF} K. M. Ting, I. Witten, Stacking Bagged and Dagged Models (1997),\url{ https://www.researchgate.net/publication/2516354_Stacking_Bagged_and_Dagged_Models}

 \bibitem{tombaritda}F Tombari (2023), Tame representations in Topological Data Analysis, \url{https://www.diva-portal.org/smash/get/diva2:1759244/FULLTEXT01.pdf}.

\bibitem{hassDis} F. J. Torres-Rojas  \& Castro-Mora,  Partially Ordered Sets and Logical Clocks for Distributed Systems

\bibitem{fCAAutoOnt}Tovar, M., Pinto, D., Rendón, A.M., Serna, J.G., \& Ayala, D.V. (2014). Identification of Ontological Relations Using Formal Concept Analysis. \textit{LANMR}.

\bibitem{envdataH} Tsakovski, V. Simeonov, Hasse diagrams as explanatory tool in environmental data mining: A case study, in: J. Owsinski, R. Bruggemann (Eds.) \textit{Multicriteria Ordering and Ranking: Partial Orders}, Ambiguities and Applied Issues, Sys. Res. Inst. Polish Acad. Sci., Warsaw, 2007, pp. 50–68.

\bibitem{biomonitoring}Pirintsos, S., Bariotakis, M., Kalogrias, V., Katsogianni, S., Brüggemann, R. (2014). Hasse Diagram Technique Can Further Improve the Interpretation of Results in Multielemental Large-Scale Biomonitoring Studies of Atmospheric Metal Pollution. In: Brüggemann, R., Carlsen, L., Wittmann, J. (eds) Multi-indicator Systems and Modelling in Partial Order. Springer, New York, NY. \url{https://doi.org/10.1007/978-1-4614-8223-9_11}

\bibitem{surveysemisuperv}van Engelen, J.E., Hoos, H.H. A survey on semi-supervised learning. Mach Learn 109, 373–440 (2020). \url{https://doi.org/10.1007/s10994-019-05855-6}

\bibitem{envpollutemon}K. Voigt, G. Welzl, R. Bruggemann, Data analysis of environmental air pollutant monitoring systems in Europe, \textit{Environmetrics} \textbf{15}(2004) 577–596.

\bibitem{envpahrm}K. Voigt, R. Bruggemann, Ranking of pharmaceuticals detected in the environment:  Aggregation and weighting procedures, \textit{Combin. Chem. High Through. Screen.} \textbf{11} (2008) 770–782.

\bibitem{envSWsoils}K. Voigt, R. Bruggemann, M. Kirchner, K.-W. Schramm, Influence of altitude concerning the contamination of humus soils in the German Alps: a data evaluation
approach using PyHasse, \textit{Environ. Sci. Pollut. Res.} \textbf{17} (2010) 429–440.

\bibitem{envSWchem}K. Voigt, R. Bruggemann, H. Scherb, H. Shen, K. H. Schramm, Evaluating therelationship between chemical exposure and cryptorchidism by discrete mathematical method using PyHasse software,\textit{ J. Environ. Model. Soft.} \textbf{25} (2010) 1801–1812.

\bibitem{envpollutionMETEOR} Voigt, K.; Brüggemann, R. Water contamination with pharmaceuticals: data availability and evaluation approach with Hasse diagram technique and METEOR. MATCH Commun. Math. Comput. Chem. 2005, 54, 671-689.

 \bibitem{prorank2env}Voigt, Kristina et al. ``A multi-criteria evaluation of environmental databases using the Hasse Diagram Technique (ProRank) software.”  \textit{ Environ. Model. Softw. }21 (2006): 1587-1597.

\bibitem{envdatahealth} K. Voigt et al., Application of the PYHASSE Program Features: Sensitivity, Similarity, and Separability for Environmental Health Data, Statistica \& Applicazioni - Special Issue, 2011, pp. 155-168.

\bibitem{ccsilearn}X. Wang and A. Gupta. Unsupervised learning of visual representations using videos. In ICCV, 2015. 

\bibitem{gnnposet11}X. Wang, P. Li, M. Zhang, Improving Graph Neural Networks on Multi-node Tasks with Labeling Tricks, (2023) arXiv:2304.10074. Available in:  \url{https://arxiv.org/abs/2304.10074}

\bibitem{swiss} C. Wendler, Machine Learning on non-Euclidean domains: powersets, lattices, posets, (2023) ETH Zurich.

\bibitem{latticehiech}R. Wille, Restructuring lattice theory: An approach based on hierarchies of concepts, in: I. Rival (Ed.), \textit{Ordered Sets}, D. Reidel Publishing, Dordrecht, 1982, pp. 445–470.

 \bibitem{decisionsupportposet} J. Wittmann,  J. Markert,  S. Plura, R. Brüggemann, A Software Platform Towards a Comparison of Cars-A Case Study for Handling Ratio-Based Decisions, EnviroInfo 2011: Innovations in Sharing Environmental Observations and Information, Shaker Verlag Aachen. Available in: \url{http://enviroinfo.eu/sites/default/files/pdfs/vol6919/0467.pdf}.

\bibitem{mwafise2022poset}
A.~M. Mwafise, 
\emph{Poset Matrix Structure Via Partial Composition Operations}, 
arXiv preprint arXiv:2212.11644, 2022.

\bibitem{gnnsurvey}Wu et al., A Comprehensive Survey on Graph Neural Networks,(2019) arXiv:1901.00596.

\bibitem{clusterhres}Wu, C., Li, H., \& Ren, J. (2021). Research on hierarchical clustering method based on partially-ordered
Hasse graph. Future Generation Computer Systems. Doi: \url{https://doi.org/10.1016/j.future.2021.07.025}
 
\bibitem{Zhipeng}Z. Xie, W. Hsu, Z. Liu, M. L. Lee: Concept Lattice based Composite Classifiers for high Predictability. Artificial Intelligence, vol. 139, pp. 253–267, Wollongong, Australia (2002).

\bibitem{mlseqdata}Xiao, S., Yan, J., Farajtabar, M., Song, L., Yang, X., \& Zha, H. (2017). Joint Modeling of Event Sequence and Time Series with Attentional Twin Recurrent Neural Networks. ArXiv, abs/1703.08524.

 \bibitem{hypergraphpaper} Y. Yadav, A. Samal, E. Saucan, A Poset-Based Approach to Curvature of Hypergraphs,  \textit{Symmetry}, 14(2):420, 2022.

\bibitem{gnnposet22} M. Yang, R. Wang, Y. Shen, H. Qi and B. Yin, `` Breaking the Expression Bottleneck of Graph Neural Networks," in \textit{IEEE Transactions on Knowledge and Data Engineering}, vol. 35, no. 6, pp. 5652-5664, 1 June 2023, doi: \url{10.1109/TKDE.2022.3168070}.

\bibitem{grapFCA}Yoneda, Y., Sugiyama, M., \& Washio, T. (2018). Learning Graph Representation via Formal Concept Analysis. \textit{ ArXiv, abs/1812.03395}.

\bibitem{ensembleRR}H. Yu, Y.  Chen, P.  Lingras, G. Wang, A three-way cluster ensemble approach for large-scale data, International Journal of Approximate Reasoning, Volume 115, 2019, Pages 32-49, \url{https://doi.org/10.1016/j.ijar.2019.09.001}.


\bibitem{fuzzysetZ}L. A. Zadeh, Fuzzy Sets, \textit{Information and Control}, \textbf{8}, 338--353 (1965).

\bibitem{classifrules}J. Zhang , Learning to Rank Cases with Classification Rules(2008),
\url{http://www.ecmlpkdd2008.org/files/pdf/workshops/pl/8.pdf}.

 \bibitem{computeconc}L. Zhang,  D. Wang , X.  Liu,   Computing Concept Lattices with Clustering Approaches.

\bibitem{fcaonto5}Zhao, M., Zhang, S., Li, W. et al. Matching biomedical ontologies based on formal concept analysis. J Biomed Semant 9, 11 (2018). \url{https://doi.org/10.1186/s13326-018-0178-9}.

\bibitem{ziaTDA}Zia, A., Khamis, A., Nichols, J. et al. Topological deep learning: a review of an emerging paradigm. Artif Intell Rev 57, 77 (2024).\url{ https://doi.org/10.1007/s10462-024-10710-9}.

\bibitem{stasdepth} Y. Zuo and R.  Serfling, General Notions of Statistical Depth Function, \textit{The Annals of Statistics}, Vol. 28, No. 2 (Apr., 2000), pp. 461-482.

 \bibitem{ucidata}UCI machine learning repository, \url{http://www.ics.uci.edu/~mlearn/MLRepository.html.}

\bibitem{selfmap}Kohonen, Teuvo (January 2013). "Essentials of the self-organizing map". \textit{Neural Networks.}  37: 52–65. doi:1\url{0.1016/j.neunet.2012.09.018}. 

\bibitem{mp11pose} MPII Human Pose Dataset, \url{http://human-pose.mpi-inf.mpg.de/}.

\bibitem{lspdataset} Leeds Sports Poset Dataset \url{https://datasets.activeloop.ai/docs/ml/datasets/lsp-dataset/}.

\bibitem{olympicdata} Olympic Sports Dataset \url{http://vision.stanford.edu/Datasets/OlympicSports/}.

\bibitem{cfqdataset} Compositional Freebase Questions (CFQ) Dataset,\url{https://github.com/google-research/google-research/tree/master/cfq}

\bibitem{UDC} Universal Dependencies Corpora \url{https://www.tensorflow.org/datasets/catalog/universal_dependencies}

 \bibitem{equidata} Istat. (2018a). Indicators for the measurement of equitable and sustainable well-being. \url{ https://www.istat.it/en/well-being-and-sustainability/the-measurement-of-well-being/indicators}.

 \bibitem{servicedata} Eurobarometer dataset for service performance \url{https://www.gesis.org/en/eurobarometer-data-service/}

 \bibitem{pythonVS}Python implementaion of deep supervised visual similarity learning(2017) In :\url{https://github.com/asanakoy/deep_unsupervised_posets} .

 \bibitem{causaldag}C.  Squires, causaldag (2018), \url{https://github.com/uhlerlab/causaldag}.


 \bibitem{hypergraphdata} Poset-Hypergraph-Curvature Datasets, \url{https://github.com/asamallab/Poset-Hypergraph-Curvature}.









\end{thebibliography}
\end{document}